\documentclass[authoryear,a4paper]{elsarticle}

\usepackage{verbatim}

\usepackage{geometry}

\usepackage{color,soul}

\usepackage[hidelinks]{hyperref}
\hypersetup{
  colorlinks   = true,
  urlcolor     = blue,
  linkcolor    = blue,
  citecolor    = blue
}

\usepackage{graphicx}
\graphicspath{{./figs/}}
\DeclareGraphicsExtensions{.pdf,.png,.jpg}
\usepackage{placeins}
\usepackage{caption}
\usepackage{subcaption}

\usepackage{booktabs}
\usepackage[table,xcdraw]{xcolor}
\usepackage{graphicx}
\usepackage{adjustbox}
\usepackage{multirow}

\usepackage{amssymb}
\usepackage{gensymb}
\usepackage[mathscr]{euscript}
\usepackage{bm}
\usepackage{mathtools}
\usepackage{nccmath}
\usepackage{amsmath}
\usepackage{braket}

\usepackage{algorithm}
\usepackage{algpseudocode}

\usepackage{lineno, hyperref}

\usepackage{lipsum}

\usepackage{lineno}

\journal{International Journal of Applied Earth Observation and Geoinformation}

\begin{document}

\captionsetup[figure]{labelfont={bf},labelformat={default},labelsep=period,name={Fig.}}
\captionsetup[table]{labelfont={bf},labelformat={default},labelsep=newline,name={Table}}

\begin{frontmatter}
\title{Meta-learning an Intermediate Representation for Few-shot Block-wise Prediction of Landslide Susceptibility}
\author[swjtu]{Li Chen}
\author[swjtu]{Yulin Ding\corref{cor1}}
\author[swjtu]{Saeid Pirasteh}
\author[swjtu]{Han Hu}
\author[swjtu]{Qing Zhu}
\author[swjtu]{Xuming Ge}
\author[swjtu]{Haowei Zeng}
\author[swjtu]{Haojia Yu}
\author[swjtu]{Qisen Shang}
\author[nsmilr]{Yongfei Song}

\cortext[cor1]{Corresponding Author: dingyulin@swjtu.edu.cn}

\address[swjtu]{Faculty of Geosciences and Environmental Engineering, Southwest Jiaotong University, Chengdu, China}
\address[nsmilr]{Ningxia Survey and Monitor Institute of Land and Resources, Yinchuan, China}

\begin{abstract}
    Predicting a landslide susceptibility map (LSM) is essential for risk recognition and disaster prevention.
    Despite the successful application of data-driven approaches for LSM prediction, most methods generally apply a single global model to predict the LSM for an entire target region.
    However, in large-scale areas with significant environmental change, various parts of the region hold different landslide-inducing environments, and therefore, should be predicted with respective models.
    This study first segmented target scenarios into blocks for individual analysis.
    Then, the critical problem is that in each block with limited samples, conducting training and testing a model is impossible for a satisfactory LSM prediction, especially in dangerous mountainous areas where landslide surveying is costly.
    To solve the problem, we trained an intermediate representation by the meta-learning paradigm, which is superior for capturing information valuable for few-shot adaption from LSM tasks. 
    We hypothesized that there are more general and vital concepts concerning landslide causes and are sensitive to variations in input features.
    Thus, we can quickly adapt the models from the intermediate representation for different blocks or even unseen tasks using very few exemplar samples.
    Experimental results on the two study areas demonstrated the validity of our block-wise analysis in large scenarios and revealed the top few-shot adaption performances of the proposed methods.
\end{abstract}

\end{frontmatter}


\section{Introduction}
\label{s:introduction}

Landslide susceptibility evaluates the likelihood of landslide occurrence, is useful for potential disaster recognition and prevention, land-use planning, and regional decision \citep{reichenbach2018review, ray2010landslide}.
Considering whether to involve the landslide mechanism, the landslide susceptibility prediction methods are classified into handcrafted-feature-based and data-driven approaches.
Handcrafted features \citep{van2003use} are artificially designed using expert knowledge associated with a complex mechanism of landslide causes.
Data-driven methods \citep{merghadi2020machine} collect samples from geographic spatial data and train their models to predict landslide susceptibility.
These methods \citep{huang2018review, zhu2021depth, van2020spatially, pham2019landslide, zhu2020map, merghadi2018landslide, ge2021target} gradually spring up in part because of the access to masses of thematic information benefited by the development of remote sensing technologies, and in part because of the deep learning method's automaticity in exploring innate causative aspects of a landslide from data analysis.

Currently, most data-driven methods predict the LSM based on \textit{global analysis}, denoted in this study as using a single model to predict landslide susceptibility of an administrative region. 
Most of these methods can provide good predictive results based on the premise that sufficient landslide/non-landslide points are fed as input samples, and each part of the target area shares similar landslide causes.  
However, considering large-scale scenarios, such as a whole province in China or other countries and regions with more than 100,000 square kilometers, a landslide-inducing environment would be expected to vary due to the territory's large span. 
In this case, the global model inferred from a specific region would fail to generalize to other areas, thus losing veracity or even rationality. 
There are a few works \citep{shahri2019landslide, shahri2021landslide} dividing study areas into pieces for analysis, but lack of consideration when there are very few samples.
Generally, adequate landslide records are available only in certain administrative regions (a few parts of large scenarios) \citep{qing2019review}, which would lead to the training of the model suffering from limited labeled samples in other regions. 
Accordingly, the principal objective of this study is to design a strategy for predicting the LSM to raise the rationality of modeling the overall landslide-inducing environment and overcoming the problem of having few available samples.

\begin{figure}[t]
	\centering
	\includegraphics[width=\textwidth]{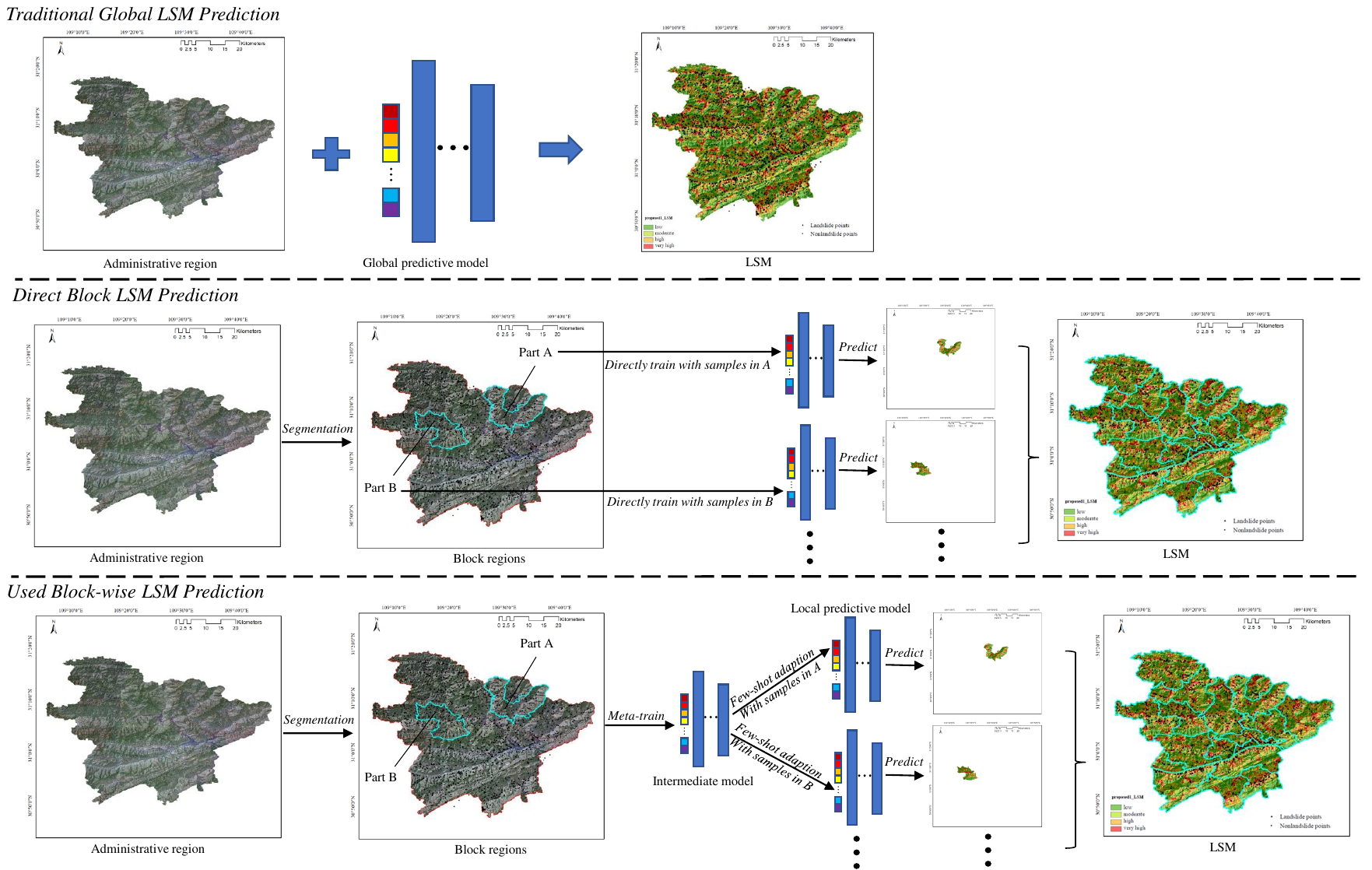}
	\caption{Three manners of predicting the LSM. Traditional global LSM prediction could lose veracity or even rationality in complex large scenarios with different landslide-inducing environments. 
    When considering segmenting the scenarios for individual analysis, directly training predictive models for each block would suffer from having few available samples. 
    Our method trains an intermediate model that can be quickly adapted to each block.}
	\label{fig:global_vs_local}
\end{figure}

\subsection{Limited generalization ability of a single model}
\label{subs:limited generalization ability of a single model}
\textit{Global analysis} can elevate the fitting performance of a single model on given samples, but it limits the transferability of the model for being generalized to various scenarios. 
Because the landslide-inducing environments of various parts of a region may be different, it might be unreasonable to use only one model to represent multiple landslide causes. 
For example, suppose that the occurrence of a landslide in one region is extremely sensitive to the slope, while the occurrence in other regions is not. 
The model fed with similar input features should give the same output, but in this case, the occurrence of a landslide in two areas could be different states, that one input caused a landslide prediction, whereas the other did not. 
One possible reason is that the dominant causative factors, for example, lithology or rainfall, resulting in such discrepancy were not captured as feeding to the training model. 
Intuitively, to solve the generalization limitation, we divide the region into blocks and build models, respectively.
We denoted this as block-wise analysis in this article.

\subsection{Few samples for block-wise analysis}
\label{subs:few samples for block-wise analysis}
In large mountainous scenarios, access to landslide events is costly and limited.
Because the collection of historical or assumptive landslide points requires not only expert knowledge but also involves an investigation of the probably dangerous site \citep{ciampalini2016landslide}.
Therefore, the related issue is how the current model can be quickly adapted to unseen areas with only a few labeled samples. 
For this question, we considered whether there is an intermediate representation that is conditionally sensitive to variations in some aspects concerning general and vital landslide causes; thus, within these concepts, the representation could be few-shot adapted to new tasks.

\subsection{Contribution}
To solve the problems mentioned above, we proposed to decompose the global analysis into \textit{local analysis} (whole study area into blocks, single task into multitasks), and meta-learns an intermediate representation that can be few-shot adapted to each task. 
The proposed pipeline contains unsupervised pretraining, scene segmentation and meta-task sampling, intermediate model meta-training, and few-shot adaption. 
Specifically, we first execute unsupervised pretraining with unlabeled samples to initialize the base model for subsequent meta-learning, with good properties, such as noise proof, disentangling correlated input features, stability to variations in training configuration, improving transferability of the base model, and embedding a more discriminative metric space. 
Then, to recast the \textit{global analysis} and the model, we utilized a modified simple linear iterative clustering (SLIC) \citep{achanta2012slic} to effectively segment the large-scale scenario into \textit{superpixels} for block-wise analysis. 
Finally, for the validity of applying meta-learner to solve the few-shot learning problem \citep{sung2018learning}, we meta-learn an intermediate model so that local tasks in unseen areas could be quickly adapted. 
Meta-learning, as well as learning to learn, means learning concepts that can better or faster guide the model to learn desired representations.
The intuition is that a more general representation is better suited for landslide susceptibility few-shot adaption.

The main contributions are listed as follows: 
(1) We propose a block-wise LSM approach to raise the veracity for modeling different parts of a scenario;
and (2) more crucial is that we propose to meta-learn an intermediate representation that can be quickly adapted to unseen tasks with very few samples and gradient descent updates. 
It is feasible and efficient for large-scale LSM tasks where landslide records are limited.
The remainder of this paper is organized as follows: 
In Section \ref{s:related works}, related works are reviewed.
In Section \ref{s:study areas and datasets}, the study area and dataset are introduced, while Section \ref{s:proposed method} presents the proposed method. 
Results and discussions are presented in Section \ref{s:experiments and analysis}, and section \ref{s:conclusion} concludes the paper.

\section{Related works}
\label{s:related works}

This section reviews related statistical landslide susceptibility prediction methods and summarizes their characteristics.

\paragraph{Supervised Regression}
Supervised regression refers to the prediction of the LSM with end-to-end nonlinear mapping trained by labeled samples \citep{chang2020landslide}.
These methods are widely applied, for example, logistical regression \citep{chen2018landslide}, decision trees \citep{hong2020modeling}, support vector machines \citep{pham2019novel}, and evidential belief functions \citep{chen2019landslide}.
Despite their outstanding performance in the face of simple scenarios, these models would lose veracity to fit the nonlinearity of complex landslide causes.
To improve the representative power, researchers have either applied ensemble techniques \citep{hu2021performance} or increased the depth and width of the model. 
For example, in \cite{pham2017hybrid}, different base classifiers were combined, and in \cite{wyner2017explaining}, an expanded multilayer perceptron (MLP) was applied, both to increase the precision of fitting the training data in a complex environment.
The state-of-the-art random forest (RF) \citep{dou2019assessment} is also regarded as a bagging method in ensemble techniques. 
Recently, convolutional neural networks (CNN) have also been applied for landslide susceptibility prediction \citep{fang2020integration, yi2020landslide}.
These methods are competent, but they can suffer from the deficiency of labeled samples to train large-capacity models in practice.
In addition, poor quality of thematic information, such as non-uniform resolution, noise disturbance, the less rigorous feeding of correlated thematic features, and inaccurate supervision where labeled samples are not always the ground truth, would adversely impact predicting reasonable landslide susceptibility.

\paragraph{Unsupervised Representation Learning}
Unsupervised learning takes a good insight into the unlabeled data structure and learns representations that are more suitable as a side effect for subsequent supervised learning \citep{amruthnath2018research}
Some landslide susceptibility evaluations applied unsupervised representation learning before training the labeled samples.
For example, \cite{xu2015stacked} used a sparse autoencoder to reduce redundancy and correlation in the conditioning factors and applied regularized greedy forests \citep{sameen2020landslide} to predict landslide susceptibility with sparse representation.
\cite{wang2020mapping} applied a deep belief network to disentangle independent concepts in a relatively low-dimensional space.
Impressively, \cite{zhu2020unsupervised} applied an adversarial manner to urge the model to learn a representation that shares as many underlying factors between source and target scenarios, which improved the transferability of the model to unseen areas.
These unsupervised techniques open up a new way for potential hazard analysis.
However, they neglect the irrationality to apply one global model to describe multi-environmental tasks and the very limited samples situation.

\paragraph{Meta Learning}
Learning quickly from very few samples is a hallmark of human intelligence \citep{hospedales2020meta}. For example, a child would recognize a horse picture based on the previous reading of an animal brochure.
In contrast to the classifier trained with thousands of labeled samples, meta-learning aims to learn more general concepts among the tasks so that the desired model can be learned from only a few samples. Different from uninterpretable output trained by black-box supervised regression,
it is worth researching if a landslide can also be quickly identified based on some meta-trained cognitive patterns.
Model-agnostic meta-learning (MAML) \citep{finn2017model} is a heuristic and typical work of meta-learning science, which innerly trains each meta-task and whose meta-objective is to minimize the sum of losses from related tasks.
To generalize the ideology of meta-training, \cite{metz2018meta} introduced an unsupervised weight update rule that applies an MLP for hyperparameter meta-training.
Despite its superiority, few studies have discussed the potential and advantages of applying meta-learning techniques to solve the problems above in LSM prediction.
In this study, we explored the superiority of meta-learning in capturing general concepts from LSM tasks and tested the few-shot adaption and generalization performance.

\section{Study areas and datasets}
\label{s:study areas and datasets}

\begin{figure}[htp]
    \centering
    \begin{subfigure}{0.48\textwidth}
        \includegraphics[width=\linewidth]{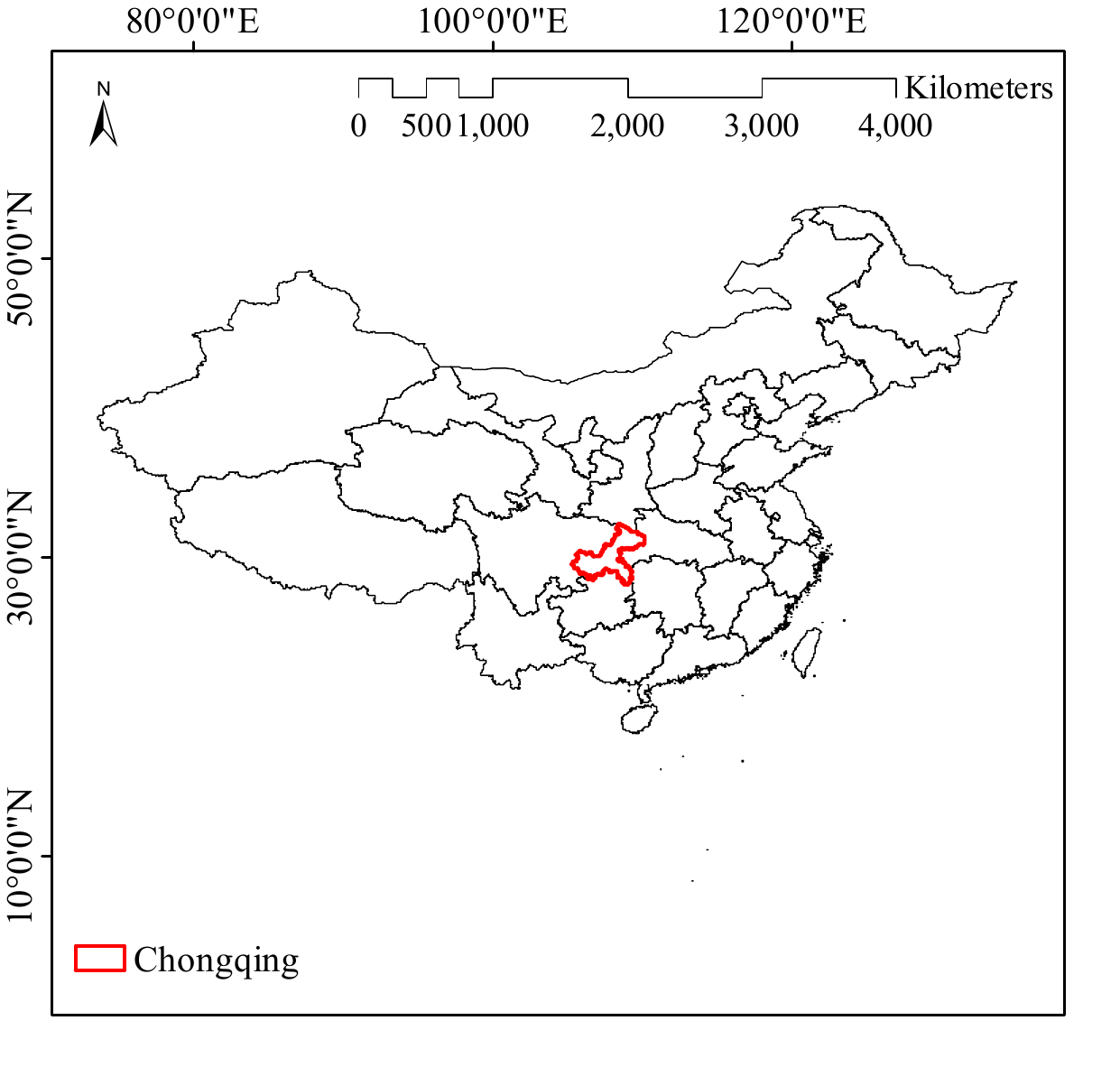}
        \caption{Administrative districts of China}
    \end{subfigure}
	\begin{subfigure}{0.48\textwidth}
        \includegraphics[width=\linewidth]{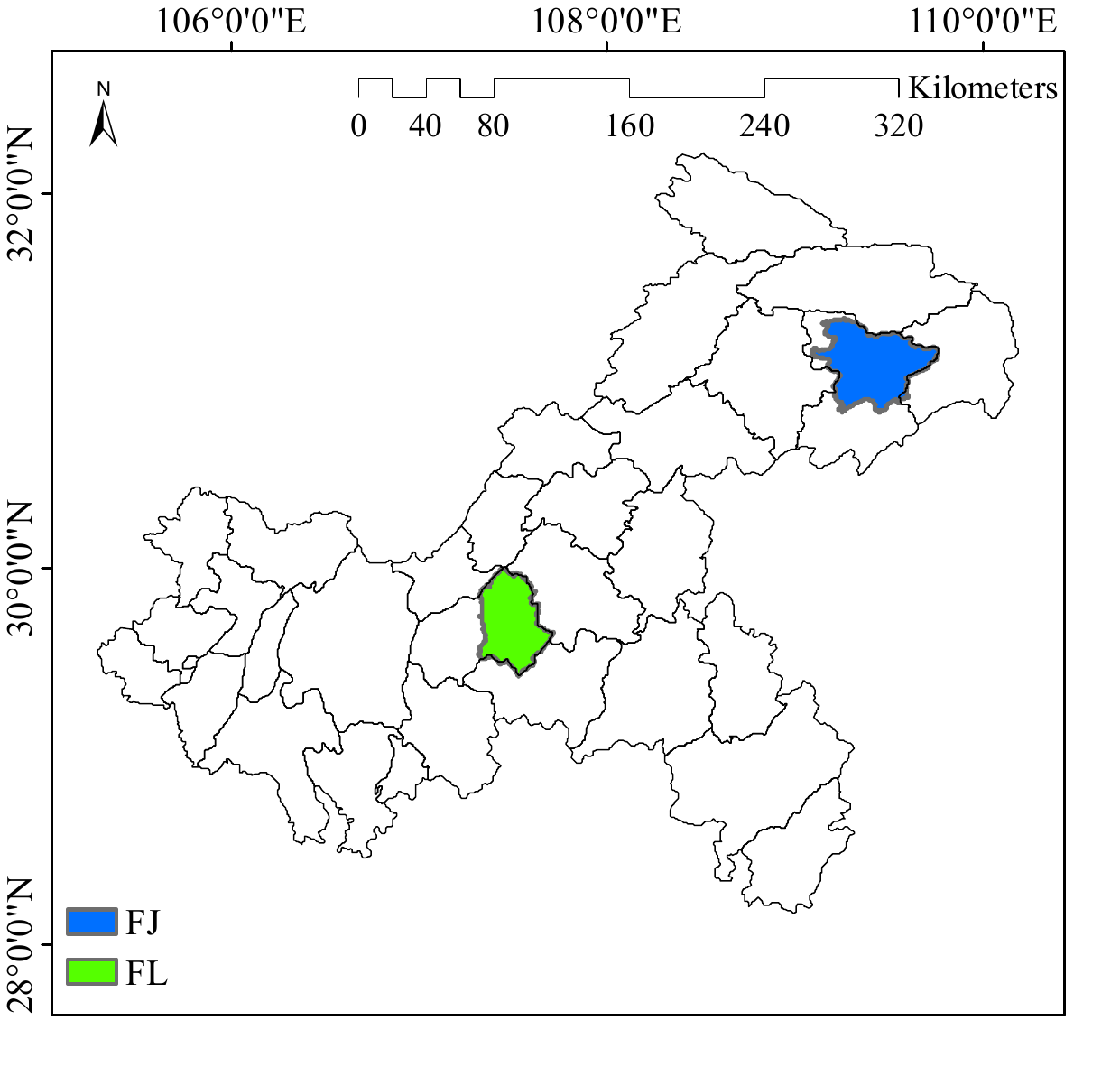}
        \caption{Chongqing City}
    \end{subfigure}
    \begin{subfigure}{0.48\textwidth}
        \includegraphics[width=\linewidth]{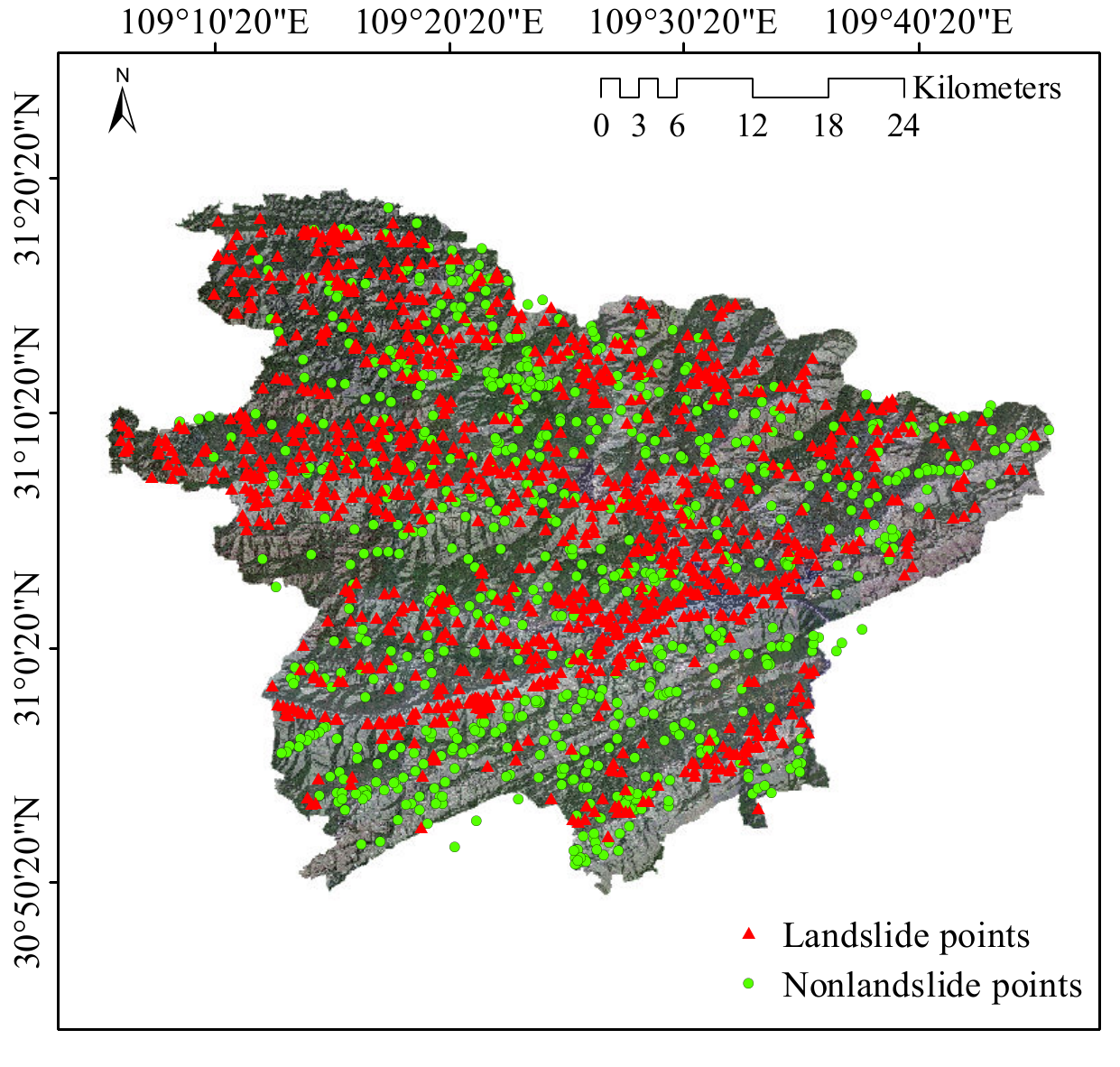}
        \caption{Fengjie County (FJ)}
    \end{subfigure}
    \begin{subfigure}{0.48\textwidth}
        \includegraphics[width=\linewidth]{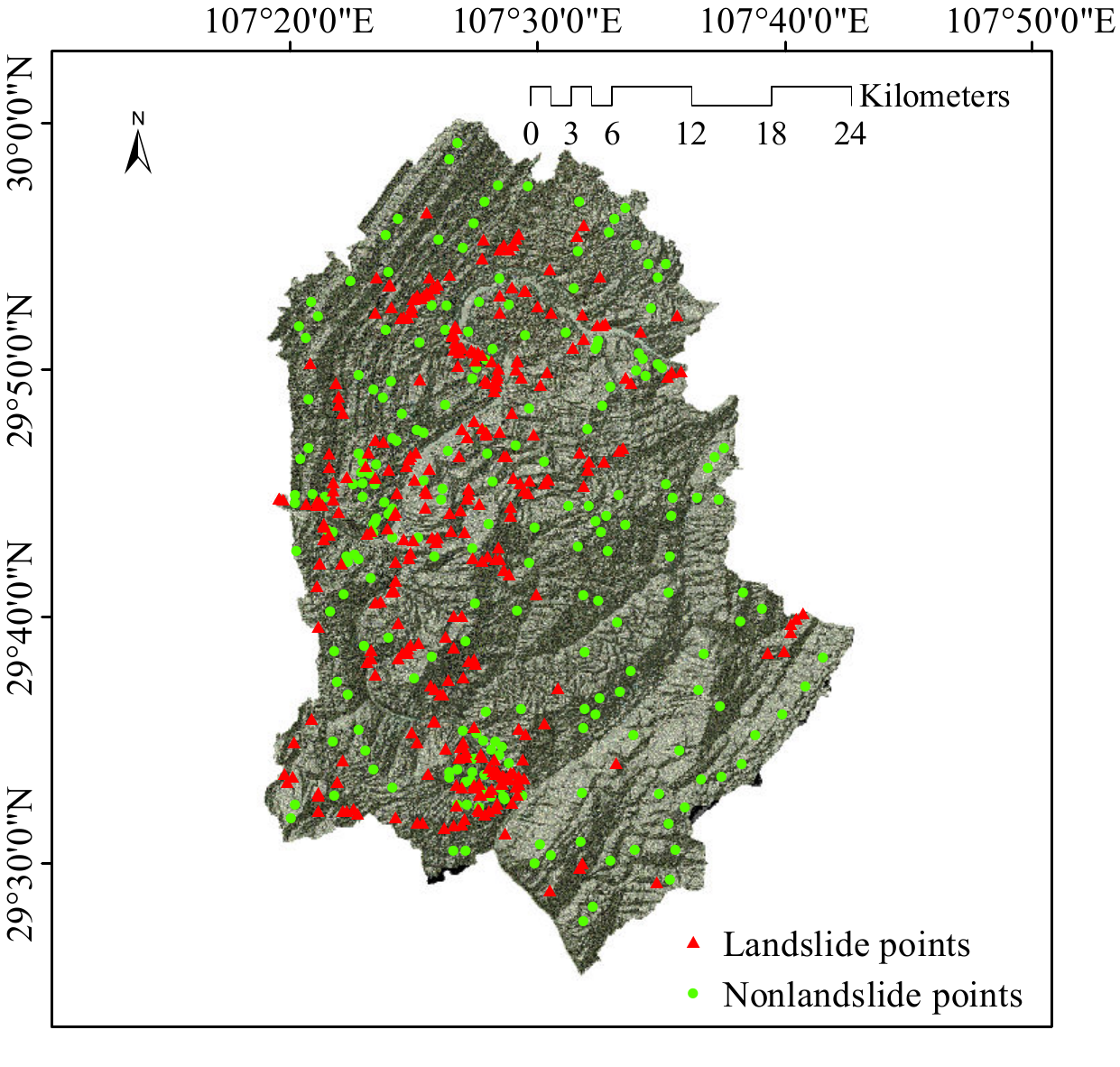}
        \caption{Fuling District (FL)}
    \end{subfigure}
    \caption{Location of the two study areas FJ and FL: (a) administrative map of China and the location of Chongqing City; (b)admistrative map of Chongqing City and the locations of FJ and FL; (c) (non)landslide distribution in FJ and FL.}
    \label{fig:location}
\end{figure}

There were two study areas (Fig. \ref{fig:location}), the Fengjie County (FJ) and Fuling District (FL), both located in Chongqing City, China.
Chongqing City is the largest city in China and features mountainous areas with complex landslide-inducing environments.
In the past, many devastating landslide events have occurred, for example, the typical Tangjiewan, Xinmo, and Mabian landslides \citep{fan2018analyzing, scaringi2018some, wei2019numerical}.

The landslide inventory was collected from the Chongqing Geomatics and Remote Sensing Center. 
In addition to historical landslide records, potential landslide locations inferred by expert knowledge and on-site investigation were also collected. 
There were approximately 1200 available landslide records in FJ, and approximately 400 landslide records for FL, as positive samples. 
We randomly selected the same amount of non-landslide points as the negative samples.

\begin{figure}[H]
    \centering
    \begin{subfigure}{0.24\textwidth}
        \includegraphics[width=\linewidth]{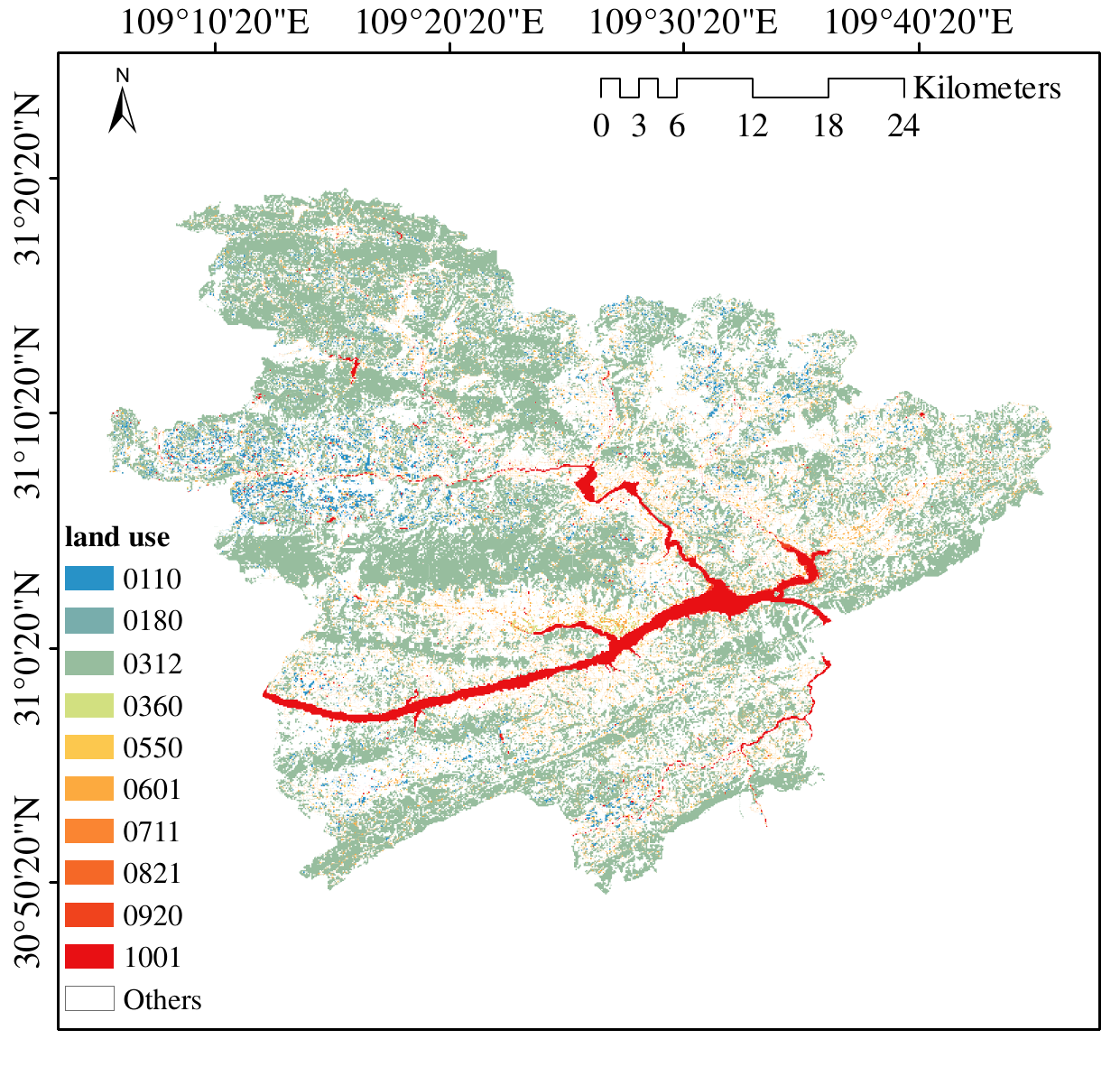}
        \caption{land use}
    \end{subfigure}
    \begin{subfigure}{0.24\textwidth}
        \includegraphics[width=\linewidth]{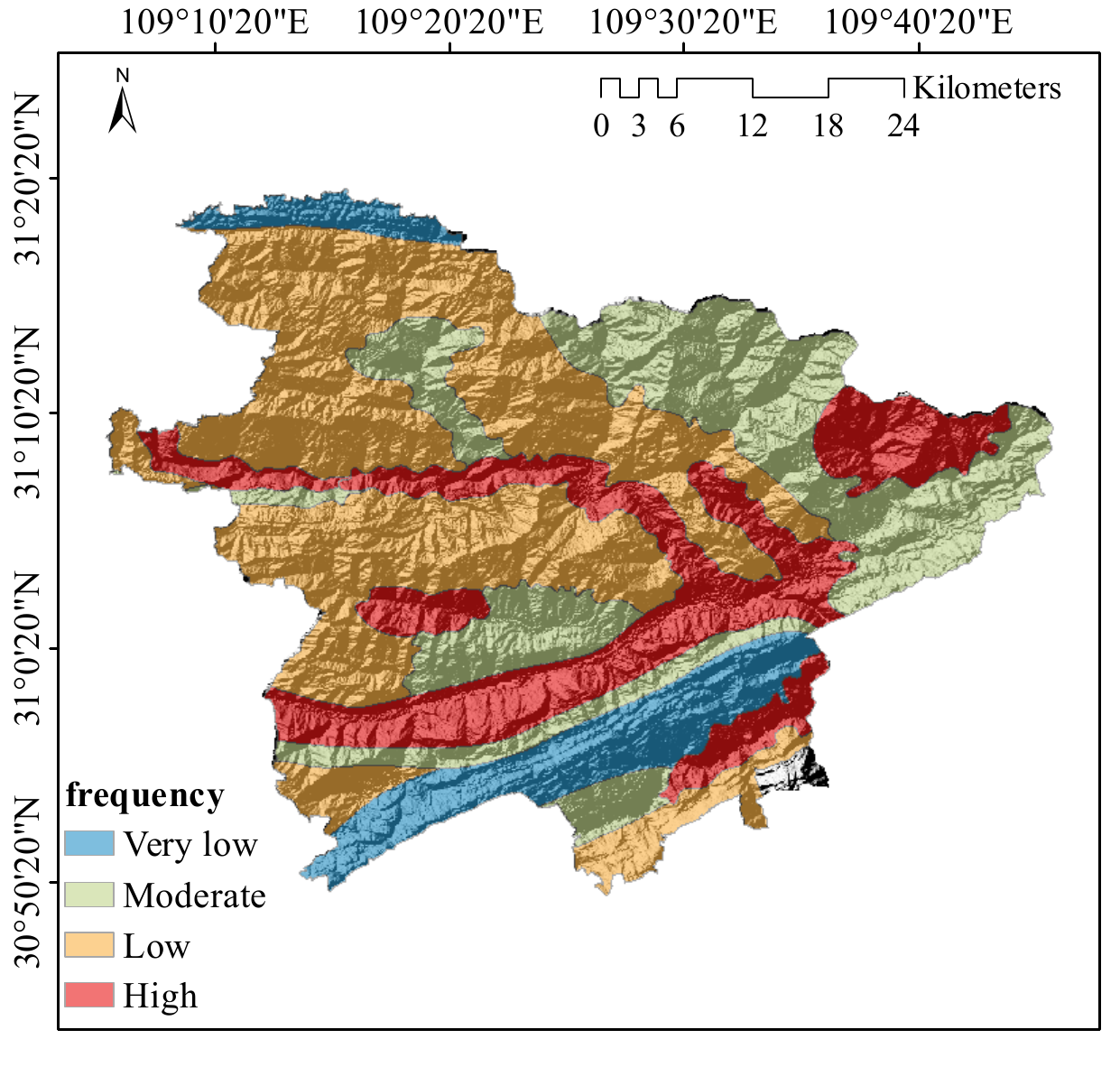}
        \caption{frequency}
    \end{subfigure}
    \begin{subfigure}{0.24\textwidth}
        \includegraphics[width=\linewidth]{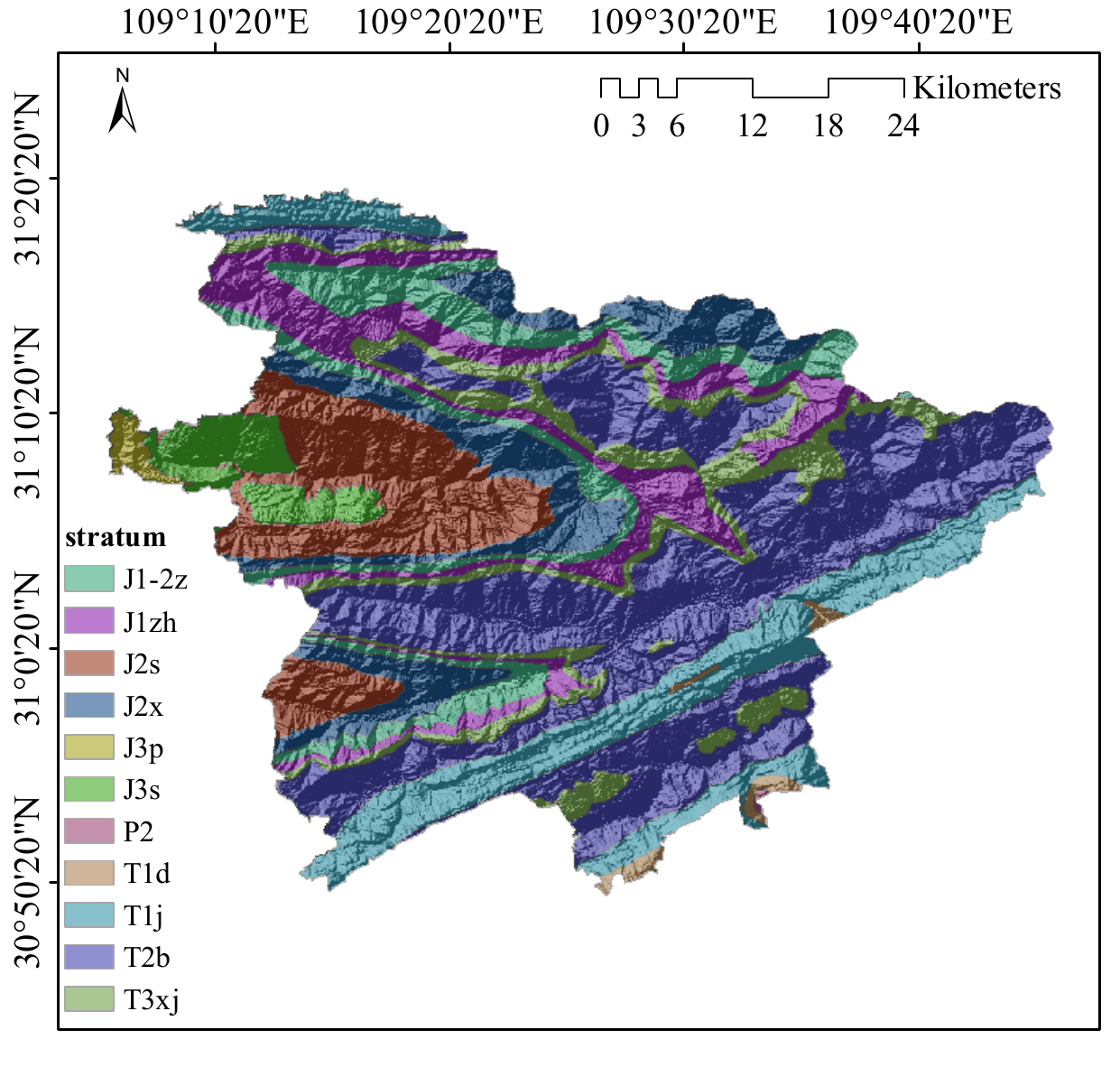}
        \caption{strata type}
    \end{subfigure}
	\begin{subfigure}{0.24\textwidth}
        \includegraphics[width=\linewidth]{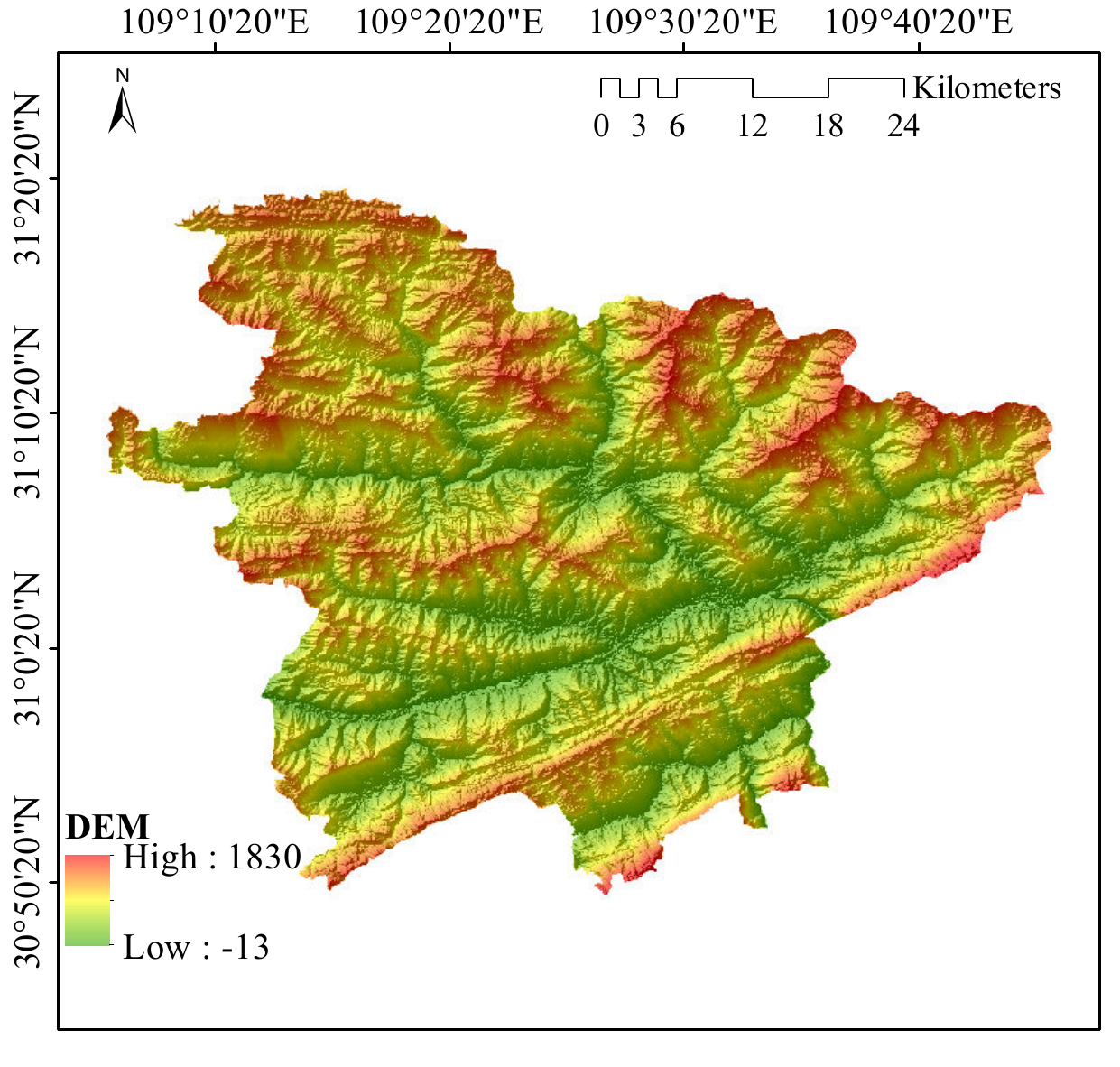}
        \caption{DEM}
    \end{subfigure}
    \begin{subfigure}{0.24\textwidth}
        \includegraphics[width=\linewidth]{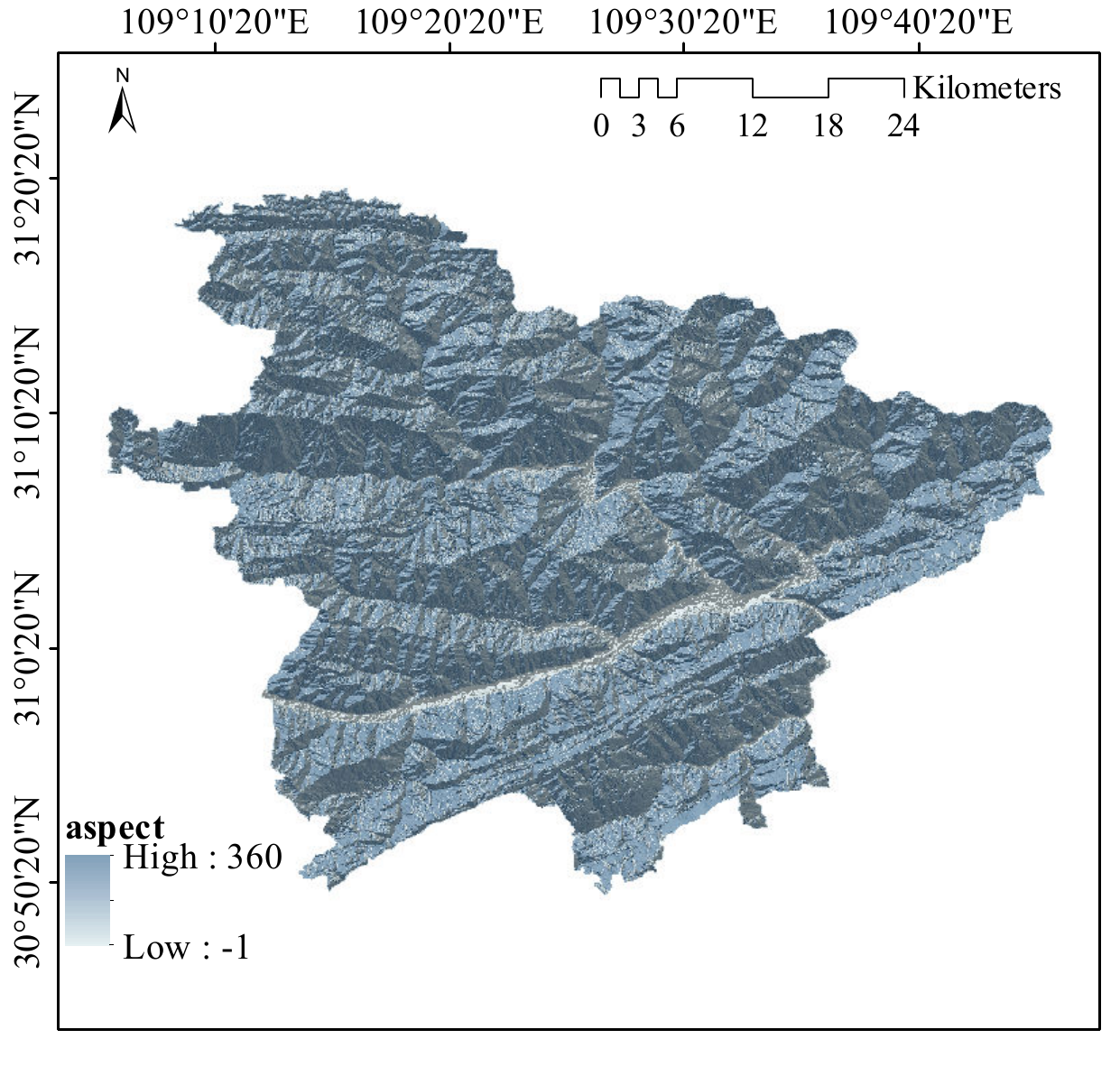}
        \caption{aspect}
    \end{subfigure}
    \begin{subfigure}{0.24\textwidth}
        \includegraphics[width=\linewidth]{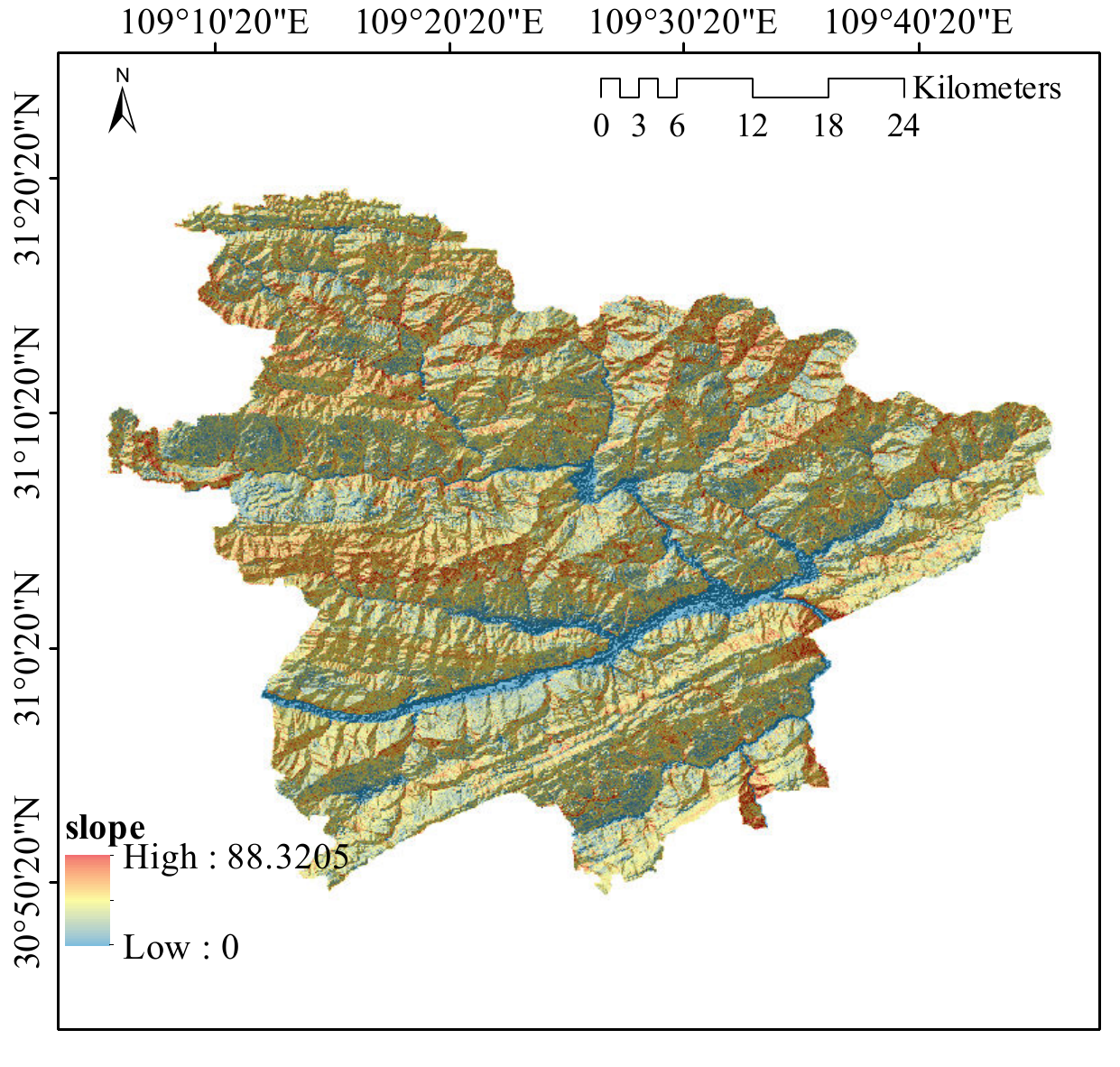}
        \caption{slope}
    \end{subfigure}
    \begin{subfigure}{0.24\textwidth}
        \includegraphics[width=\linewidth]{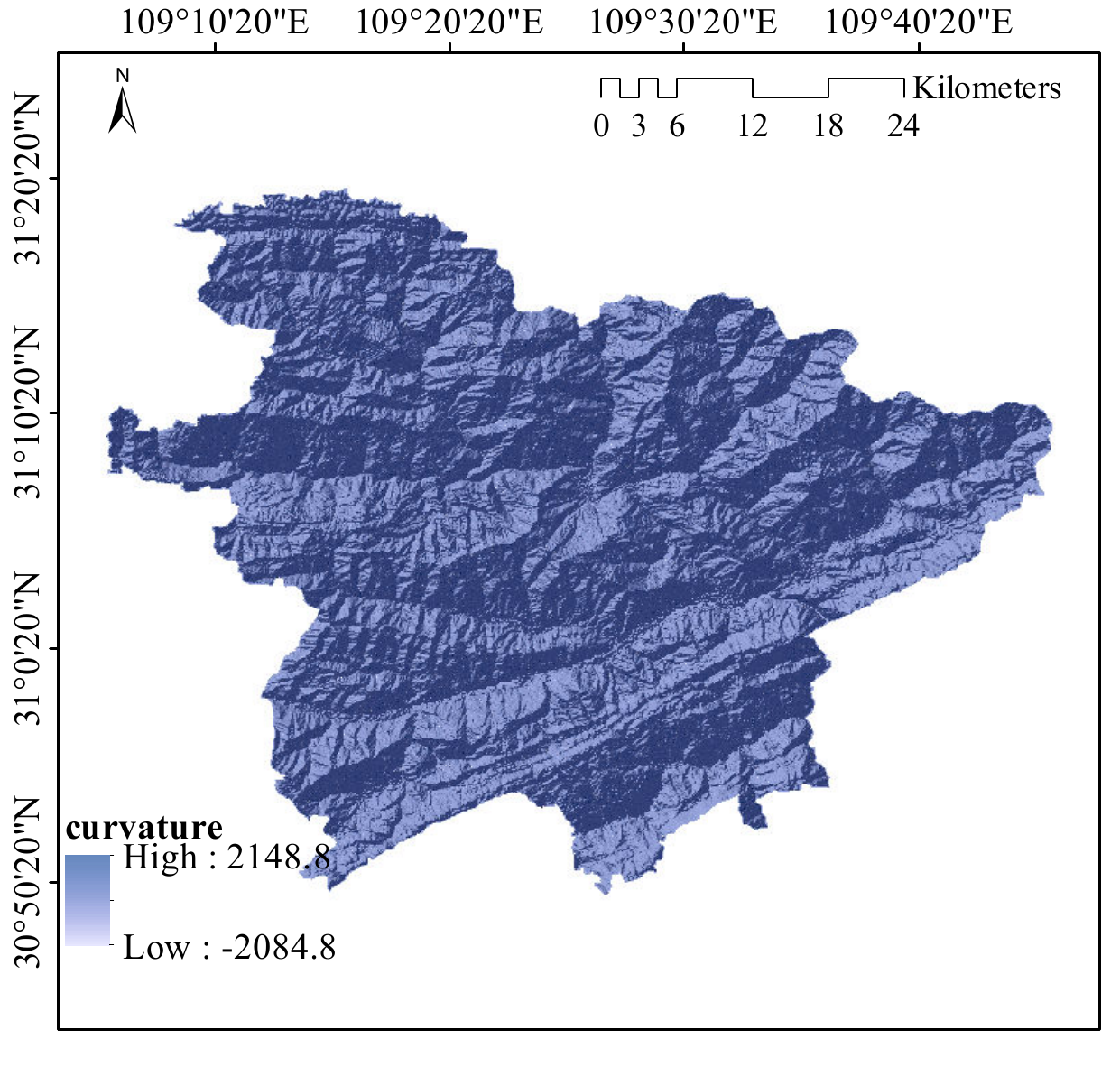}
        \caption{curvature}
    \end{subfigure}
    \begin{subfigure}{0.24\textwidth}
        \includegraphics[width=\linewidth]{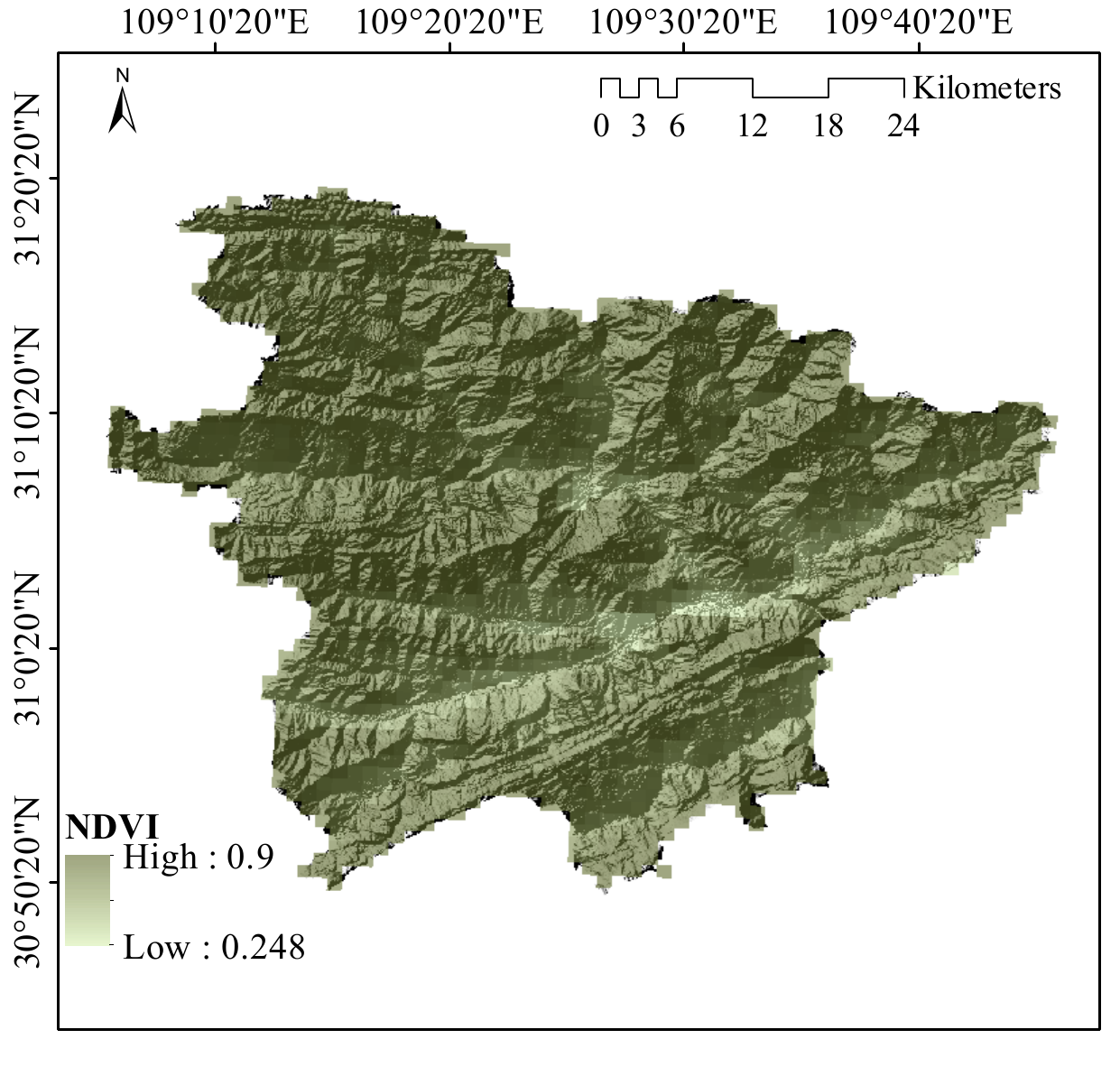}
        \caption{NDVI}
    \end{subfigure}
    \begin{subfigure}{0.24\textwidth}
        \includegraphics[width=\linewidth]{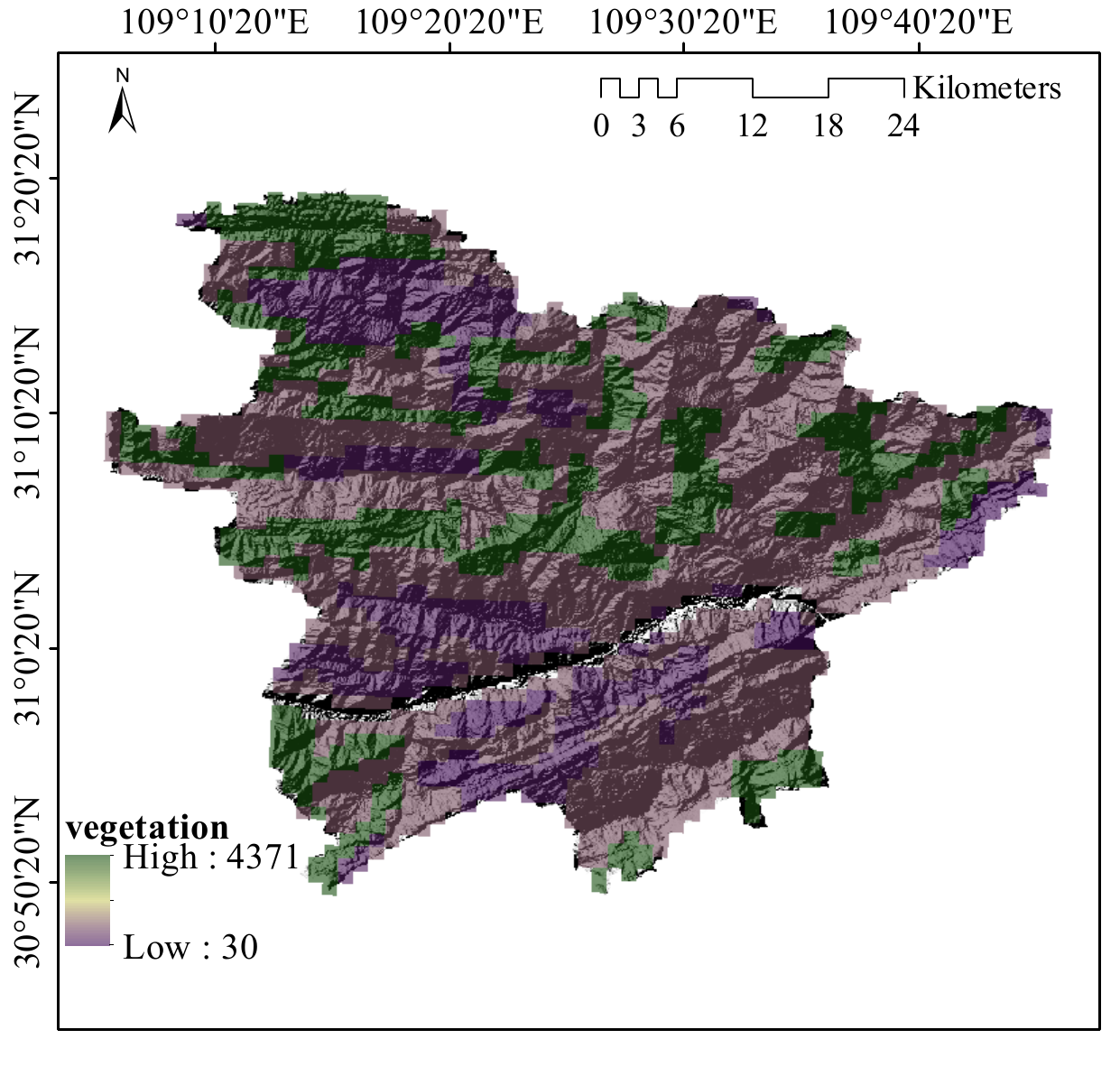}
        \caption{vegetation coverage}
    \end{subfigure}
    \begin{subfigure}{0.24\textwidth}
        \includegraphics[width=\linewidth]{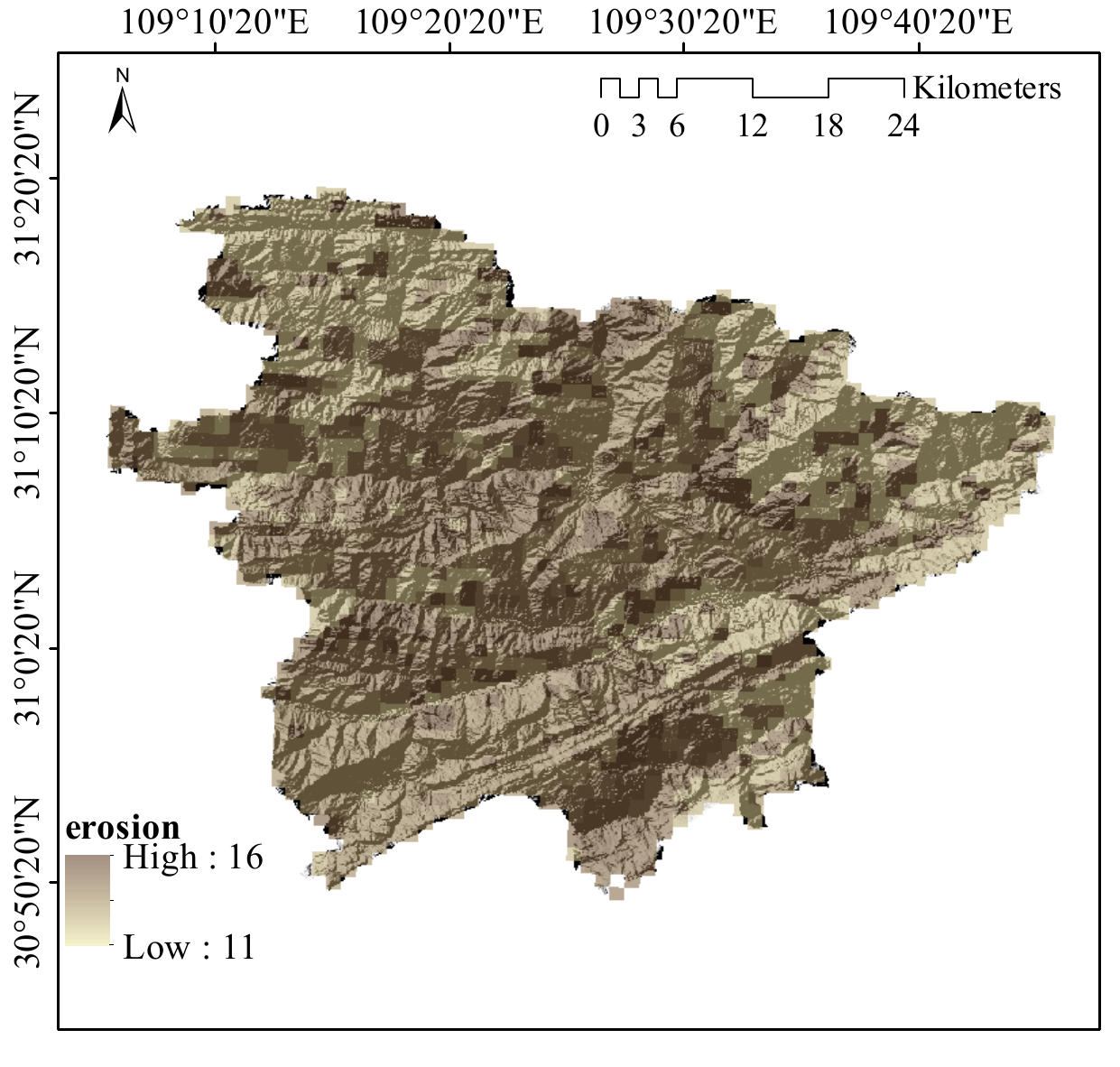}
        \caption{soil erodibility}
    \end{subfigure}
    \begin{subfigure}{0.24\textwidth}
        \includegraphics[width=\linewidth]{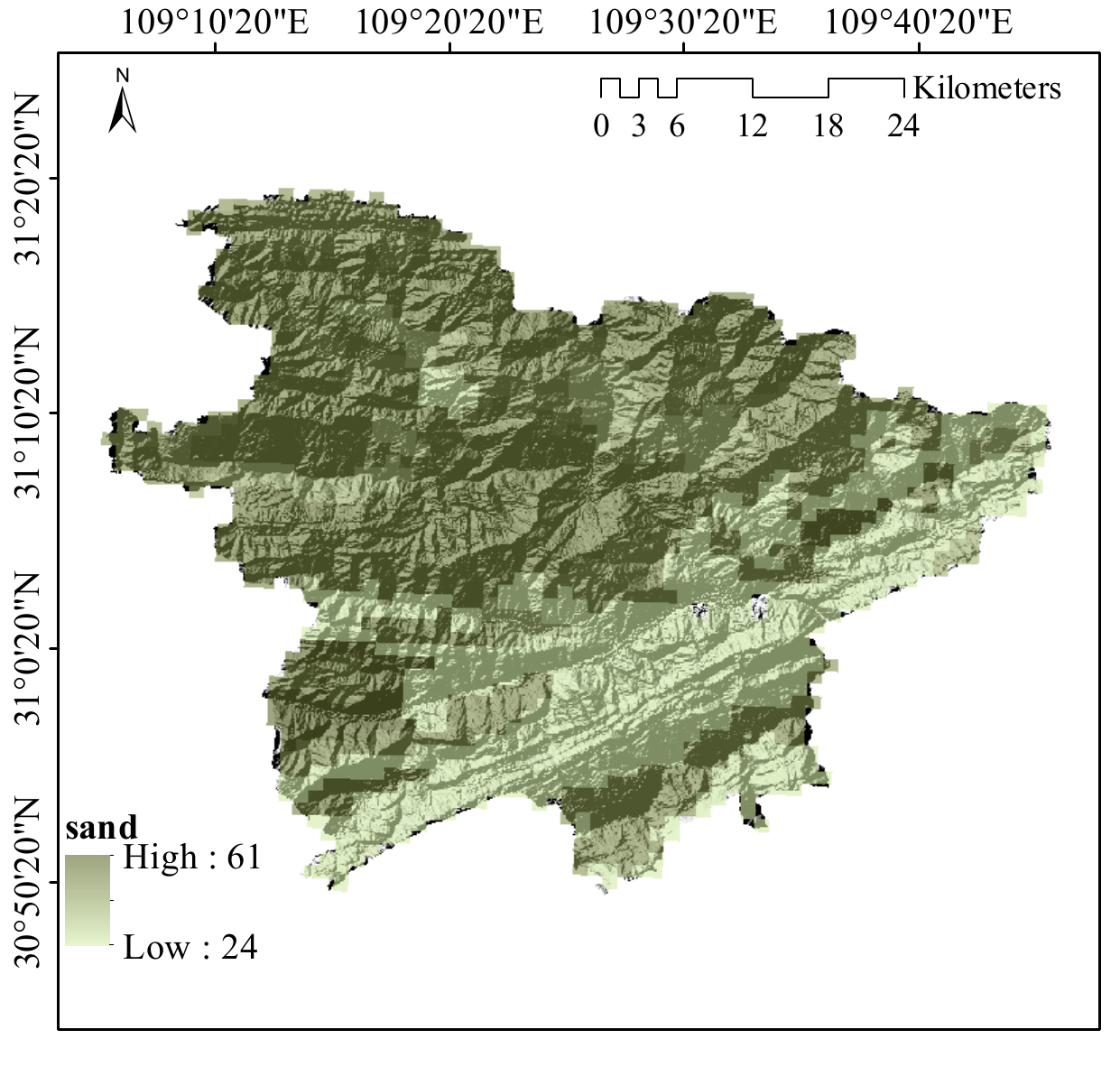}
        \caption{sand coverage}
    \end{subfigure}
    \begin{subfigure}{0.24\textwidth}
        \includegraphics[width=\linewidth]{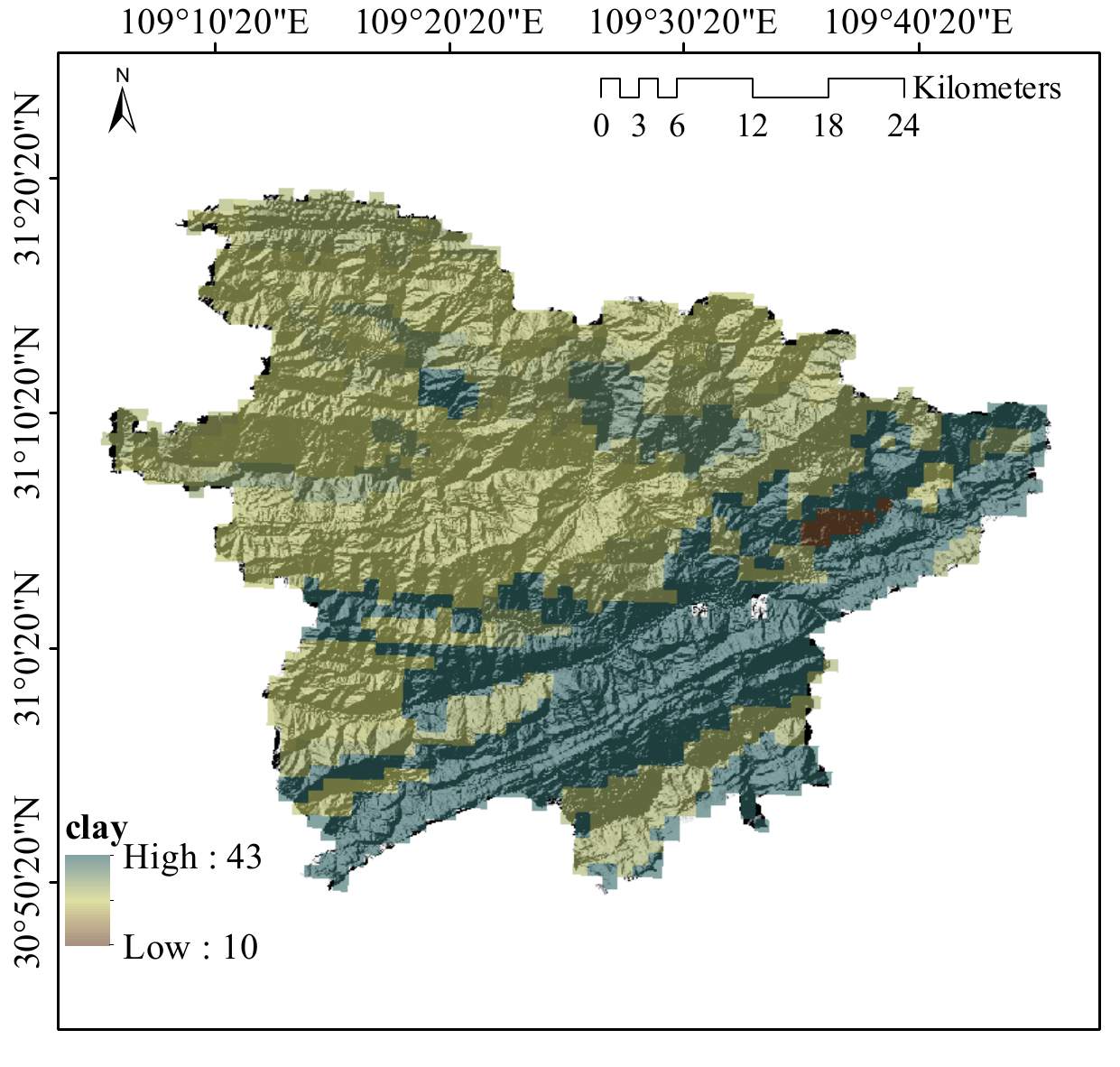}
        \caption{clay coverage}
    \end{subfigure}
    \begin{subfigure}{0.24\textwidth}
        \includegraphics[width=\linewidth]{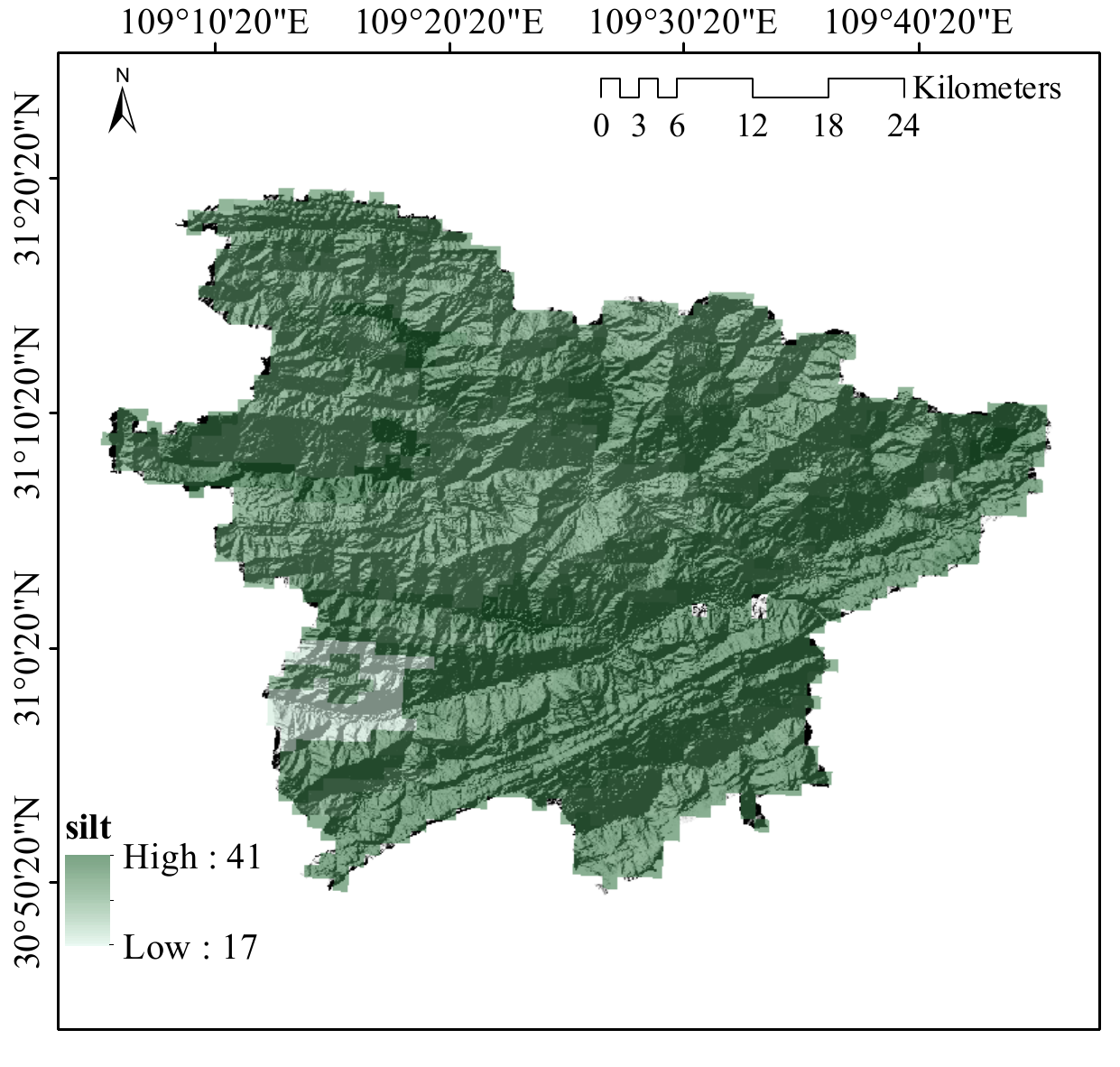}
        \caption{silt coverage}
    \end{subfigure}
    \begin{subfigure}{0.24\textwidth}
        \includegraphics[width=\linewidth]{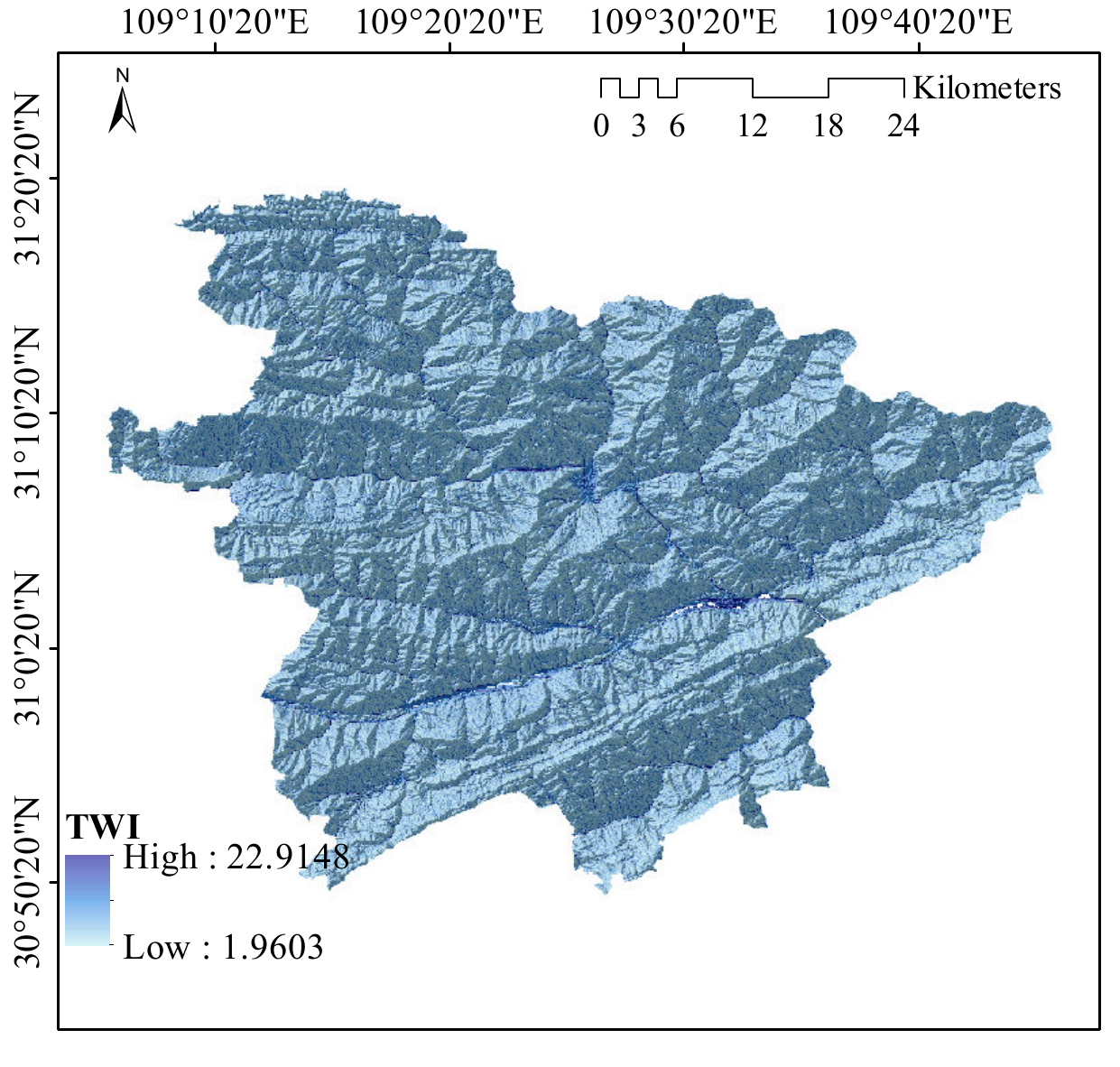}
        \caption{TWI}
    \end{subfigure}
	\begin{subfigure}{0.24\textwidth}
        \includegraphics[width=\linewidth]{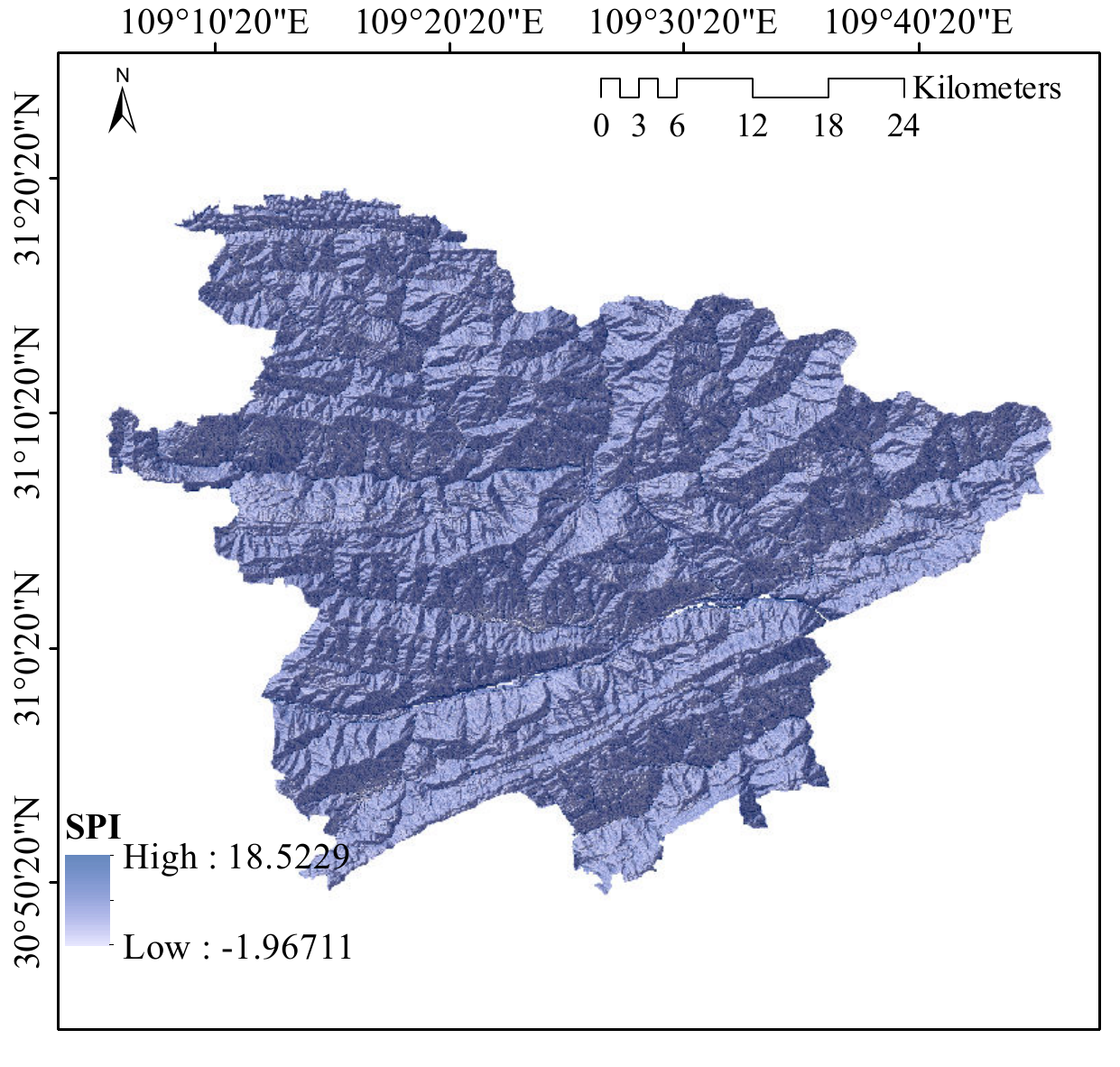}
        \caption{SPI}
    \end{subfigure}
    \begin{subfigure}{0.24\textwidth}
        \includegraphics[width=\linewidth]{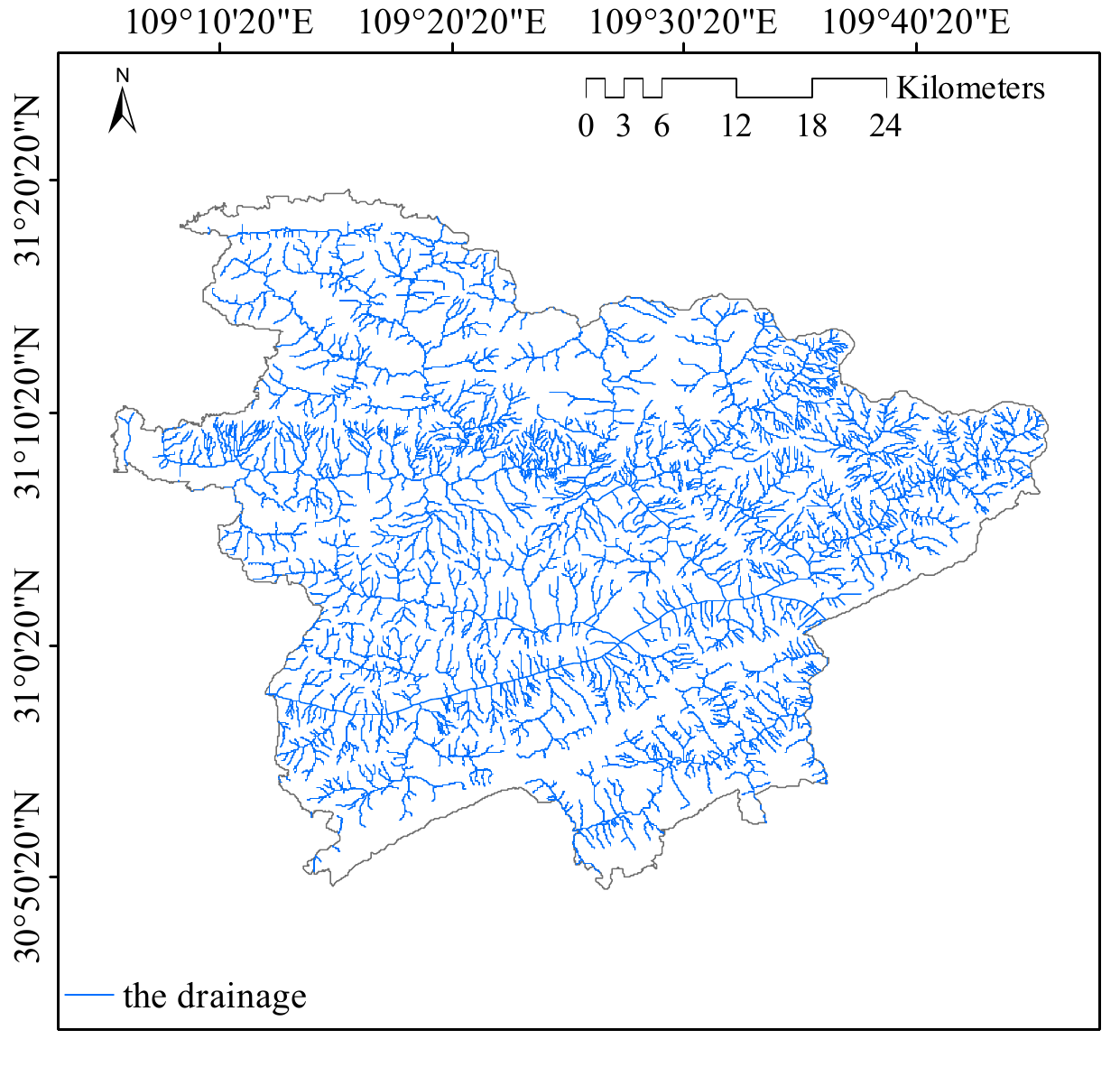}
        \caption{drainage}
    \end{subfigure}
    \caption{Thematic maps of Fengjie County. (a) Land use, (b) landslide occurrence frequency, (c) strata type, (d) DEM, (e) aspect, (f) slope, (g) curvature, (h) NDVI,
    (i) vegetation coverage, (j) soil erodibility, (k)sand coverage, (l) clay coverage, (m) silt coverage, (n) TWI, (o) SPI, and (p) drainage.}
    \label{fig:Thematic maps of FJ}
\end{figure}

\begin{figure}[H]
    \centering
    \begin{subfigure}{0.24\textwidth}
        \includegraphics[width=\linewidth]{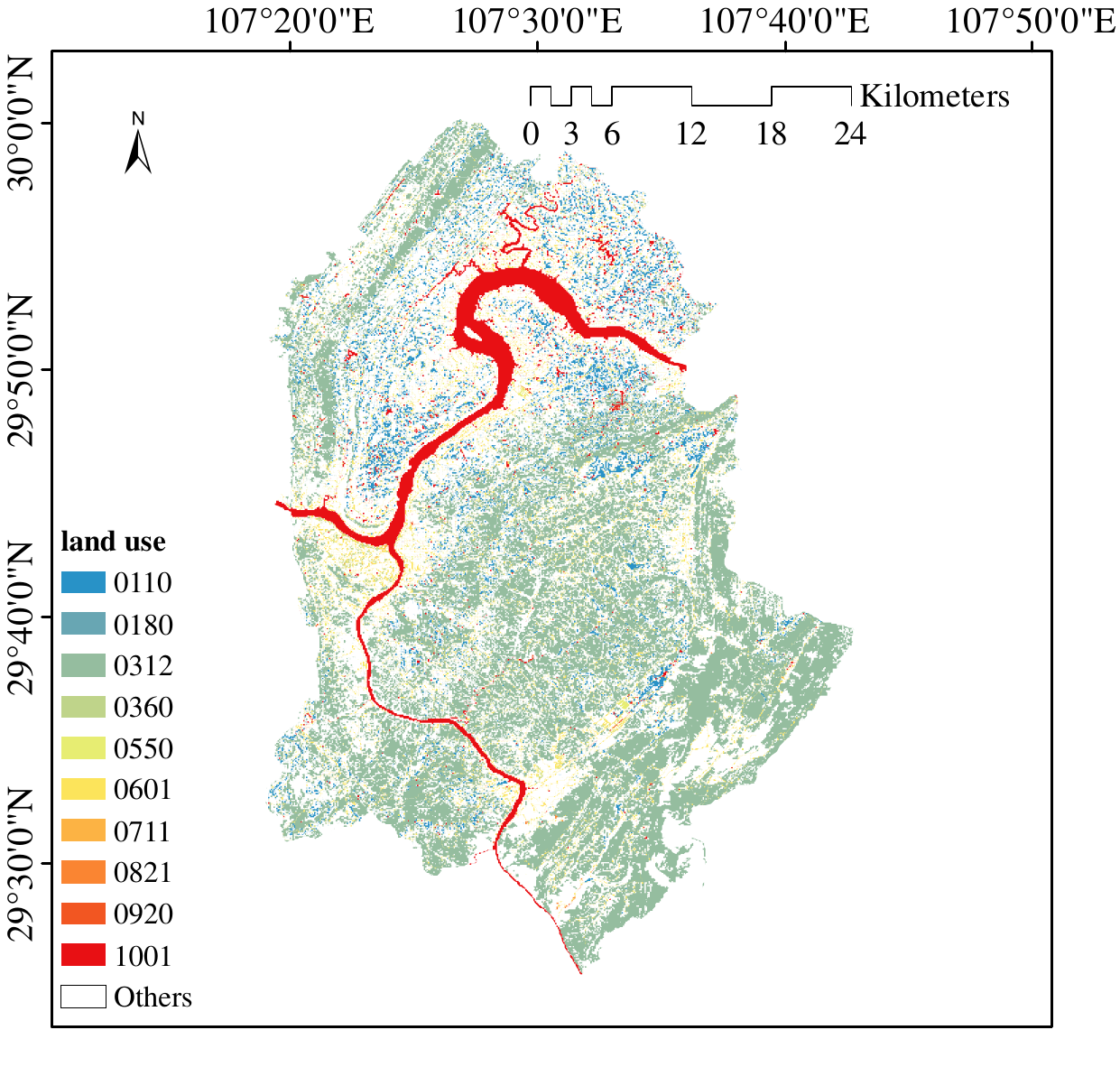}
        \caption{land use}
    \end{subfigure}
    \begin{subfigure}{0.24\textwidth}
        \includegraphics[width=\linewidth]{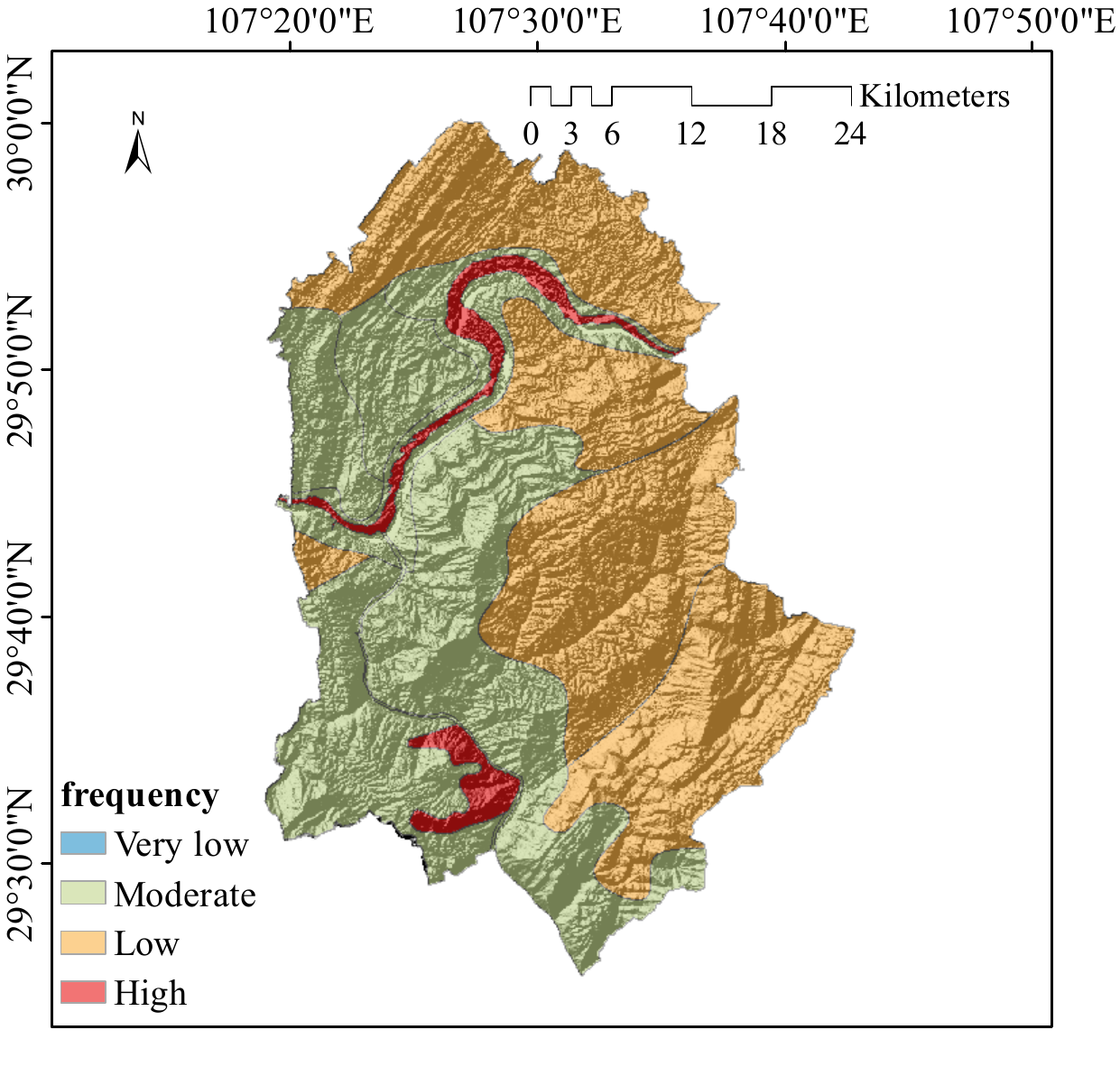}
        \caption{frequency}
    \end{subfigure}
    \begin{subfigure}{0.24\textwidth}
        \includegraphics[width=\linewidth]{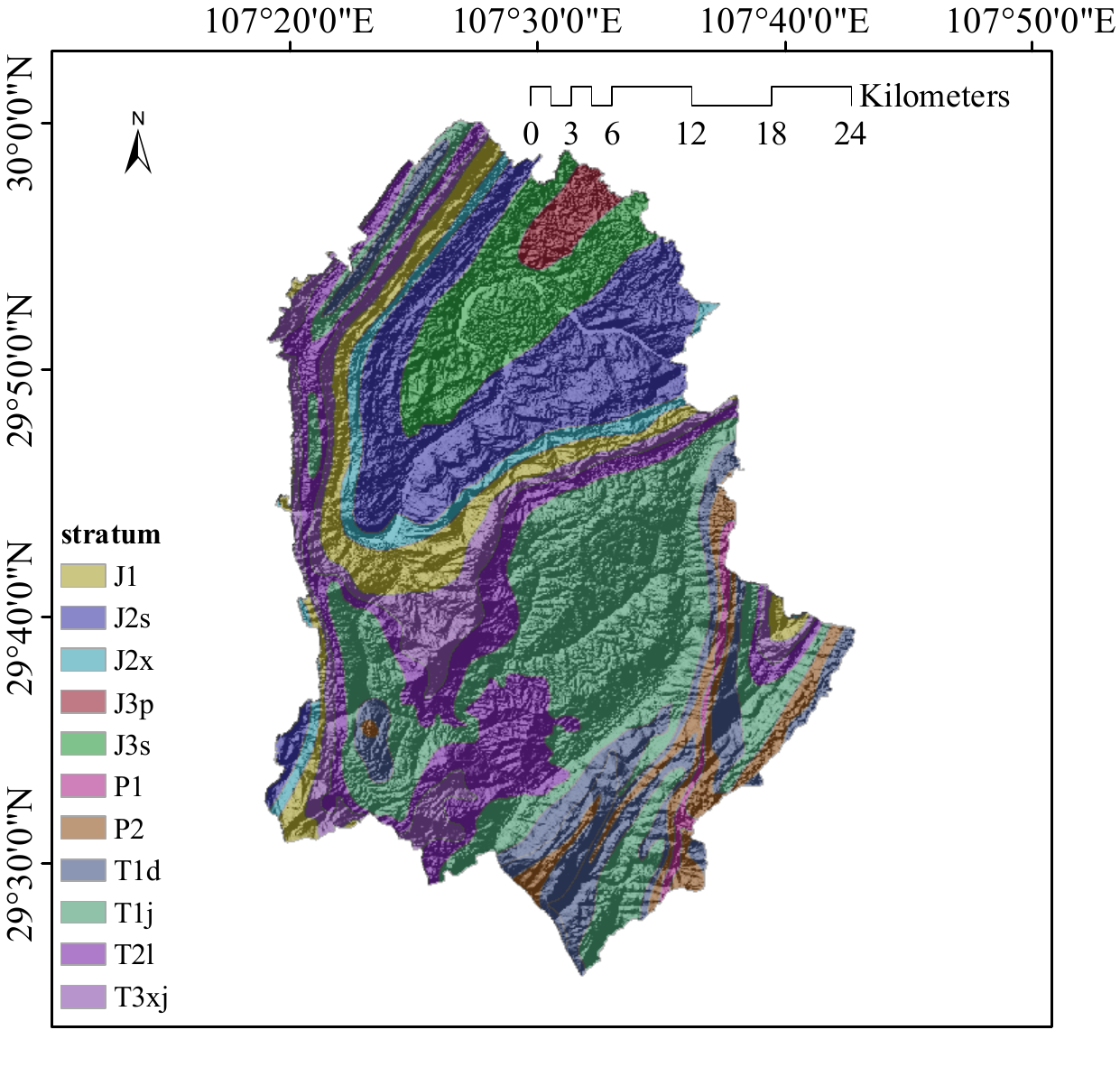}
        \caption{strata type}
    \end{subfigure}
	\begin{subfigure}{0.24\textwidth}
        \includegraphics[width=\linewidth]{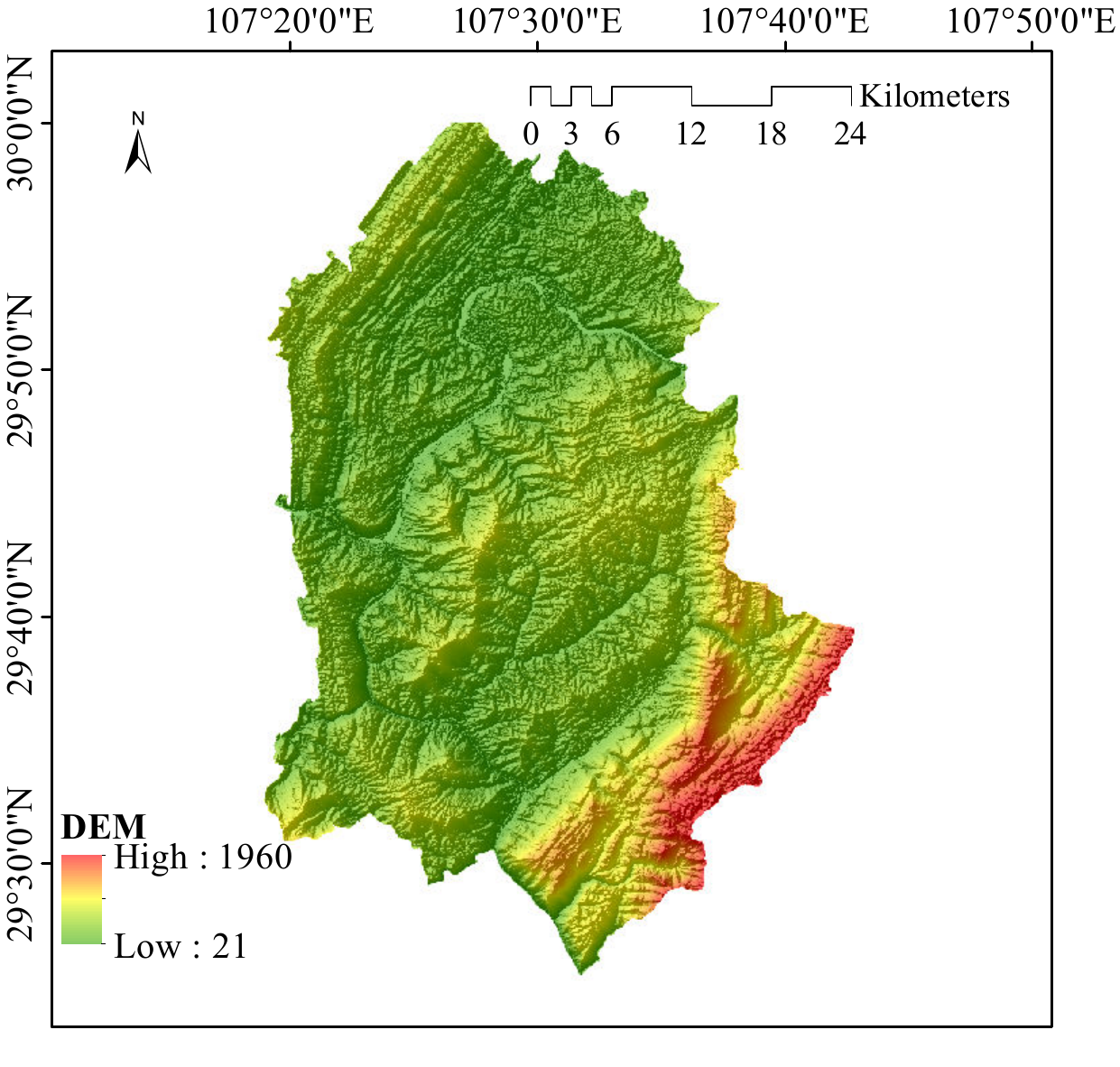}
        \caption{DEM}
    \end{subfigure}
    \begin{subfigure}{0.24\textwidth}
        \includegraphics[width=\linewidth]{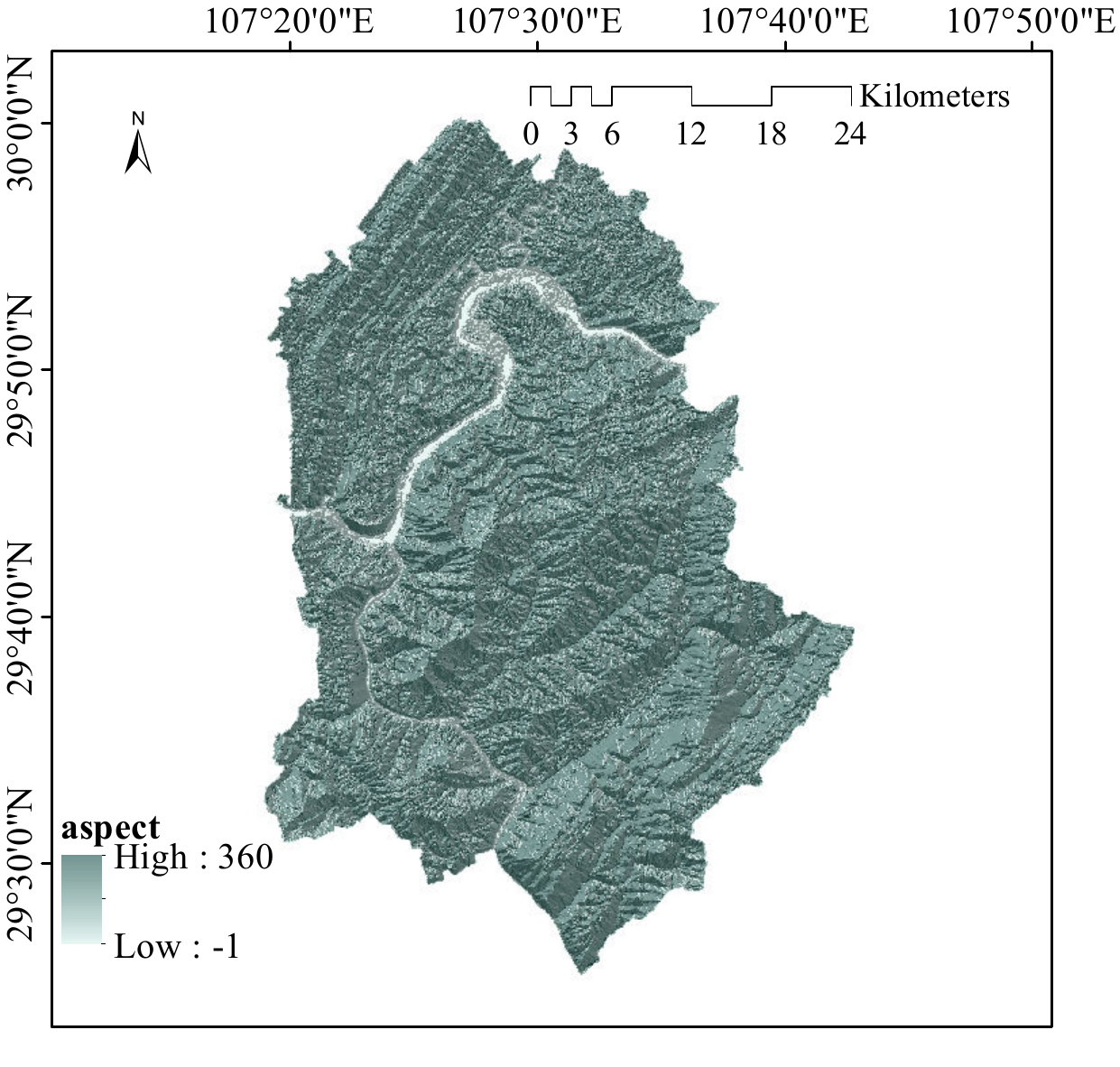}
        \caption{aspect}
    \end{subfigure}
    \begin{subfigure}{0.24\textwidth}
        \includegraphics[width=\linewidth]{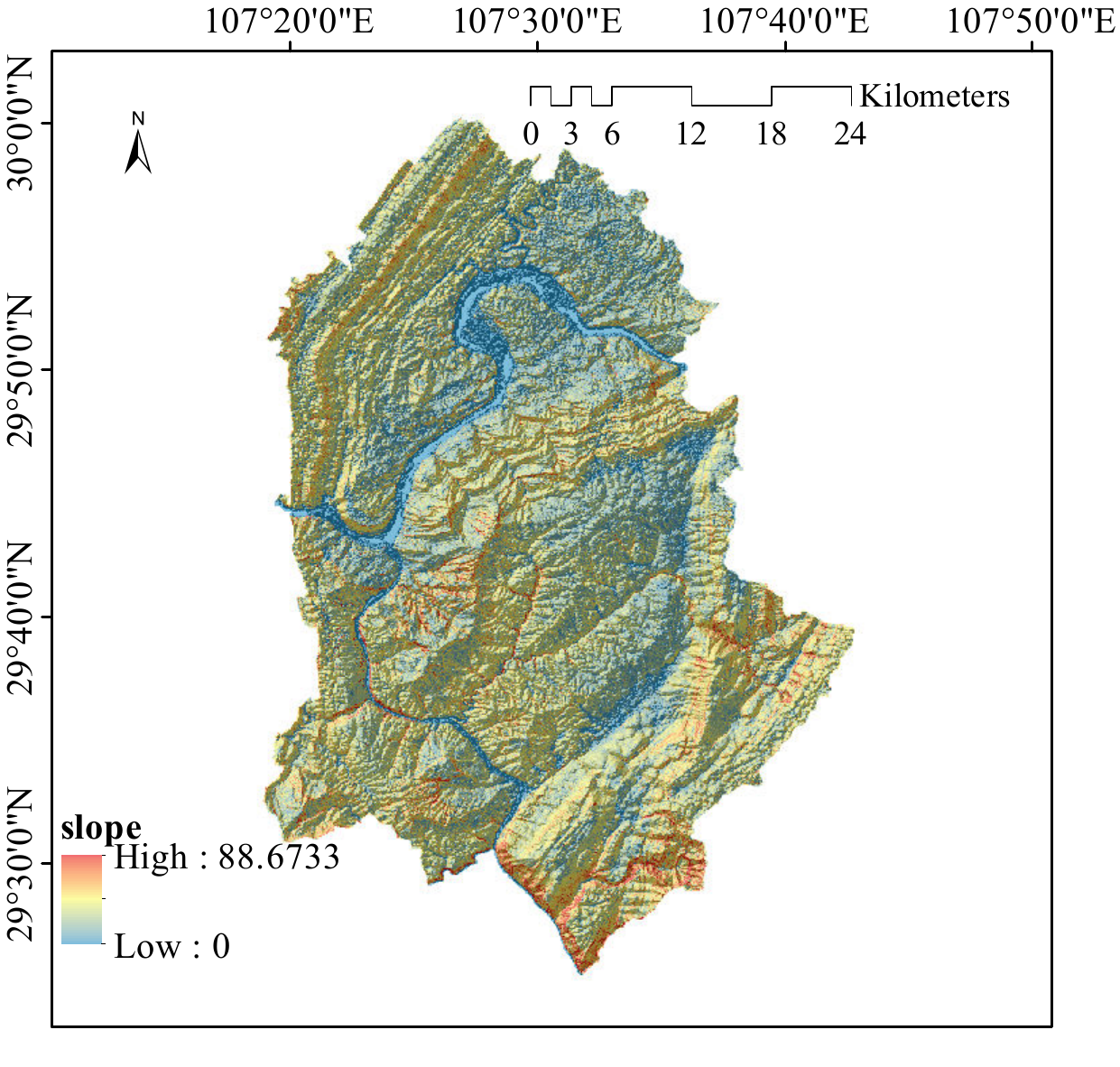}
        \caption{slope}
    \end{subfigure}
    \begin{subfigure}{0.24\textwidth}
        \includegraphics[width=\linewidth]{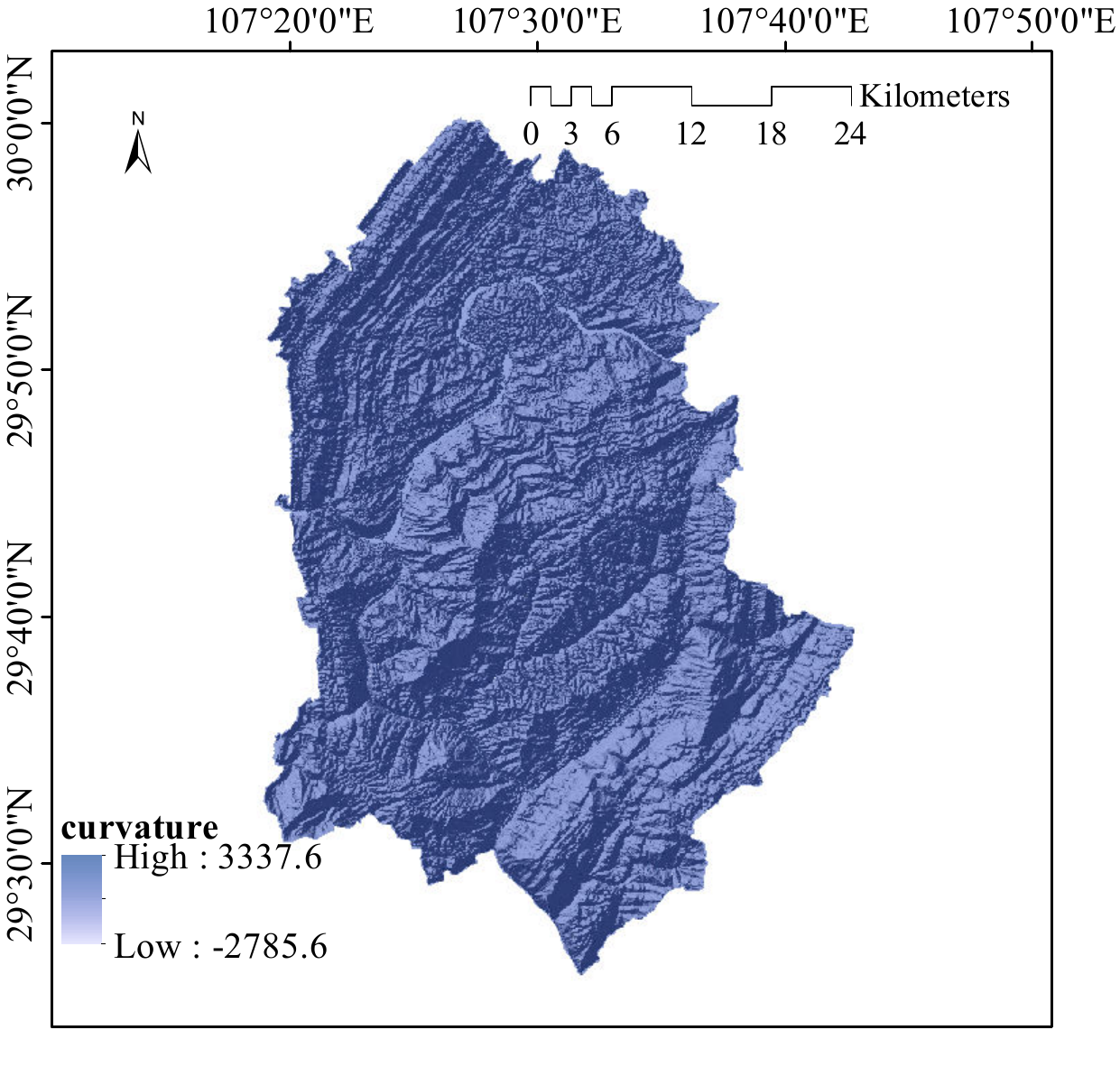}
        \caption{curvature}
    \end{subfigure}
    \begin{subfigure}{0.24\textwidth}
        \includegraphics[width=\linewidth]{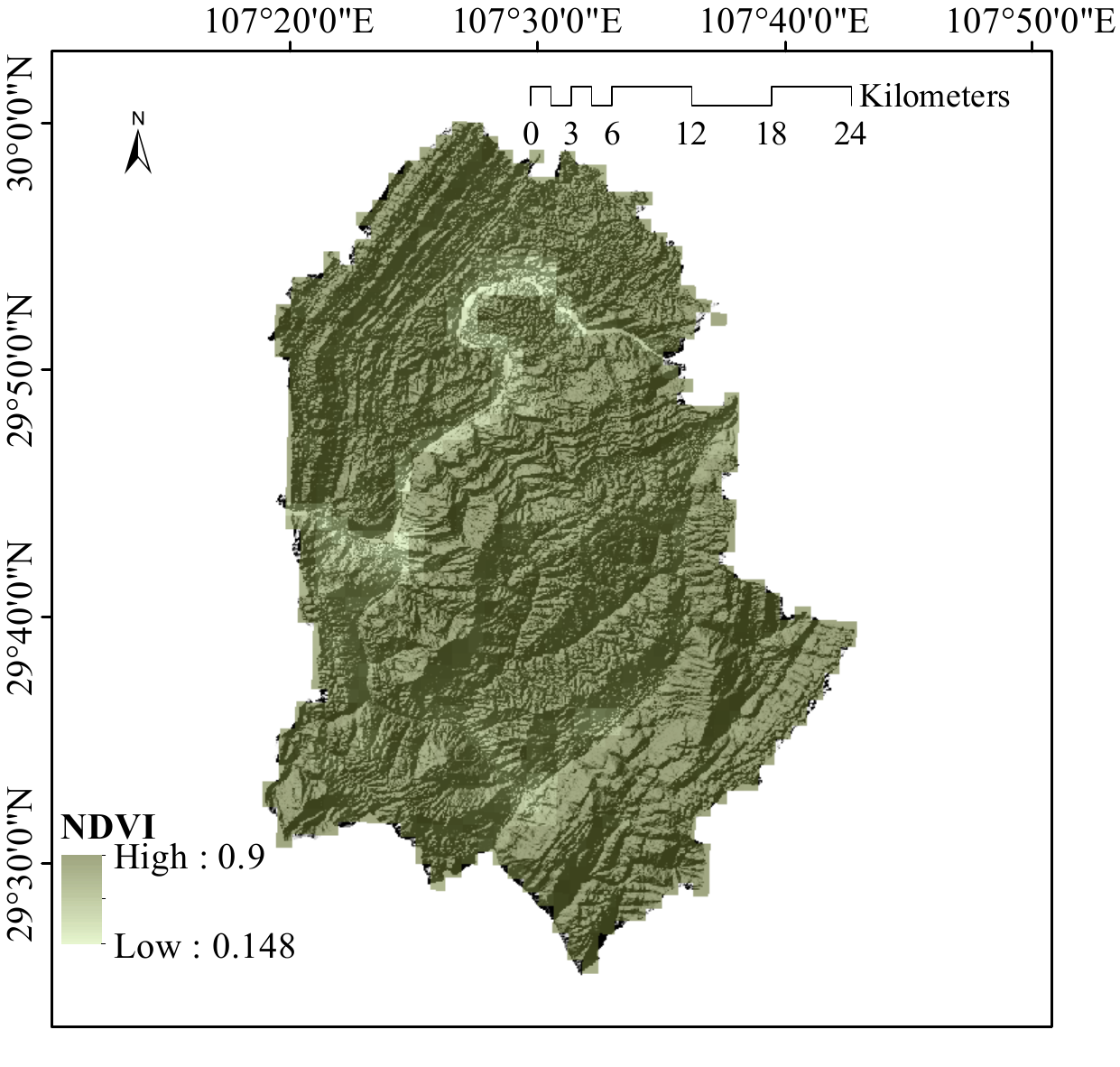}
        \caption{NDVI}
    \end{subfigure}
    \begin{subfigure}{0.24\textwidth}
        \includegraphics[width=\linewidth]{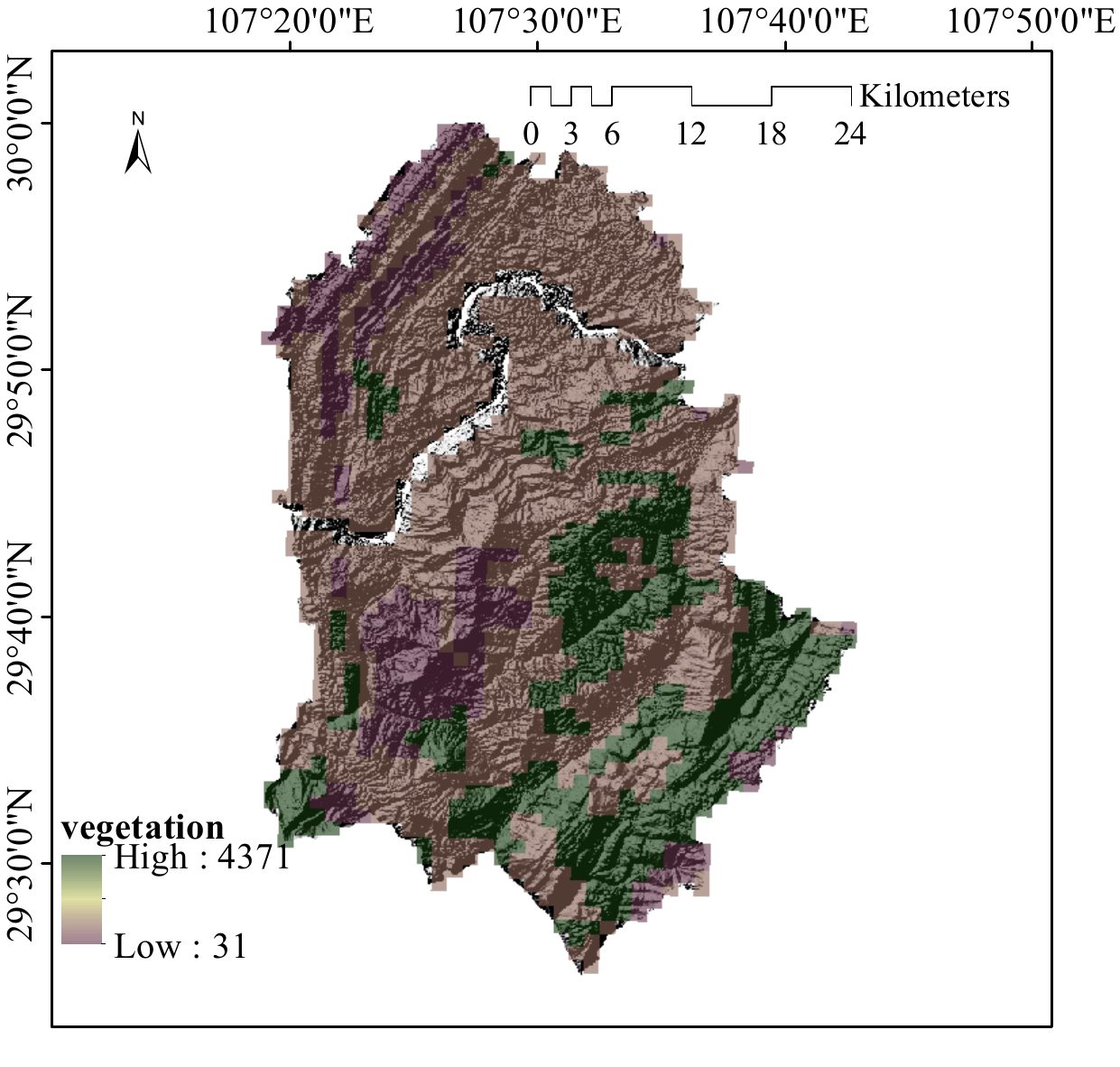}
        \caption{vegetation coverage}
    \end{subfigure}
    \begin{subfigure}{0.24\textwidth}
        \includegraphics[width=\linewidth]{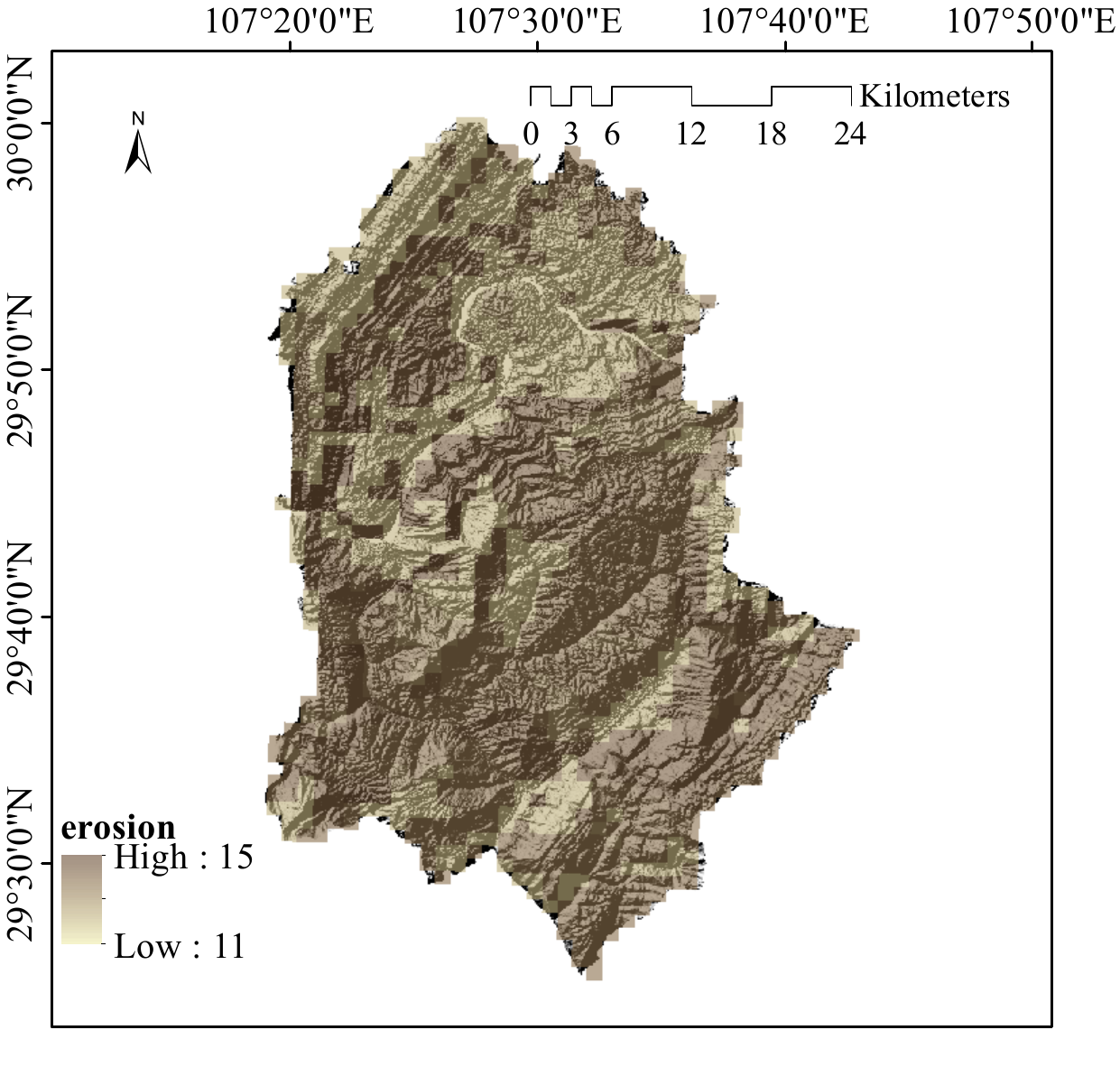}
        \caption{soil erodibility}
    \end{subfigure}
    \begin{subfigure}{0.24\textwidth}
        \includegraphics[width=\linewidth]{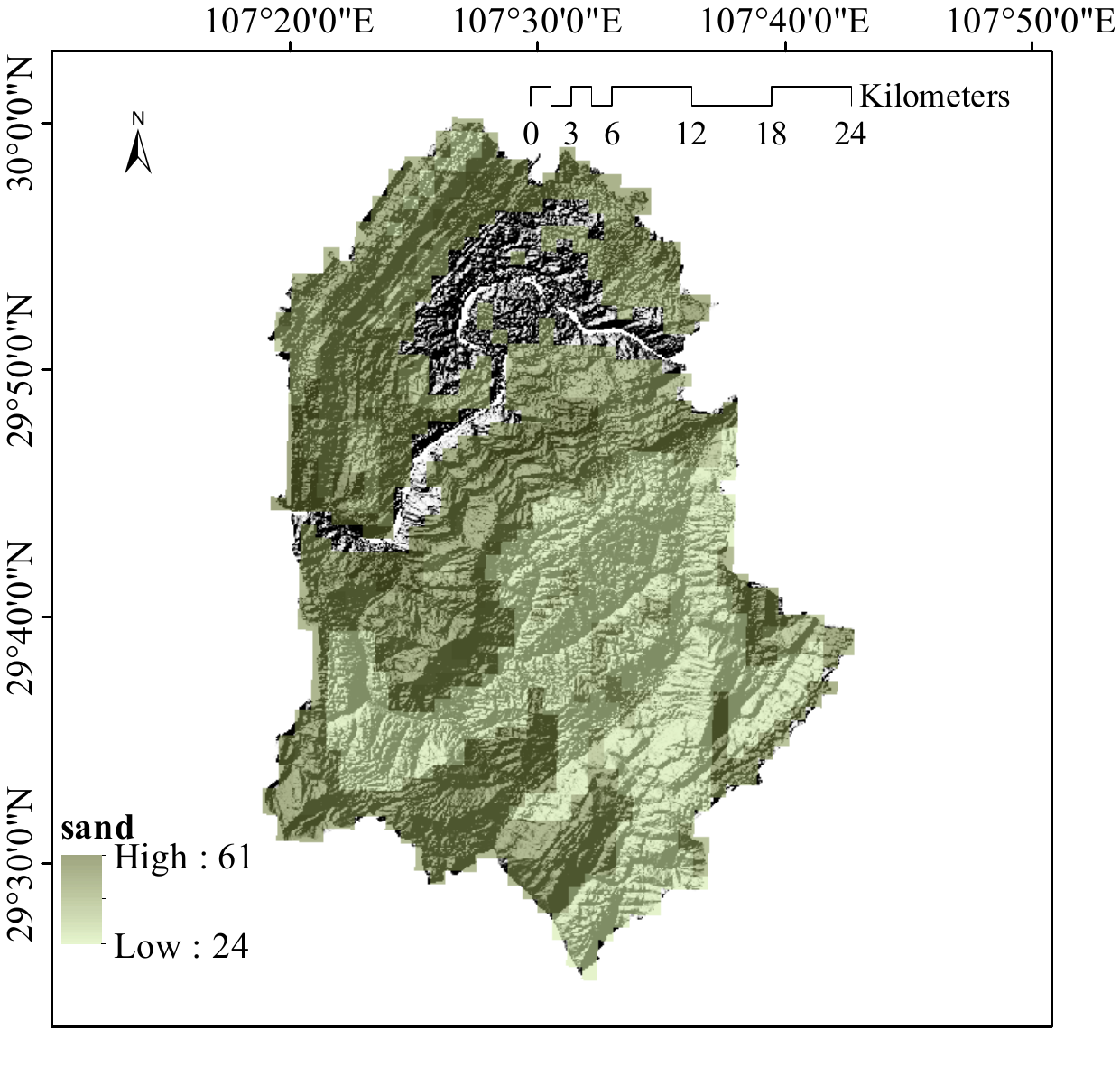}
        \caption{sand coverage}
    \end{subfigure}
    \begin{subfigure}{0.24\textwidth}
        \includegraphics[width=\linewidth]{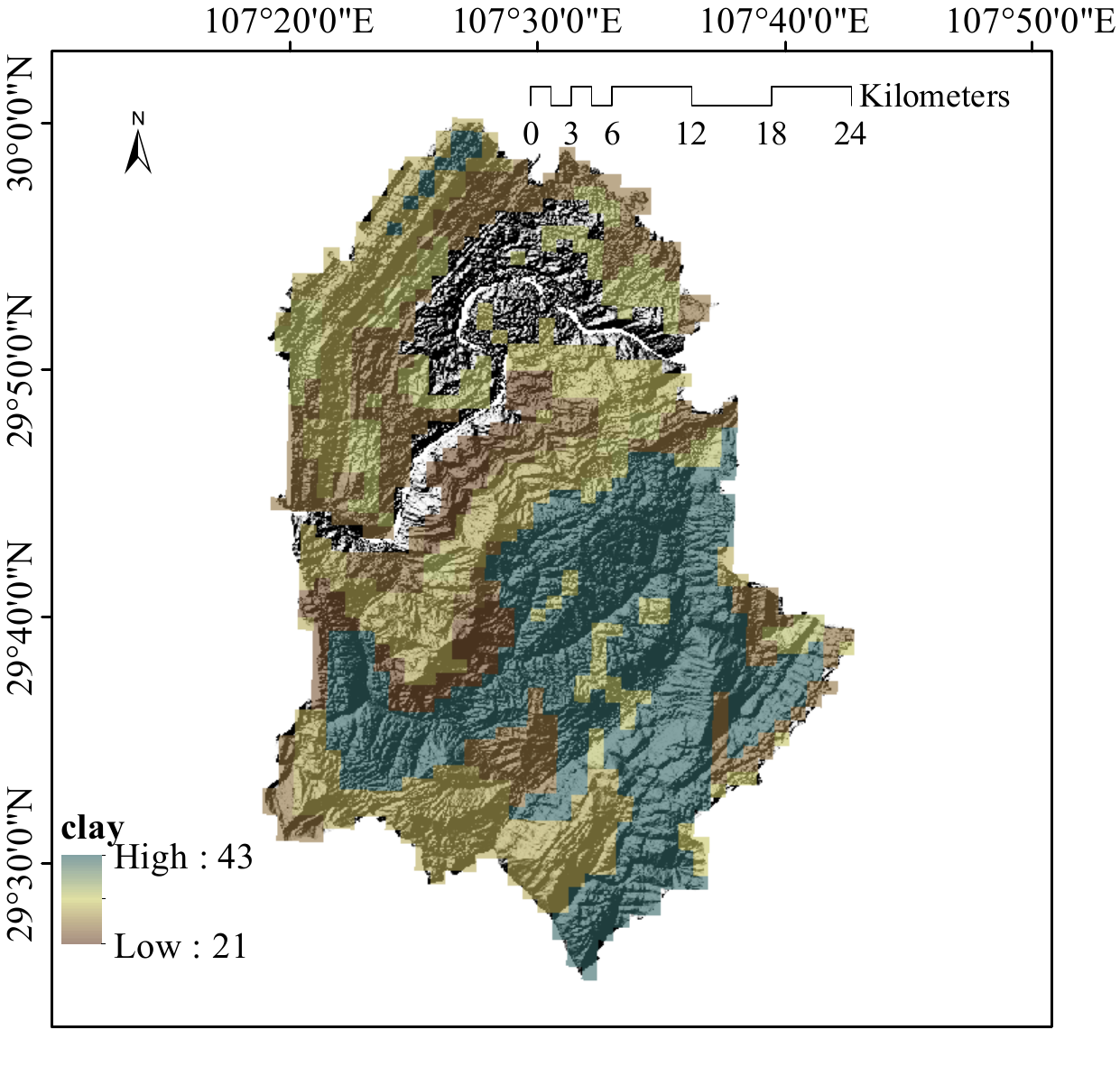}
        \caption{clay coverage}
    \end{subfigure}
    \begin{subfigure}{0.24\textwidth}
        \includegraphics[width=\linewidth]{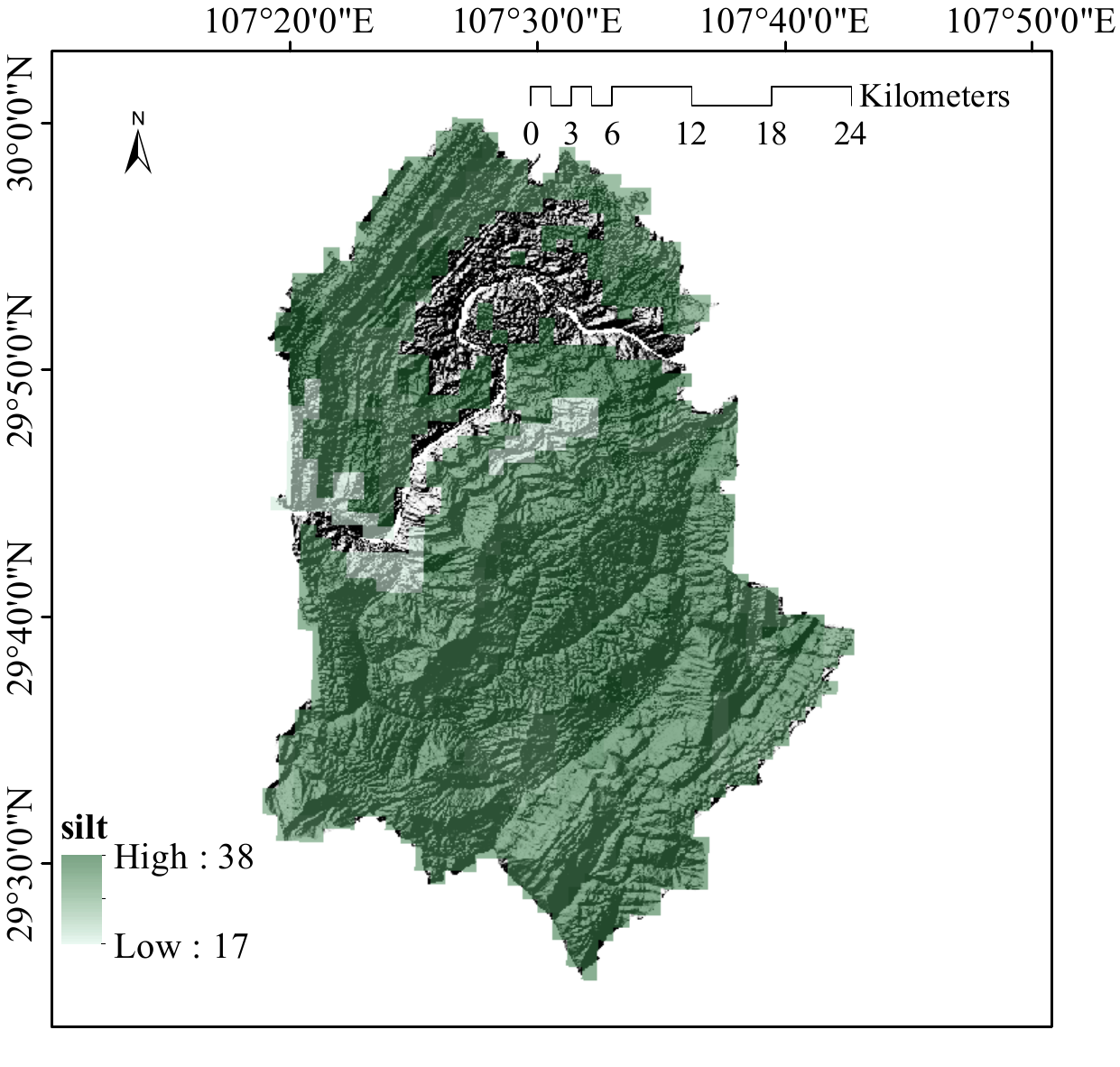}
        \caption{silt coverage}
    \end{subfigure}
    \begin{subfigure}{0.24\textwidth}
        \includegraphics[width=\linewidth]{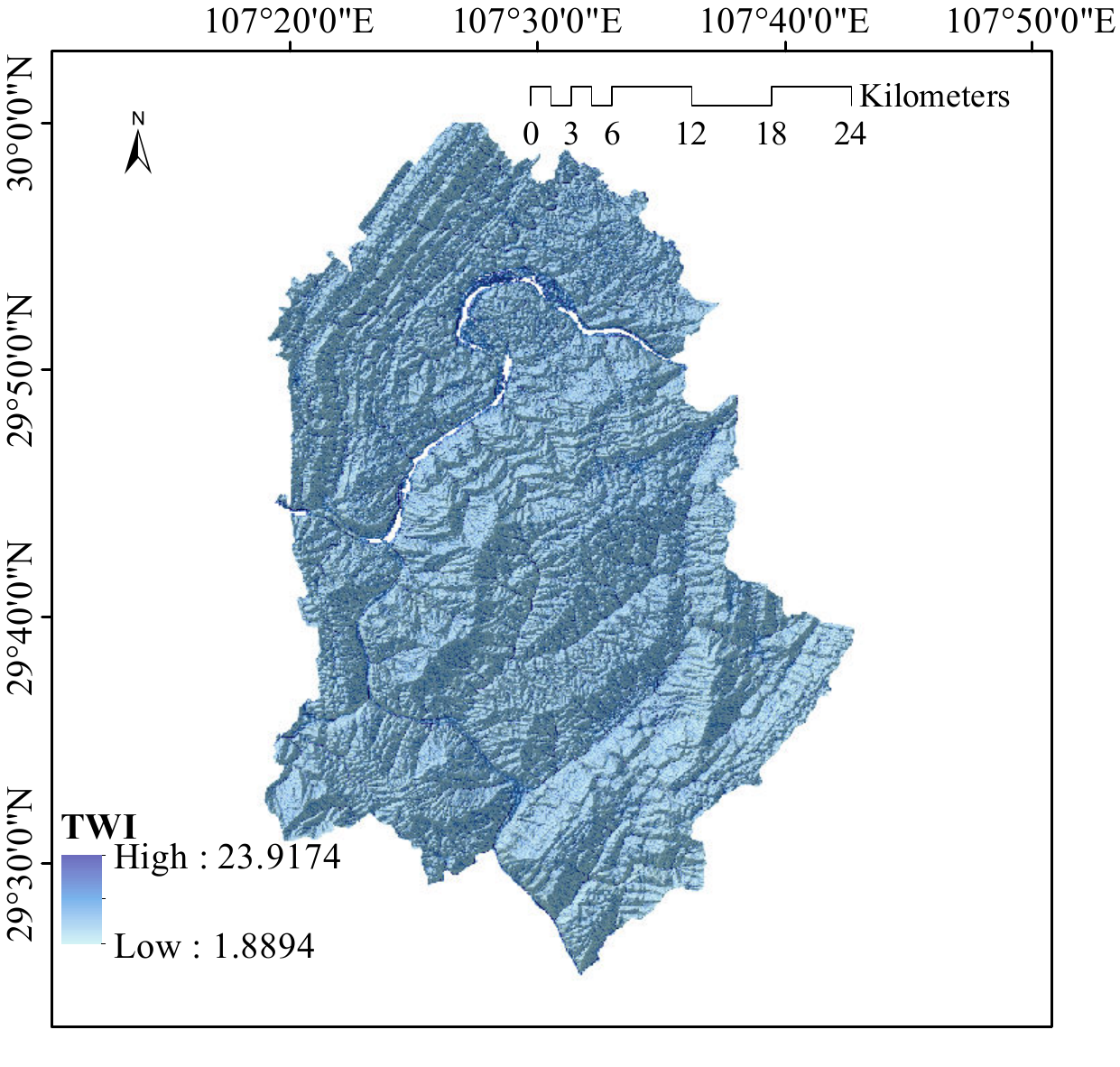}
        \caption{TWI}
    \end{subfigure}
	\begin{subfigure}{0.24\textwidth}
        \includegraphics[width=\linewidth]{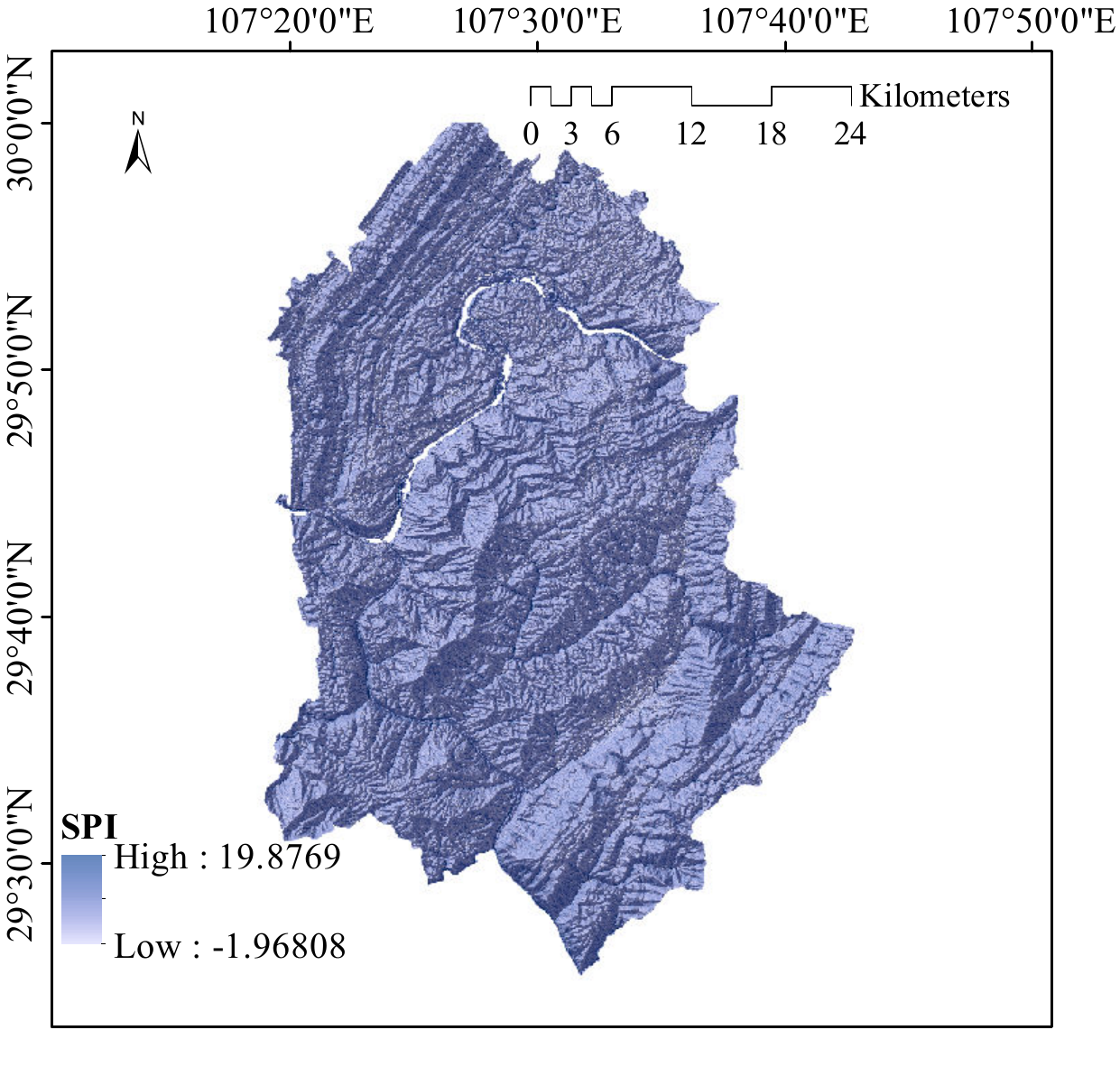}
        \caption{SPI}
    \end{subfigure}
    \begin{subfigure}{0.24\textwidth}
        \includegraphics[width=\linewidth]{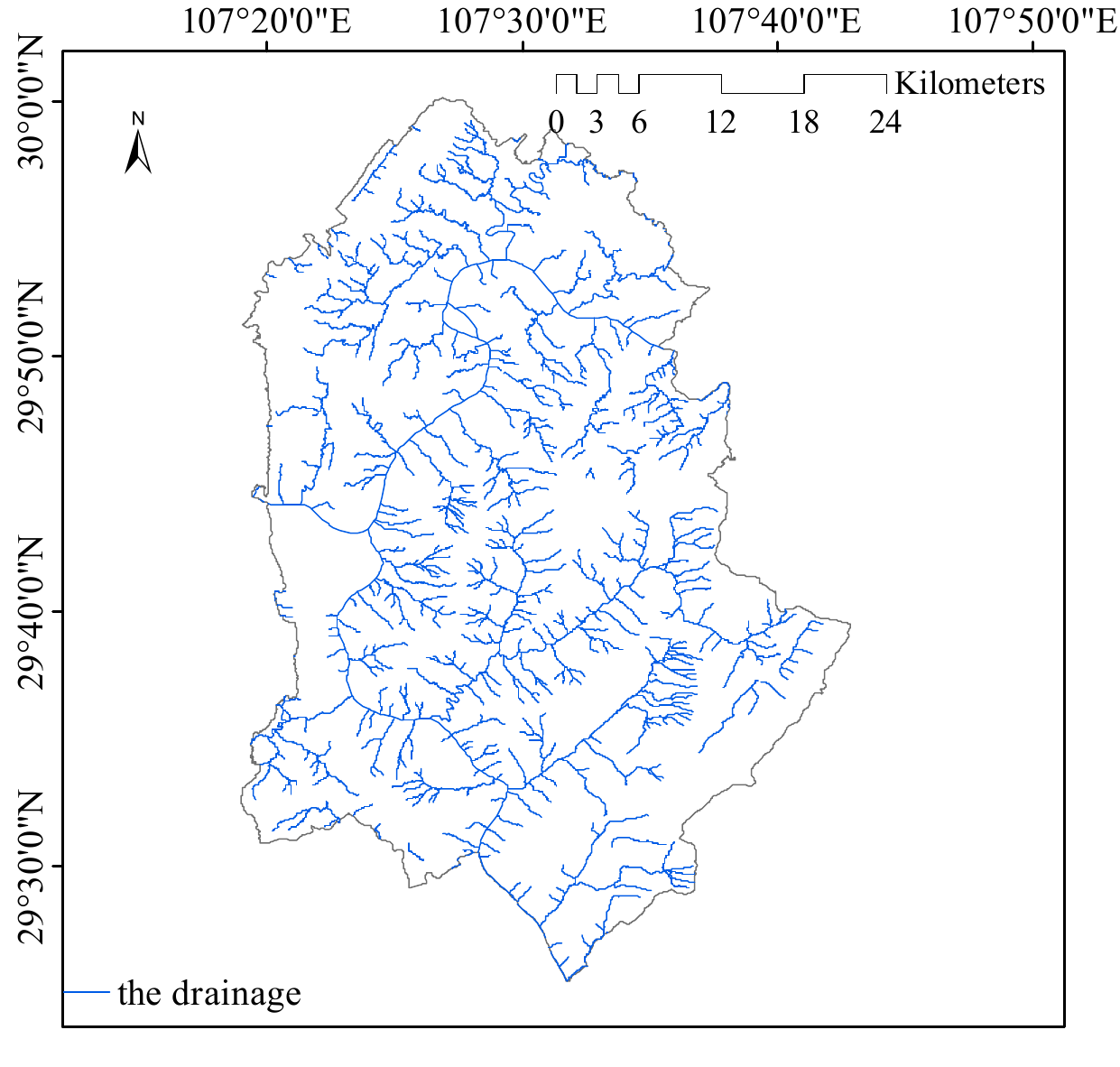}
        \caption{drainage}
    \end{subfigure}
    \caption{Thematic maps of Fuling District. (a) land use, (b) landslide occurrence frequency, (c) strata type, (d) DEM, (e) aspect, (f) slope, (g) curvature, (h) NDVI,
    (i) vegetation coverage, (j) soil erodibility, (k) sand coverage, (l) clay coverage, (m) silt coverage, (n) TWI, (o) SPI, and (p) drainage.}
    \label{fig:Thematic maps of FL}
\end{figure}

The thematic maps related to landslide susceptibility analysis were downloaded from online platforms Computer Network Information Center (CNIC), Chinese Academy of Science, and U.S. Geological Survey (USGS) \citep{cas2021gdc, usgs2021software}.
According to \cite{jebur2014optimization}, the thematic maps used to feature an input sample vector in this study consisted of landslide occurrence frequency, strata type, digital elevation model (DEM), aspect, slope, curvature, normalized difference vegetation index (NDVI), vegetation coverage, soil erodibility, sand coverage, clay coverage, silt coverage, topographic wetness index (TWI), stream power index (SPI), distance to drainage, and distance to the road.
Each thematic map is thought to contain some aspects possibly related to the occurrence of a landslide.
The thematic maps of FJ and FL are shown in Fig. \ref{fig:Thematic maps of FJ} and Fig. \ref{fig:Thematic maps of FL}, respectively, together with the distribution of the landslide locations. 
The sampling method for each landslide record of this article refers to \cite{zhu2020unsupervised}.
It is worth mentioning that different thematic maps have different resolutions, but it does not affect the sampling process.
Our method dealt directly with the original numerical values featured in the thematic maps.
Further, to avoid numerical problems, all sample vector dimensions were normalized \citep{sun2021assessment}.

\section{Proposed method}
\label{s:proposed method}

\subsection{Overview and problem setup}
\begin{figure}[H]
    \centering
    \includegraphics[width=\textwidth]{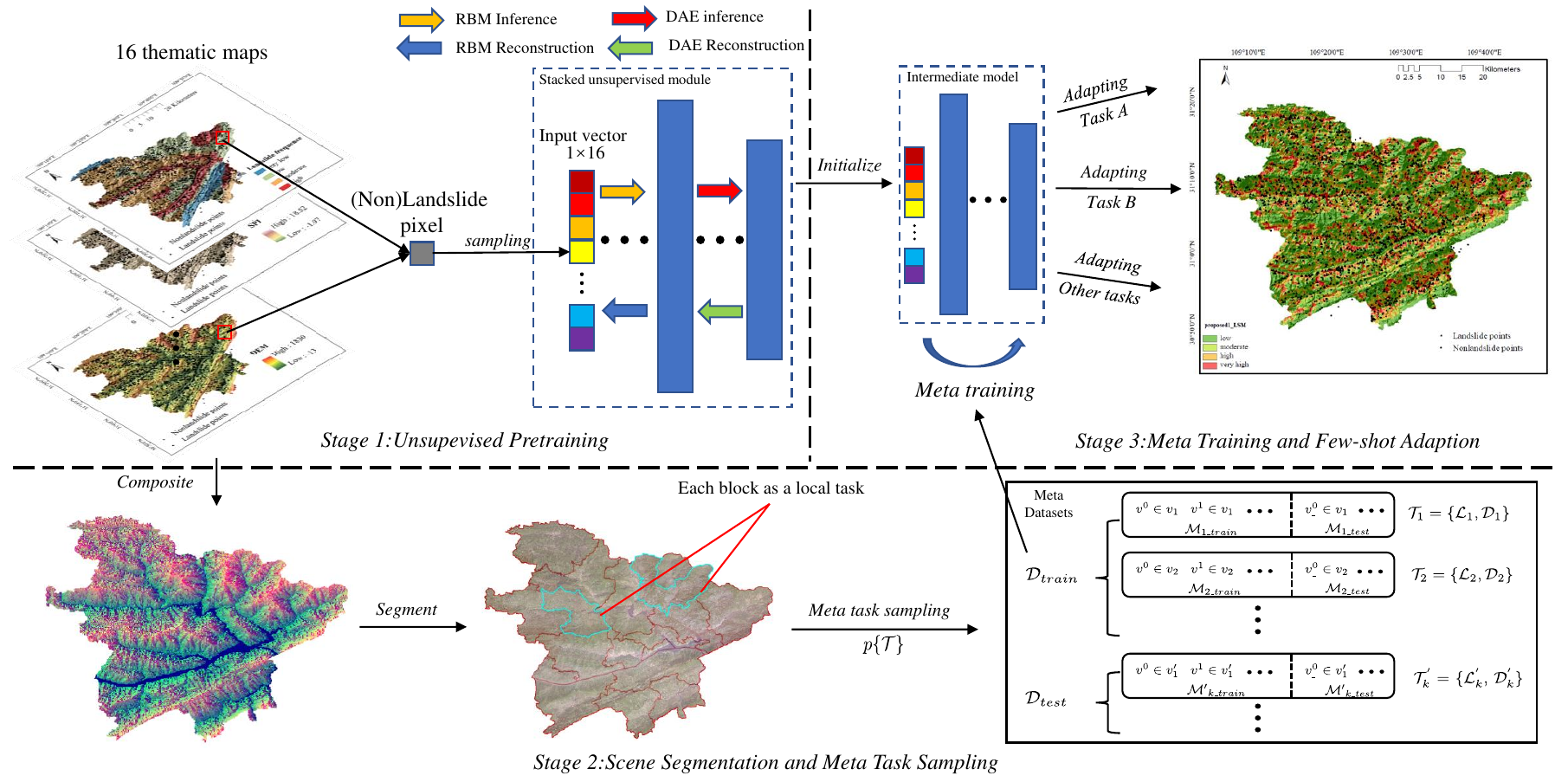}
    \caption{Overview of the proposed method. Our method contains an unsupervised pretraining stage for learning superior representations with good properties; 
    a scene segmentation and task sampling stage for producing meta-datasets; 
     and the meta-training stage for learning intermediate model suited for few-shot adaption of block-wise LSM prediction, which is further explained in Section \ref{subs:meta-learner}.}
    \label{fig:overview}
\end{figure}
\subsubsection{Overview of the methods}
\label{subsubs:overview}
The global prediction of the LSM would lose veracity and even rationality in circumstances where various parts of the target region hold different landslide-inducing environments. 
Thus, we used multiple models to learn the LSM predictive model for different parts of the region. 
Based on this, however, the model of each part of the region would be vulnerable to very few labeled samples. 
To solve this problem, we introduced meta-learning technology for few-shot adaption \citep{ren2018meta}.
An overview of the proposed method is presented in Fig. \ref{fig:overview}, containing three steps: 
(1) \emph{unsupervised pretraining} to transform original data into embedding space with good properties; 
(2) \emph{scene segmentation} for local task sampling to generate meta-datasets; 
(3) and \emph{meta-training} to obtain an intermediate representation that is suited for few-shot adaption to unseen tasks.

\subsubsection{Problem setup}
\label{subsubs:problem setup}
Formally, the initial inputs consist of $M$ thematic maps $\varGamma \{\varGamma_1,...,\varGamma_M\}$ and $N$ landslide (and non-landslide) samples $\mathcal{S}\in\mathbb{Z}^{2 \times N}$ with latitude and longitude information within the World Geodetic System. 
Each sample $s_i$ in $\mathcal{S} \in \mathbb{Z}^{2 \times N}$ features from $\varGamma$ such that sample vectors $\mathcal{V}\in\mathbb{Z}^{M \times N}$ are generated.
Within our framework, $\mathcal{V}$ is first fed as unlabeled samples into the unsupervised pretraining stage to initialize the base model, denoted $f$, which maps input features to output representation $f_{0}$.
After the initialization, we apply SLIC \citep{achanta2012slic} to segment the target scenarios into blocks $\mathcal{B}\{b_1,...,b_K\}$, and then divide the vectors $\mathcal{V}$, labeled as $\mathcal{V}_{l}\{v_1,...,v_K\}$, according to their location in $\mathcal{B}$, thus building meta-task distribution $\mathcal{T}_{k}\{\mathcal{L}_k, \mathcal{D}_k\} \sim p(\mathcal{T})$. 
$\mathcal{L}_k$ is the inner loop loss function, and $\mathcal{D}_k$ is the meta-datasets.
The purpose is then to meta-learn an intermediate model $f'$ that can be adapted to unseen tasks $\mathcal{T'}_{k}$, which is further discussed in Section \ref{subs:meta-learner}.
Next, we fine-tune $f'$ with $v_{i} \in \mathcal{V}_{l}$ to obtain the adapted model $f'_i$ for each block $b_{i} \in \mathcal{B}$.
Finally, we use $f'_i$ to predict the LSM of the region associated with $b_{i} \in \mathcal{B}$.

\subsection{Representation learning for initialization of meta-learner}
\label{subs:unsupervised pretraining of LSM predictive model}
Despite the convenience to be available to online data, several problems remain, for example, (1) data are generally noisy with various resolutions, owing to the high cost of access to high-quality thematic information of large-scale areas; 
(2) inducing factors inferred from thematic maps can be correlated and explain the same aspect of the occurrence of a landslide; 
and (3) unreasonable selection of thematic information would be expected to lead to redundant or confusing conceptions for supervised regression.

The first stage of the proposed method serves as a pretraining process to train the base model of the meta-learner with unlabeled samples for learning the representation $f_0$ featuring robustness, independence, and compactness.
Specifically, the model is composed of sequential stacked unsupervised learning modules, which include two restricted Boltzmann machines (RBMs) \citep{nair2010rectified} and a denoising autoencoder (DAE) \citep{vincent2008extracting}. 
Regarding the RBM, the objective is to minimize the Kullback-Leibler divergence \citep{hinton2002training} between a given sample distribution and the distribution inferred from the model.
The essential problem in the objective is to calculate the two intractable expectations in Eq. \ref{eq:RBM objective},
\begin{equation} 
    \begin{split}
        \frac{\partial{lnP{v^{i}}}}{\partial\theta} &= \mathbb{E}_{P_{data}}[\cdot]+\mathbb{E}_{P_{model}}[\cdot]
    \label{eq:RBM objective}
    \end{split}
    \end{equation}
where $P{v^{i}}$ is the probability of the $i$th sample vector $v^{i} \in \mathcal{V}$, $\mathbb{E}_{P_{data}}[\cdot]$, and $\mathbb{E}_{P_{model}}[\cdot]$ denote data-dependent expectation and model's expectation.
A standard method is to approximate the intractable expectation $\sum _{v}[\cdot]$ using Gibbs sampling \citep{nair2010rectified}, which describes the Markov process carrying out many state transfers between visible and hidden states to approach the stationary distribution represented by the model. 
Nevertheless, a stochastic state still requires too many cycles to reach Gibbs distribution. 
Aiming to improve the learning efficiency, \cite{hinton2002training} proposed Contrastive Divergence (CD-k algorithm), in which data sample is used as initial visible state and proved that by just applying $k$ ($k=1$ usually) step(s) Gibbs sampling, we could sufficiently approach the distribution that the model represents.
Regarding the DAE, the objective is to minimize the reconstruction error between the corrupted input and the reconstructed representation \citep{vincent2008extracting}.


The unsupervised training strategy follows the \textit{greedy layerwise pretraining} rule in \cite{salakhutdinov2009deep} to train unsupervised modules sequentially.
We were based on the premise of RBM that the hidden units are conditionally independent of each other \citep{hinton2007boltzmann}, thus disentangling correlated input units and making the hidden space more discriminative.
We applied DAE to enhance the noise-proof of the model and compress the output representation.
The stacked modules will enhance the representative power of the model to fit complex tasks.
Futhermore, we increased hidden units to generate more independent and descriptive hidden units, which borrows the part of thought in the $1 \times 1$ convolution \citep{lin2013network}. 

\subsection{Scene segmentation and meta-task sampling}
\label{subsubs:scene segmentation and meta-task sampling}
The \textit{global analysis} will build a reasonable predictive model if there are plenty of landslide/non-landslide points within a single scene, where most landslides occurred for the same causes. 
However, in most cases, especially for the large-scale scenario, the causes of a landslide in each part of that scenario are pretty different. 
Accordingly, we segmented the scenario into several parts and modeled them respectively.

\subsubsection{Scene segmentation using combined feature and spatial distance}
\label{subsubs:scene segmentation}
The segmentation approach is based on simple linear iterative clustering (SLIC) \citep{achanta2012slic} for its simplicity and efficiency.
It is superior when processing large-scale areas. 
The collected thematic maps $\varGamma =\{\varGamma_1,...,\varGamma_M\}$ are first clipped with the study areas and then merged into a composition.
Next, according to the input parameters $K$, which denotes the number of output \textit{superpixels} (blocks), the study area is evenly divided into $K$ grids $\mathcal{B}\{b_1,...,b_K\}$ with a size of $S=\sqrt{N/K}$, where $N$ denotes the number of pixels within the study area. 
The cluster center $C_{k}$ of each block $b_K$ is then initialized by the grid center with features $ft_{k}(z_{1},...,z_{M})$ and position ($x_{k}$, $y_{k}$). 
To refine this rough initialization for the subsequent pixel assignment process, we move the cluster centers to the lowest gradient position in a $5 \times 5$ neighborhood, in case the cluster centers lie on the edge of the scene or noisy pixels, which could cause a negative effect when calculating the distance measure $D$, $D$ is given as in Eq. \ref{eq:distance measure}:

\begin{equation}
    \begin{split}
    \begin{aligned}
        d_{f}&=\Vert ft_{j}(w_{1}z_{1},w_{2}z_{2},...)-ft_{i}(w_{1}z_{1},w_{2}z_{2},...) \Vert^{2}, \\
        d_{s}&=\sqrt{(x_{j}-x_{i})^{2}+(y_{j}-y_{i})^{2}}, \\
        D&=\sqrt{d_{f}^{2}+(\frac{d_{s}}{S})^{2}M^{2}},
    \label{eq:distance measure}
    \end{aligned}
    \end{split}
\end{equation}
where $d_{f}$ is the feature distance, $x_{k}$ is the $k$th dimension of $f$, $w_{k}$ is the weighting value of the $i$th dimension, $d_{s}$ is the spatial distance, and $M$ weights the relative importance between feature similarity and spatial proximity. 
Notice that inducing factors could exert different levels of contribution to the occurrence of a landslide. Thus, each considered feature $x_{m}$ is weighted with $w_{m}$ based on \citep{popescu2002landslide} in this study.

In the pixel assignment process, for each pixel $P_{i}$, the corresponding sample vector $v_i\in\mathcal{V}_{l}$ is located in, $l_{i}$ is initialized as -1 and $d_{i}$ as $\infty$, where $l_{i}$ denotes which block $b_{i}$ is the pixel belonging to, and $d_{i}$ denotes the closest distance from a \textit{superpixel} center $C_{k}$ to that pixel. 
Next, we search in a $2S \times 2S$ region around the cluster center $C_{k'}$ to identify the pixels that have the closest distance $D_{min}$ to that center compared to others. 
Then, we assign $l_{i}$ with $k'$, $d_{i}$ with $D_{min}$.
Once the assignment is completed, labeled sample vectors are grouped into $\mathcal{V}_{l}\{v_1,...,v_K\}$ by $l_{i}$.

\subsubsection{Meta-task generation}
\label{meta-task sampling}
After the scene segmentation process, $\mathcal{B}\{b_1,...,b_K\}$ is used for meta-task sampling $\mathcal{T}_{k}\sim p(\mathcal{T})$. 
Specifically, each segmented \textit{superpixel} is simply regarded as a task scenario $\mathcal{T}_{k}=\{\mathcal{L}_{k}, \mathcal{D}_{k}\}$, where $\mathcal{L}_{k}$ is the inner loop loss function and $\mathcal{D}_{k}$ is the collection of sample vectors $v_{i} \in \mathcal{V}_{label}$ of $\mathcal{T}_{k}$. 
From these meta-datasets, we randomly choose 60\% of $\mathcal{D}$ as the training set $\mathcal{D}_{train}$ for meta-training, and the other as the testing set $\mathcal{D}_{test}$ for validation. 
For each task $\mathcal{T}_{k}$ in $\mathcal{D}_{train}$, $\mathcal{D}_{k}$ is divided into a training part $\mathcal{M}_{k\_train}$ (also known as the support set), and a testing part $\mathcal{M}_{k\_test}$ (also known as the query set) serving separately for the inner and outer loop optimization. 
The meta-datasets are organized as in Fig. \ref{fig:overview}. 
$\mathcal{D}_{k}$ has only a few samples.


\subsection{The meta-learner}
\label{subs:meta-learner}
\begin{figure}[H]
    \centering
    \includegraphics[width=0.75\textwidth]{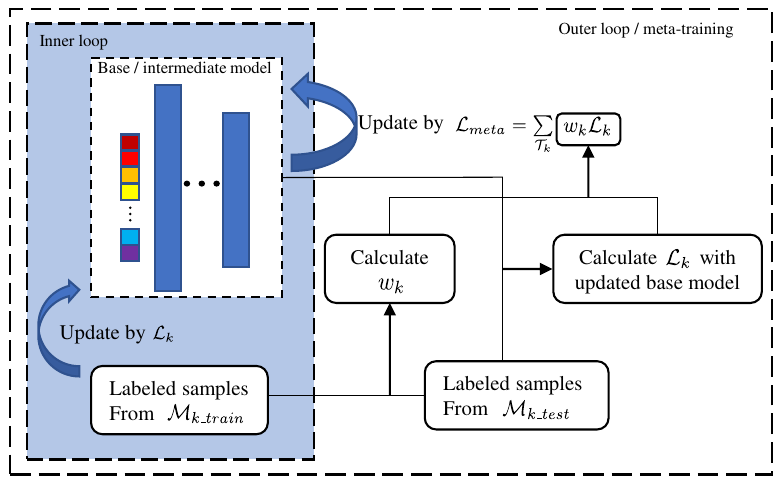}
    \caption{The meta-learner.}
    \label{fig:Meta learner}
\end{figure}

In particular, utilizing only one predictive model can lose veracity when large-scale scenarios consist of districts with various landslide-inducing environments. 
In addition, the traditional data-driven method is vulnerable to the deficiency of high-quality thematic information and historical landslide records because of the vast costs of data collection. 
As a resolution, the meta-learner, shown in Fig. \ref{fig:Meta learner}, is designed for learning a general intermediate model for few shot adaption.
The meta-learning strategy consists of two levels of loops of computation \citep{goldblum2020unraveling}, an inner loop for the optimization process, and an outer loop for optimizing some aspects of the inner loop, for example, the hyperparameters or even the optimization manner. 

Formally, concerning the inner loop, suppose that we have constructed a model parameterized with $\theta$ that is initialized by $f_0$, pre-trained in Section \ref{subs:unsupervised pretraining of LSM predictive model}, we use $\mathcal{M}_{k\_train}$ in $\mathcal{D}_{k}$ to train the model with few gradient descent updates by $\mathcal{L}_{k}$. 
Eq. \ref{eq:inner update} gives the optimization,

\begin{equation}
\begin{split}
  \theta_{k}^{'}=\theta-\alpha\nabla_{\theta}\mathcal{L}_{k},
  \label{eq:inner update}
\end{split}
\end{equation}
where $\theta_{k}^{'}$ denotes the updated parameter, which can be represented by $\theta$, $\alpha$ denotes the inner learning rate, and we choose cross entropy as $\mathcal{L}_{k}$, as given in Eq. \ref{eq:cross entropy},

\begin{equation}
\begin{split}
  \mathcal{L}_{k}=\frac{1}{N}\sum\limits_{k}-[y_{k} \cdot log(p_{k})+(1-y_{k}) \cdot log(1-p_{k})],
  \label{eq:cross entropy}
\end{split}
\end{equation}
where $y_{k}$ denotes the label of the $k$th sample, $p_{k}$ represents the probability that the $k$th sample is predicted to be positive.

With such few samples and training steps, the meta-objective is then to minimize the sum of test errors of tasks in $\mathcal{D}_{train}$, which urges the model to excavate the conception sensitive to task variations so that a few samples and iterations could make large improvements to new task adaption. 
For this study, we consider the circumstances of inaccurate supervision, where the given landslide labels are not always the ground truth, which could cause highly adverse effects on meta-learning because only very few samples are used in the inner optimization or model fine-tuning \citep{hoffmann2011knowledge}. 
In other words, the losses computed from wrong labels could account for a large proportion of the task loss $\mathcal{L}_{k}$, thus misleading the gradient in the descending direction. 
To alleviate the adverse effect, we introduce a method of weighting $\mathcal{L}_{k}$ to restrain tasks with very few samples, assuming that it is more likely to neutralize the wrong loss in a task with relatively more samples. 
Accordingly, the meta-objective $\mathcal{L}_{meta}$ is given by Eq. \ref{eq:meta-objective},

\begin{equation}
\begin{split}
  \mathcal{L}_{meta}=min\sum\limits_{\mathcal{T}_{k} \sim p(\mathcal{T})}w_{k}\mathcal{L}_{k},
  \label{eq:meta-objective}
\end{split}
\end{equation}
where $w_{k}$ is the weight of $\mathcal{T}_{k}$. This promotes the stability and reliability of training an intermediate model, which is further discussed in Section \ref{subs:analysis of the meta-learner}. 
In this study, during the iteration when generating the task batch, $w_{k}$ of task $\mathcal{L}_{k}$ is given by Eq. \ref{eq:weighting},

\begin{equation}
\begin{split}
  w_{k}=\frac{exp(n^{k}/n_{total})}{\sum\limits_{i}exp(n^{i}/n_{total})},
  \label{eq:weighting}
\end{split}
\end{equation}
where $n_{total}$ is the number of total samples in FJ and FL, $n^{k}$ is the number of samples in $\mathcal{T}_{k}$, and $i$ denotes the index of the task in a meta-training batch. 

\subsection{Few-shot adaption and block-wise LSM prediction}
Once the intermediate model $f'$ is meta-trained, the few-shot adaption or validation process for $\mathcal{T}_{k}$ and $\mathcal{T'}_{k}$ simply repeats the optimization in Eq. \ref{eq:inner update}, using sample vectors $\mathcal{M}_{k\_train}$ and $\mathcal{M'}_{k\_train}$, and with very few iterations.
Thus, we obtain adapted model $f'_i$ for each block $b_i \in \mathcal{B}$.
We then rasterize the FJ and FL areas and feature grids within each block $b_i$ into $g_i \in G\{g_1,...,g_K\}$ by means introduced in Section \ref{s:study areas and datasets}. 
Next, for each block $b_i$, we match corresponding predictive model $f'_i$ to predict the landslide susceptibility of $g_i$.
Finally, we integrate $G\{g_1,...,g_K\}$ and output the landslide susceptibility map.
The intermediate model $f'$ is meta-trained together with FJ and FL tasks, such that more general concepts can be learned from various landslide-inducing environments.

\subsection{Implement details}
\label{subs:implementation details}
The proposed method was implemented based on the tensorflow framework (version Tensorflow-gpu 1.9.0). 
In the stage of initializing the meta learner (unsupervised pretraining stage), RBMs and DAE were trained sequentially by the \textit{Adam} optimizer from the bottom to top, with learning rates of 1e-3 and 1e-5, respectively. 
Both training epochs were set to 20. 
In the scene segmentation stage, limited to the number of samples, $K$ of FL and FJ were set to 196 and 64, respectively.
The iteration loop of the pixel assignment process was set to 5. 
In the meta-learning stage, we initialized the inner learning rate $\alpha$ as 1e-1, which was relatively large and suited for quick adaption. 
The numbers of iterations of the inner optimization and few-shot adaption were both set to 5. 
The epoch of training meta-objective was set to 5000 with a learning rate of 1e-4.
The code and data were made publicly available on the website\footnote{\url{https://github.com/Young-Excavator/Meta_LSM}}.

\section{Results and discussion}
\label{s:experiments and analysis}
In this section, we first predict landslide susceptibility and evaluate the performance. 
Then, we analyze the effect of unsupervised pretraining and how our meta-learner weakens the adverse effects of inaccurate supervision and realizes faster adaption to new unseen \textit{local tasks}.

\subsection{Evaluation and comparison}
\label{subs:evaluation and comparison}

\subsubsection{LSM predction results}
\label{subs:LSM prediction results}
\begin{figure}[H]
    \centering
    \begin{subfigure}{0.32\textwidth}
        \includegraphics[width=\linewidth]{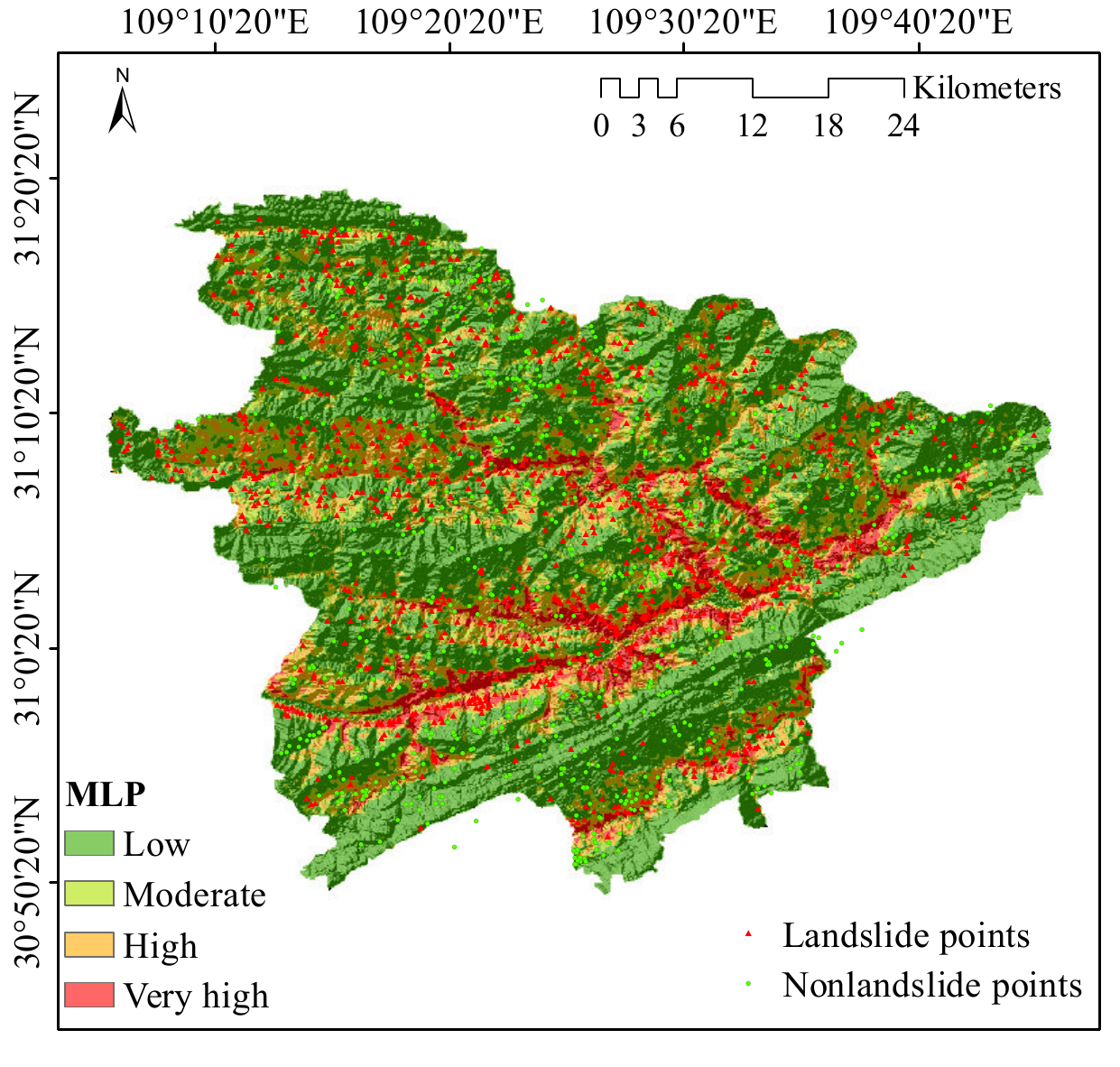}
        \caption{MLP-based}
    \end{subfigure}
    \begin{subfigure}{0.32\textwidth}
        \includegraphics[width=\linewidth]{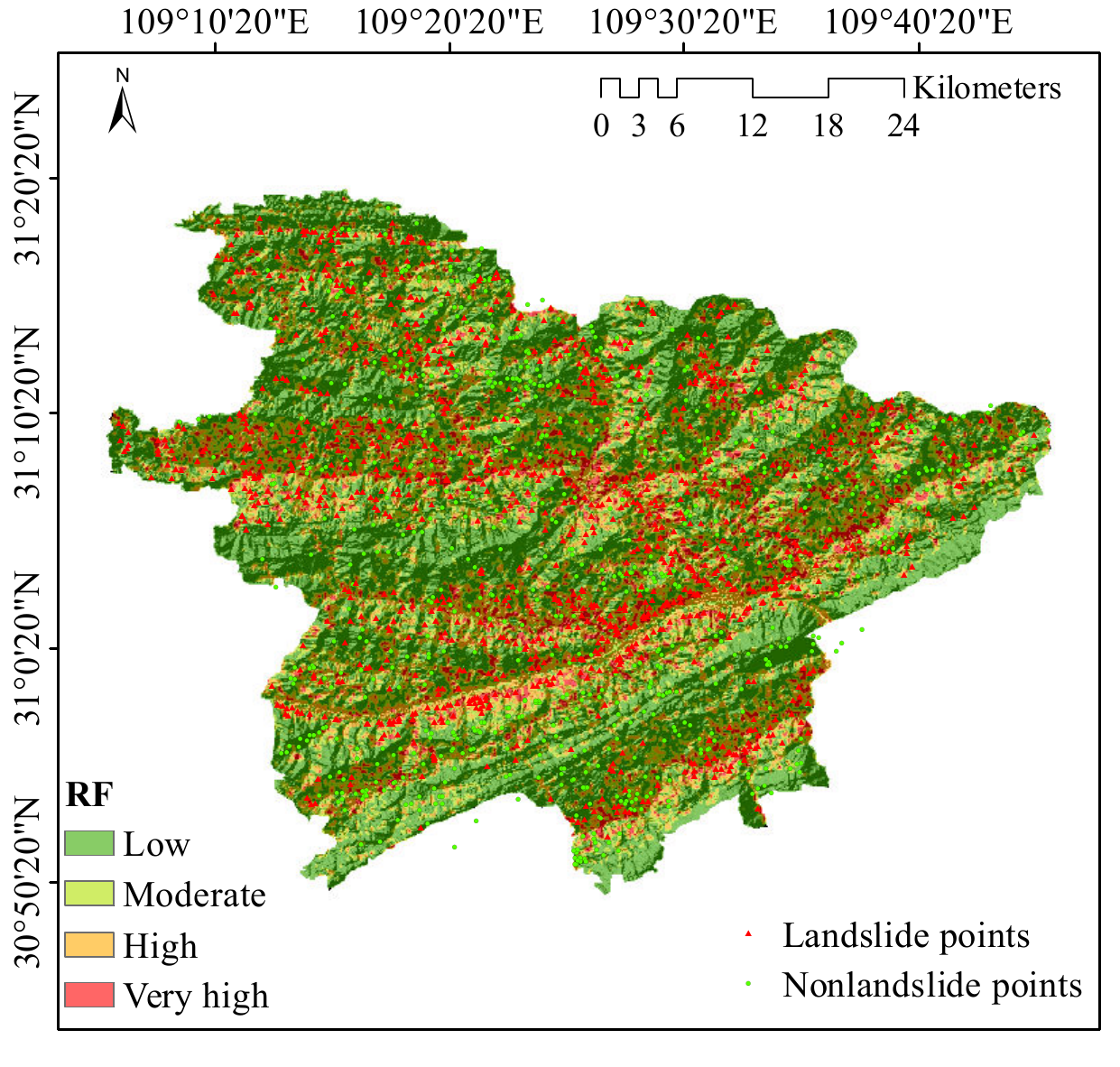}
        \caption{RF-based}
    \end{subfigure}
	\begin{subfigure}{0.32\textwidth}
        \includegraphics[width=\linewidth]{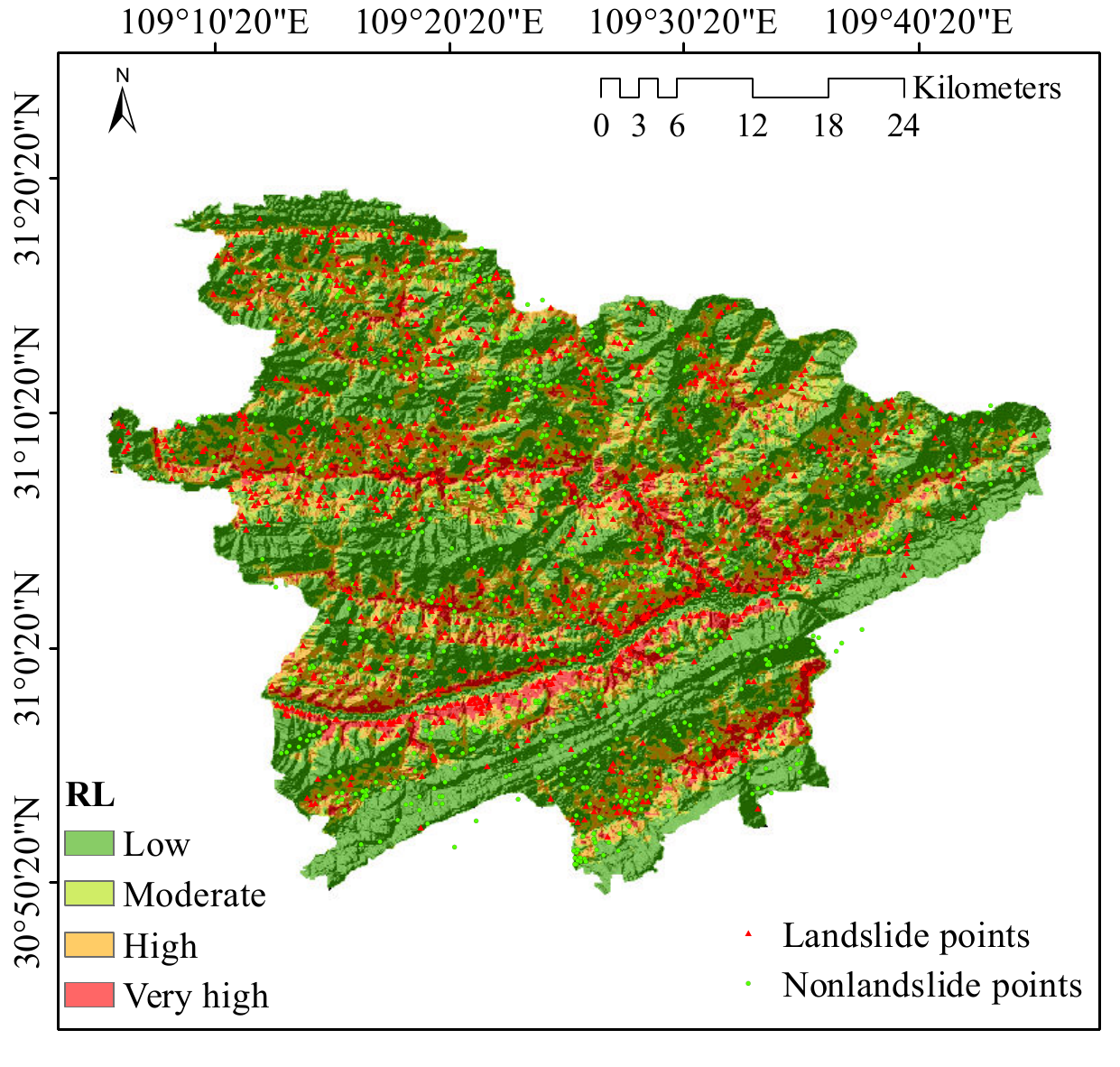}
        \caption{RL-based}
    \end{subfigure}
    \begin{subfigure}{0.32\textwidth}
        \includegraphics[width=\linewidth]{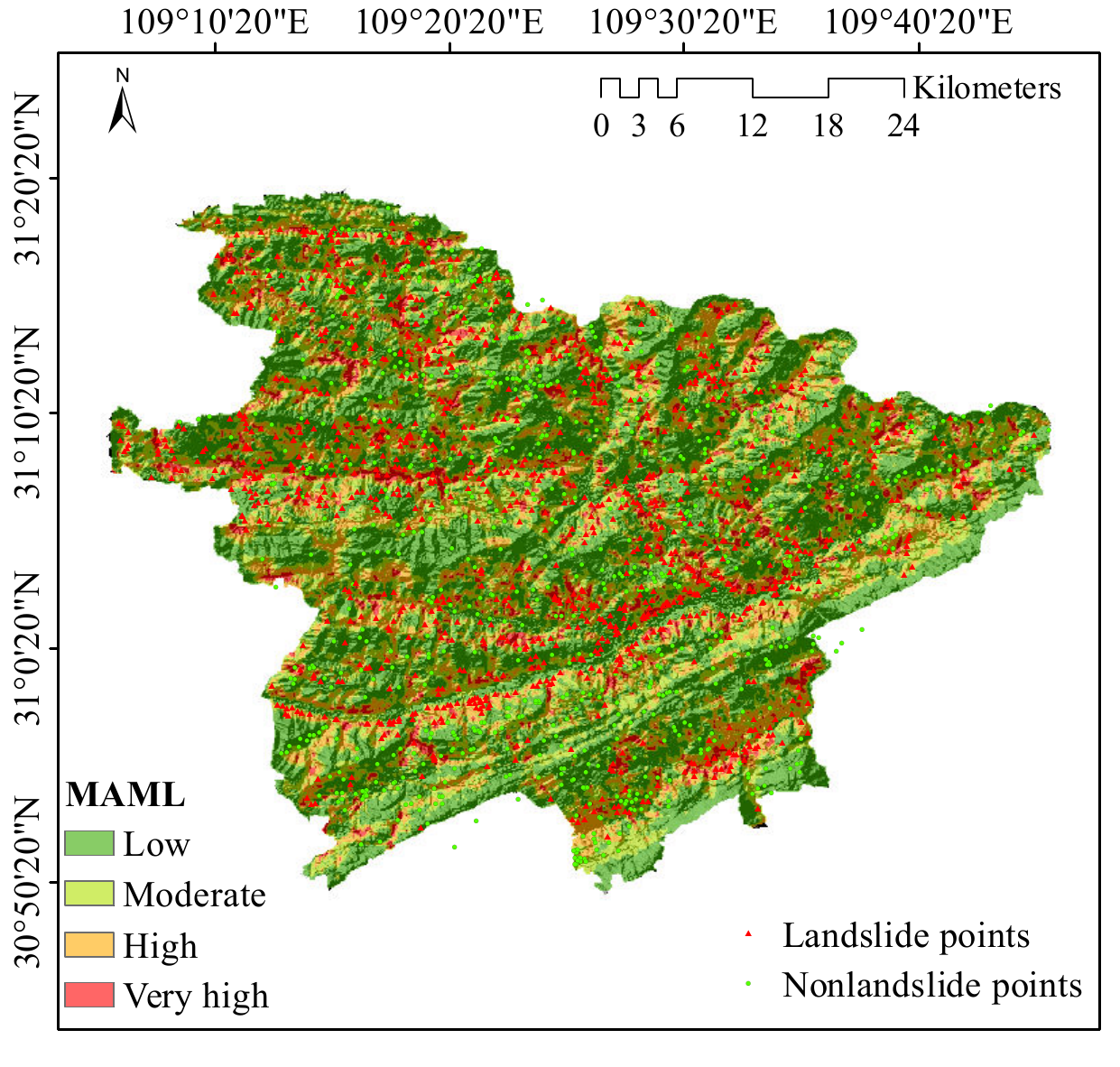}
        \caption{MAML-based}
    \end{subfigure}
    \begin{subfigure}{0.32\textwidth}
        \includegraphics[width=\linewidth]{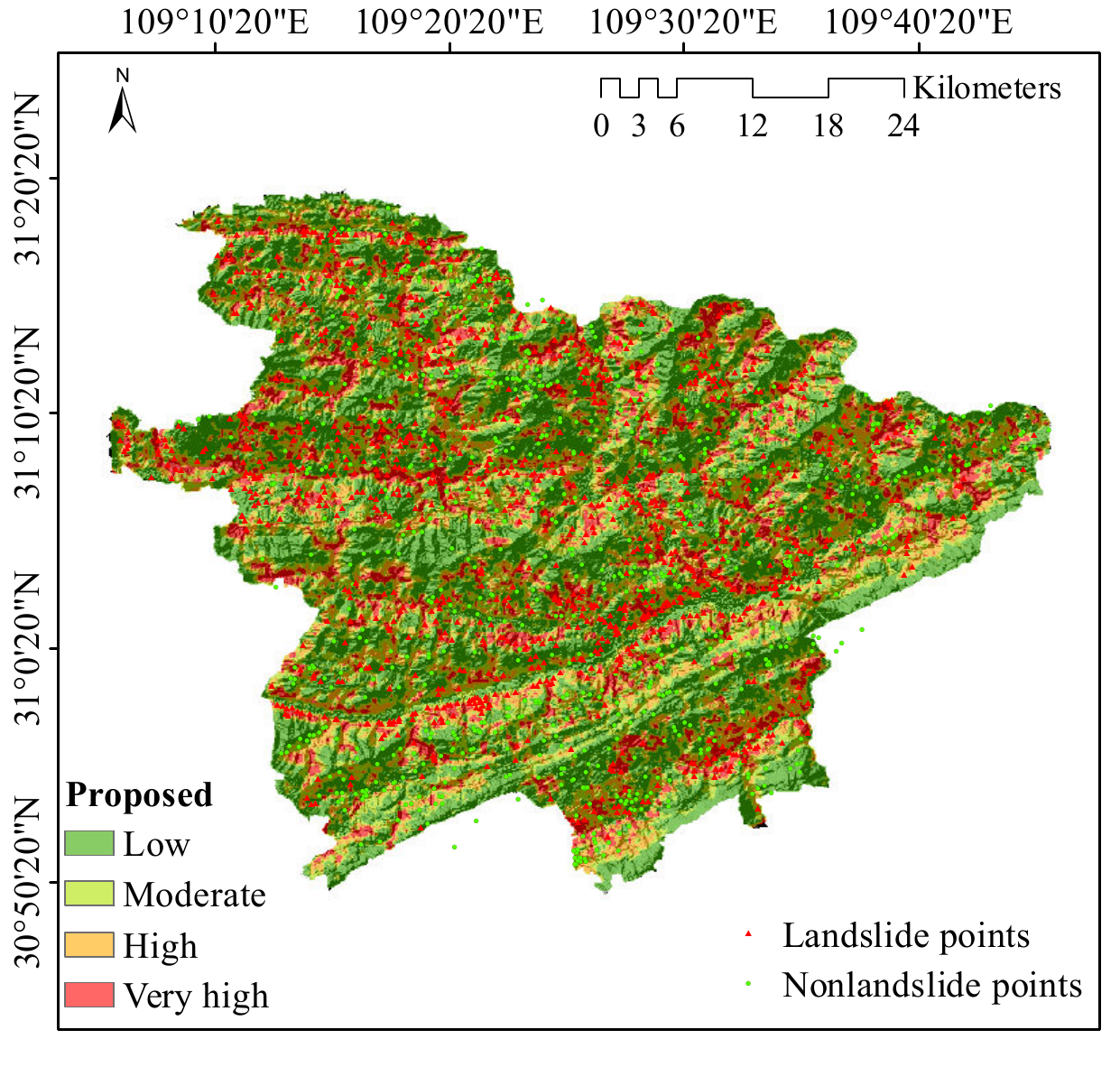}
        \caption{proposed}
    \end{subfigure}
    \caption{LSM prediction in FJ with various methods.}
    \label{fig:LSM of FJ}
\end{figure}

\begin{figure}[H]
    \centering
    \begin{subfigure}{0.32\textwidth}
        \includegraphics[width=\linewidth]{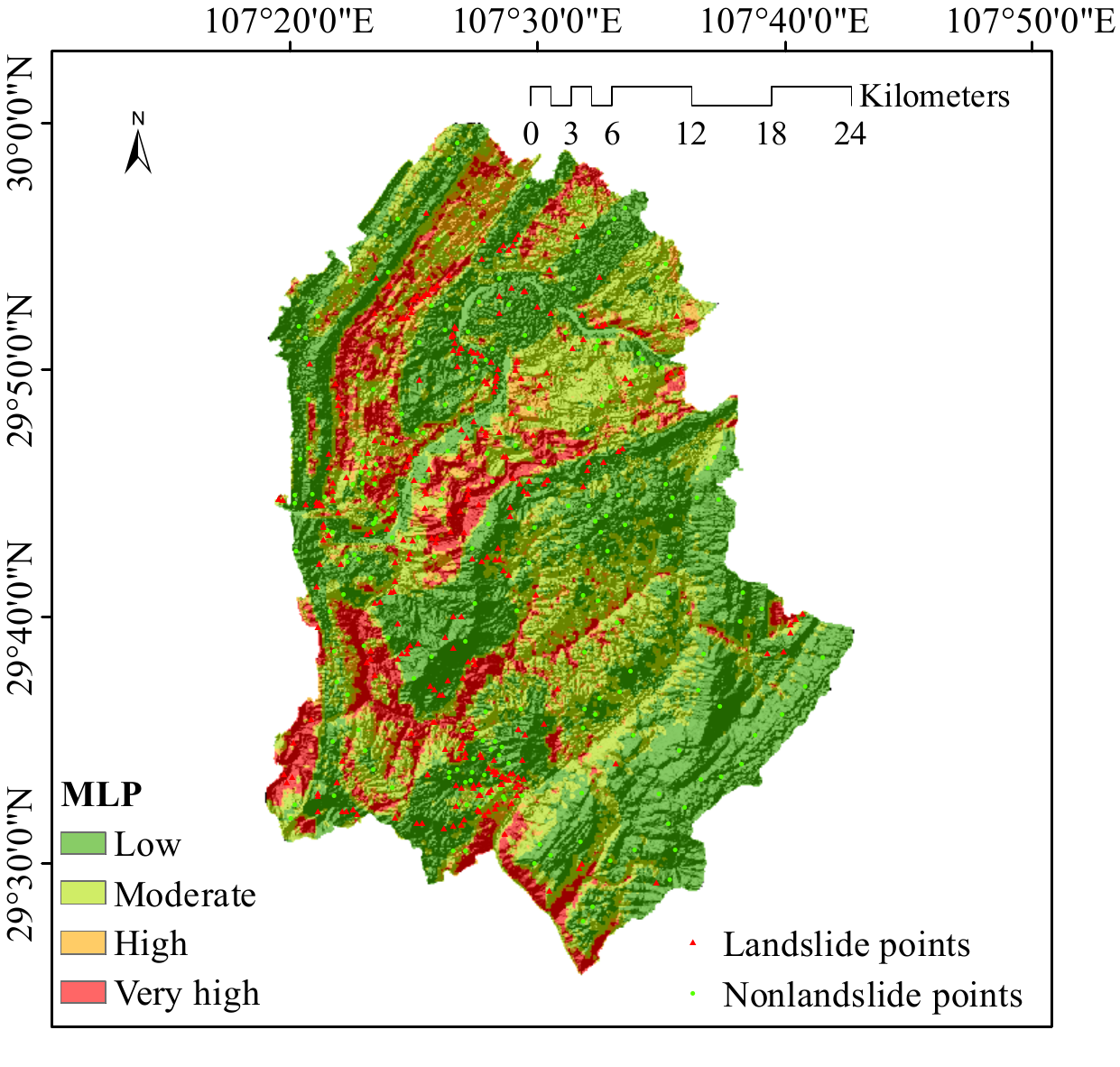}
        \caption{MLP-based}
    \end{subfigure}
    \begin{subfigure}{0.32\textwidth}
        \includegraphics[width=\linewidth]{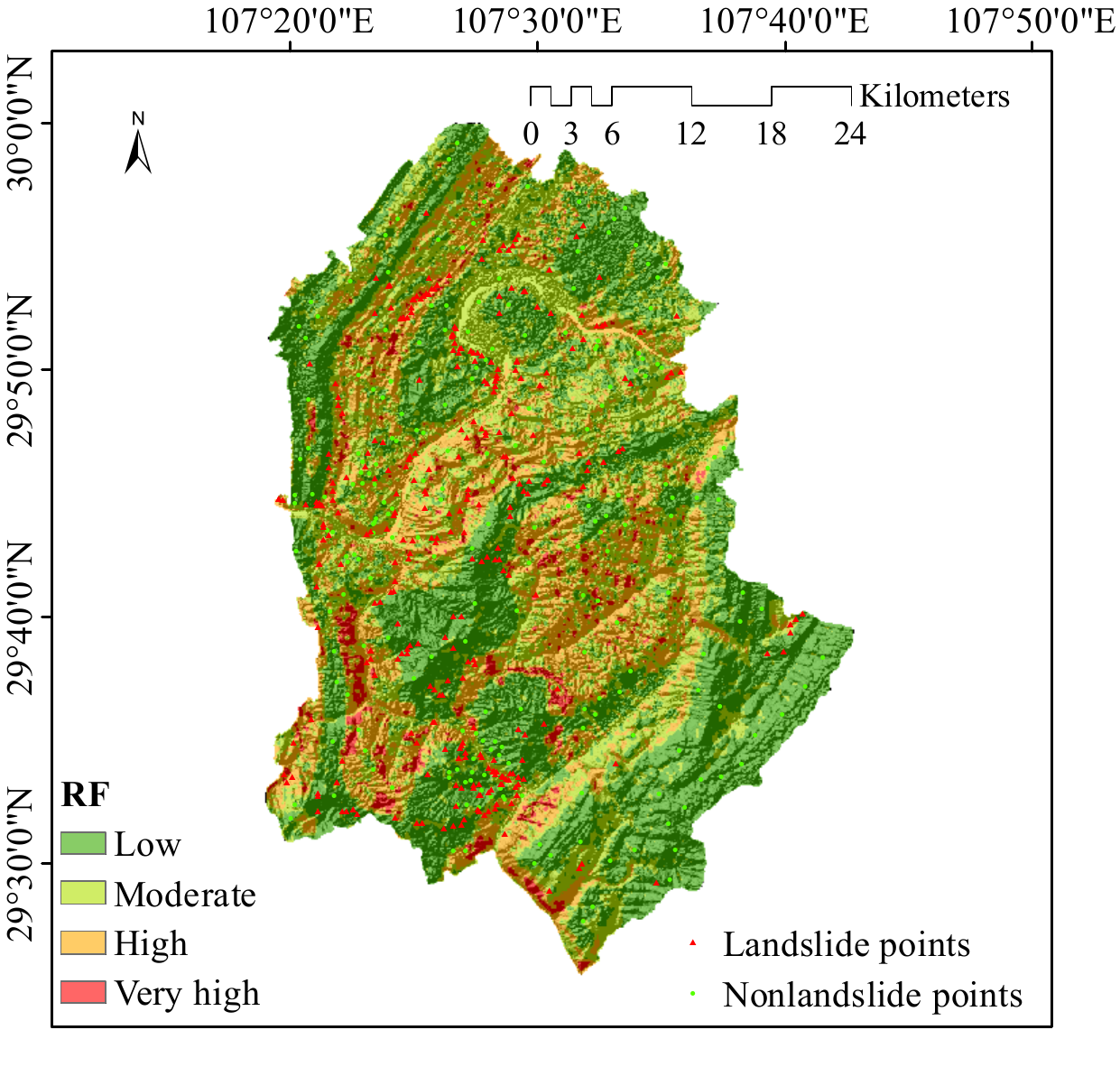}
        \caption{RF-based}
    \end{subfigure}
	\begin{subfigure}{0.32\textwidth}
        \includegraphics[width=\linewidth]{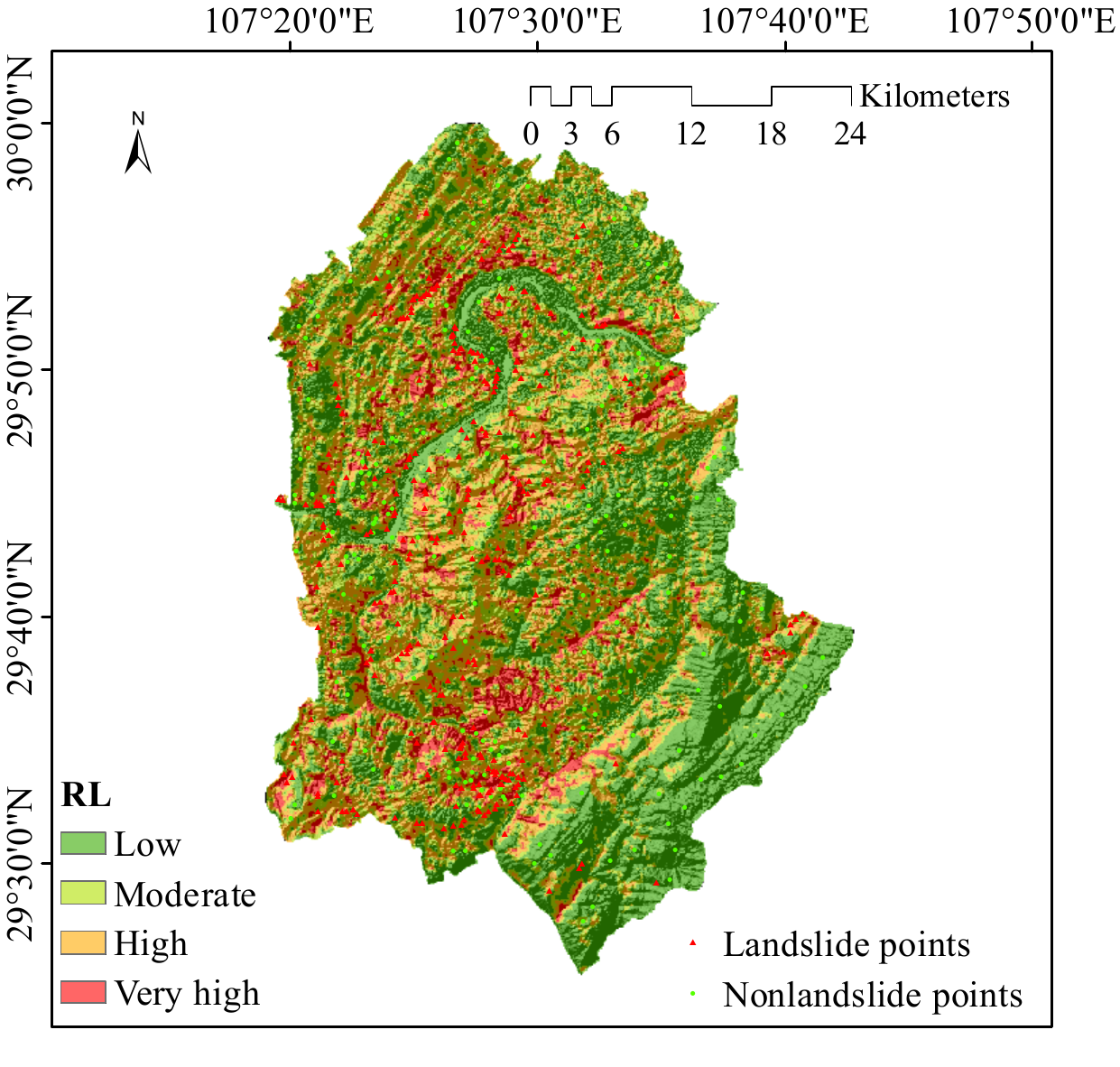}
        \caption{RL-based}
    \end{subfigure}
    \begin{subfigure}{0.32\textwidth}
        \includegraphics[width=\linewidth]{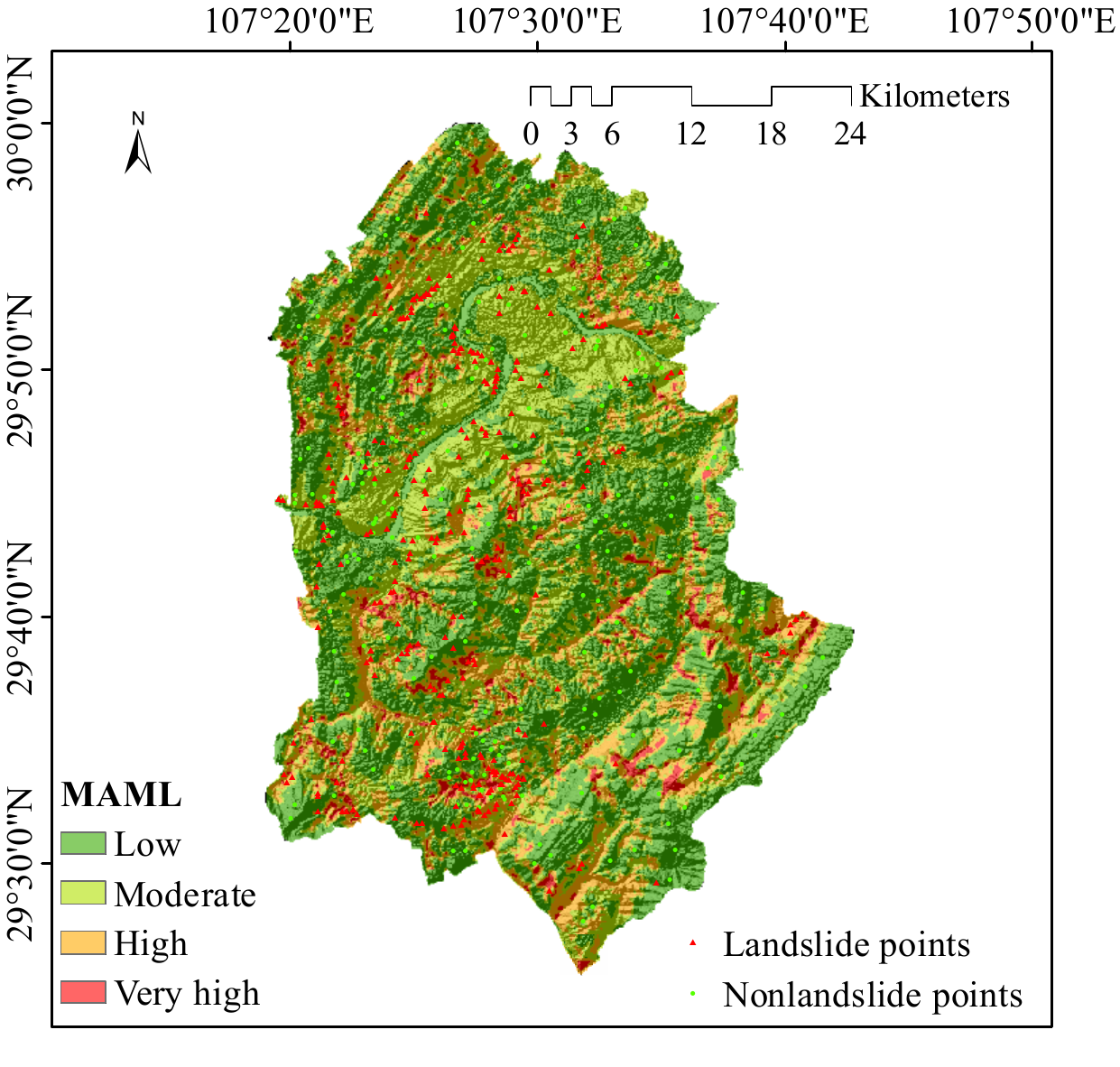}
        \caption{MAML-based}
    \end{subfigure}
    \begin{subfigure}{0.32\textwidth}
        \includegraphics[width=\linewidth]{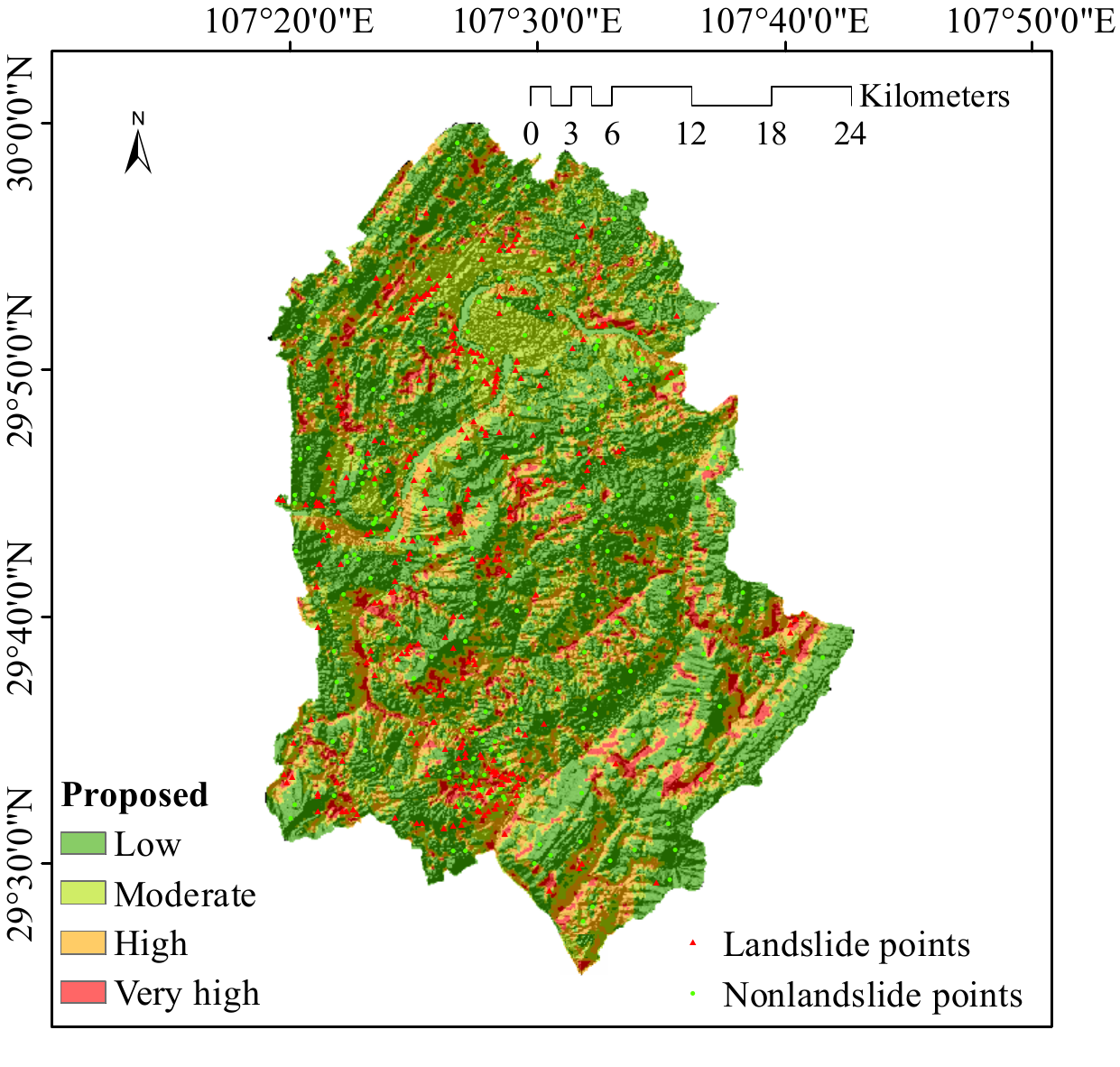}
        \caption{proposed}
    \end{subfigure}
    \caption{LSM prediction in FL with various methods.}
    \label{fig:LSM of FL}
\end{figure}

We compare our method with MLP-based \citep{zare2013landslide}, state-of-the-art RF-based \citep{catani2013landslide}, and an unsupervised representation-learning-based (RL-based) \citep{zhu2020unsupervised} approaches. 
In addition, to further demonstrate the validity of the proposed method, we compare it with the typical few-shot learning approach MAML \citep{finn2017model}. Fig. \ref{fig:LSM of FJ} and Fig. \ref{fig:LSM of FL} depict the LSMs of FJ and FL by different approaches. 
There are four levels of susceptibility separated by predictive value, with a nearly equal number of raster samples in each susceptibility range, for intuitive comparison.

All methods predicted plausible LSMs in the FJ area with numerous evenly distributed samples, whereas in the sample-scarce area FL, the predictions were noticeably different. 
The MLP-based method showed a preference for positive prediction to the west of FL, where landslide samples were concentrated, while negative prediction was shown to the east of FL, where landslide samples were scarce, indicating that the predictive model was trapped in the local minimum. 
The LSM predicted by RF tended to produce discontinuous patches, reflecting from the side that the model suffered from \textit{overfitting}. \cite{zhu2020unsupervised} gives a more evident and interpretable result, in that some dominant causative factors, for example, distance to the drainage, slope, and strata type, should exert more significant influence on predicting the LSM. 
However, that method was also vulnerable to having very few samples because it failed to capture prominent conceptions in the east FL quickly. 
Additionally, the model lost sight of the discrepancy between landslide-inducing environments in different parts of the FL. 
In comparison, our method learns a general intermediate model such that vital conception can be quickly learned for local tasks with fewer samples and iterations.

\begin{figure}[H]
    \centering
    \begin{subfigure}{0.48\textwidth}
        \includegraphics[width=\linewidth]{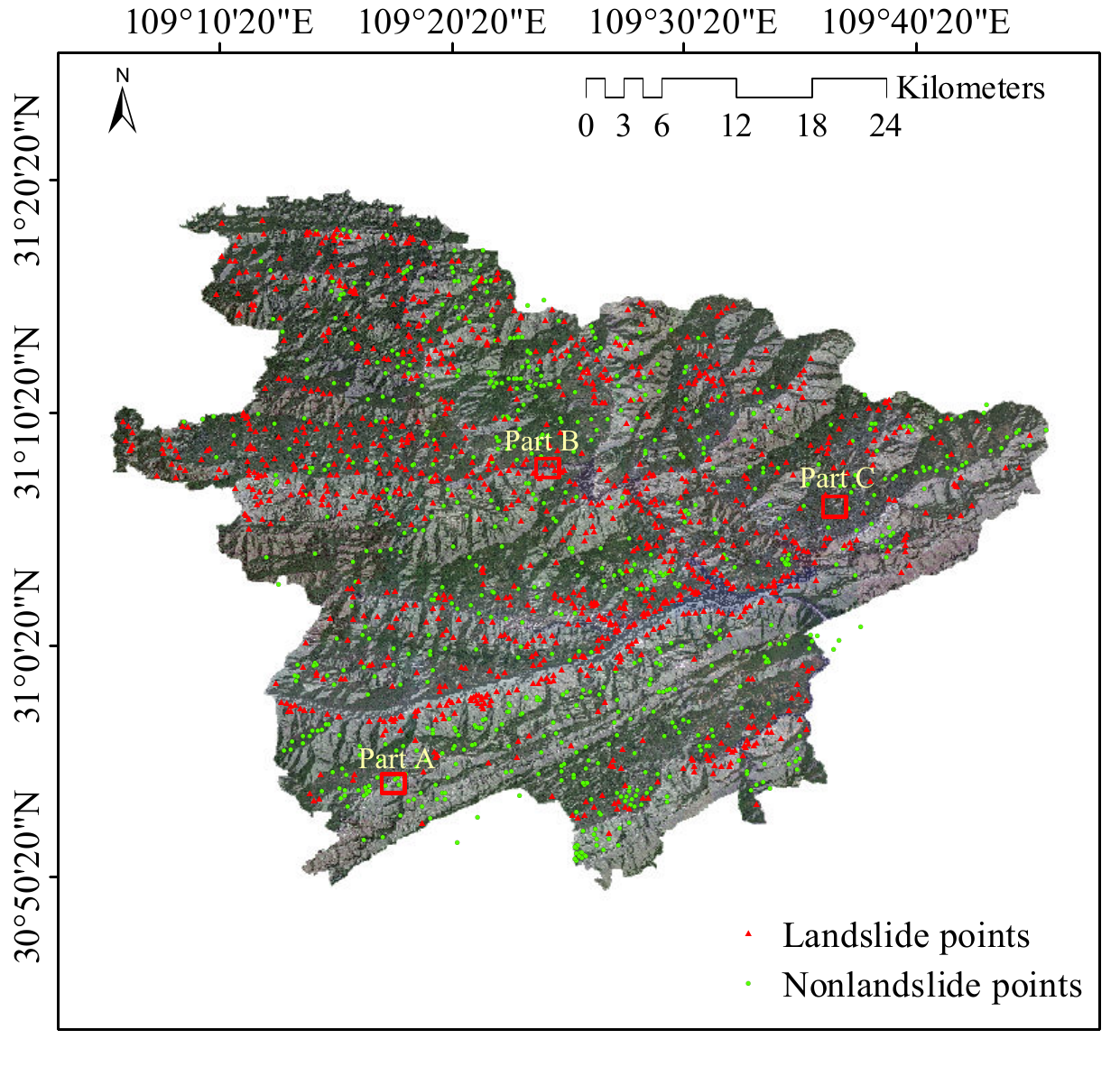}
        \caption{FJ area}
    \end{subfigure}
    \begin{subfigure}{0.48\textwidth}
        \includegraphics[width=\linewidth]{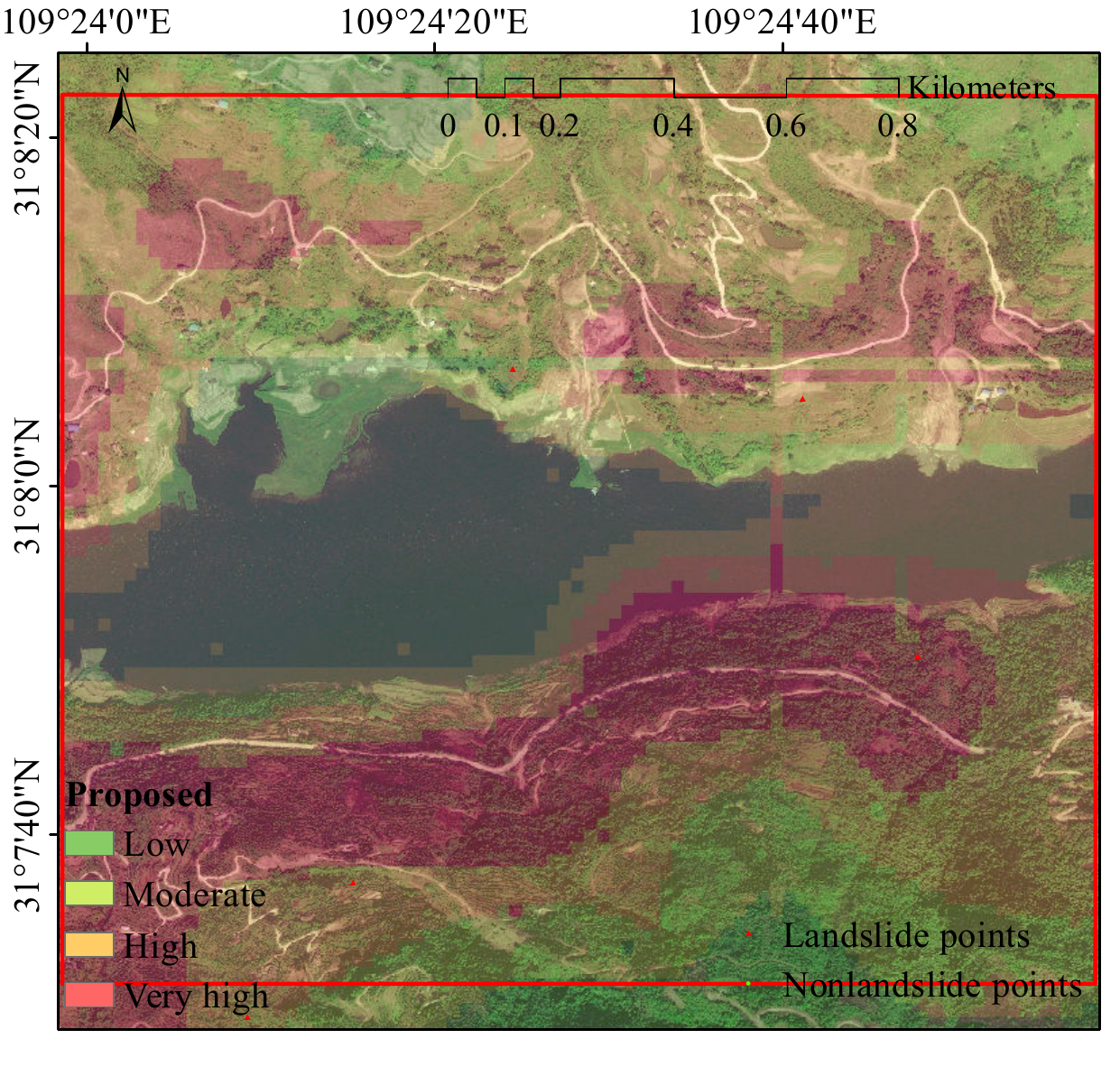}
        \caption{LSM of Part A}
    \end{subfigure}
	\begin{subfigure}{0.48\textwidth}
        \includegraphics[width=\linewidth]{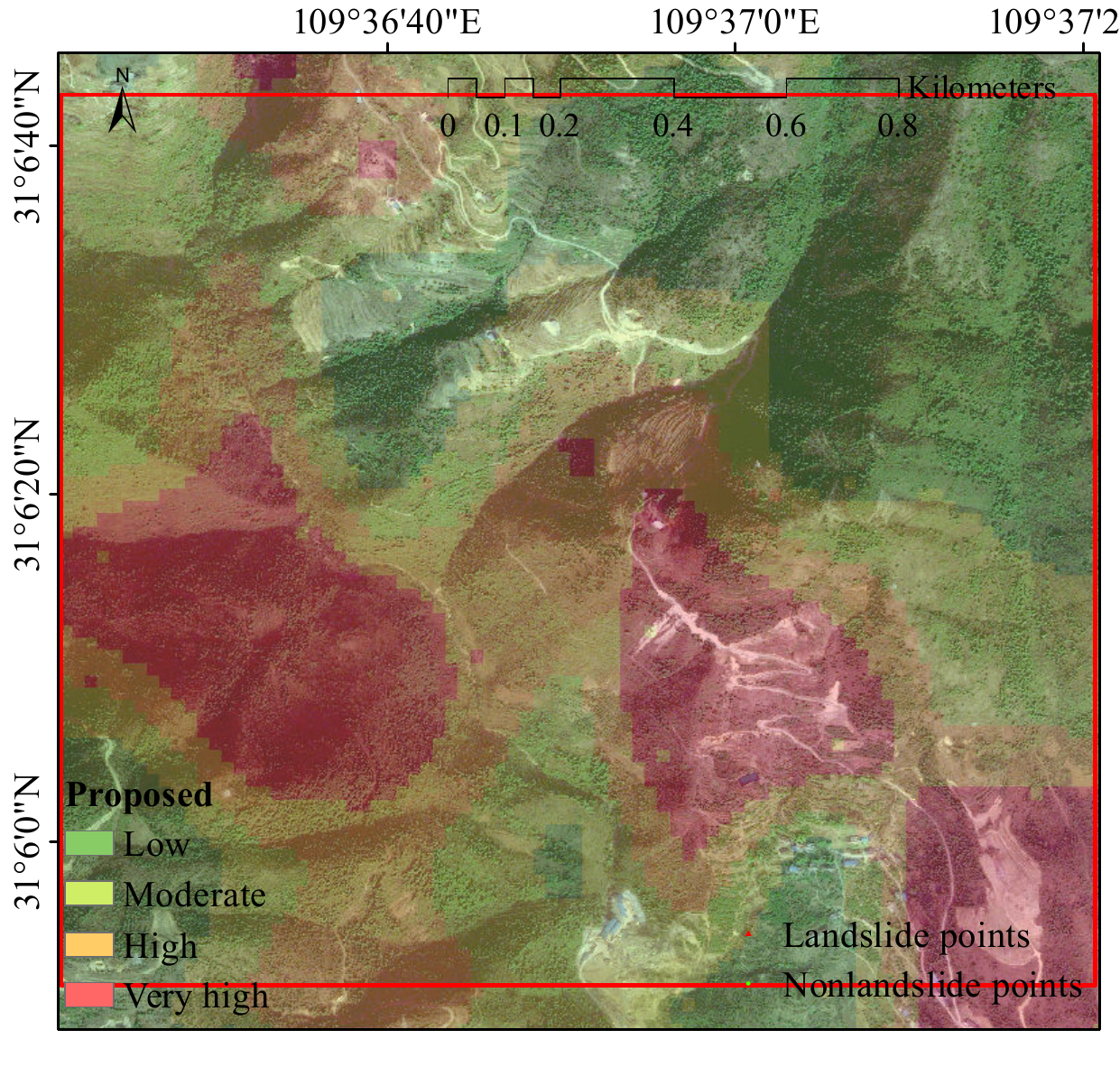}
        \caption{LSM of Part B}
    \end{subfigure}
    \begin{subfigure}{0.48\textwidth}
        \includegraphics[width=\linewidth]{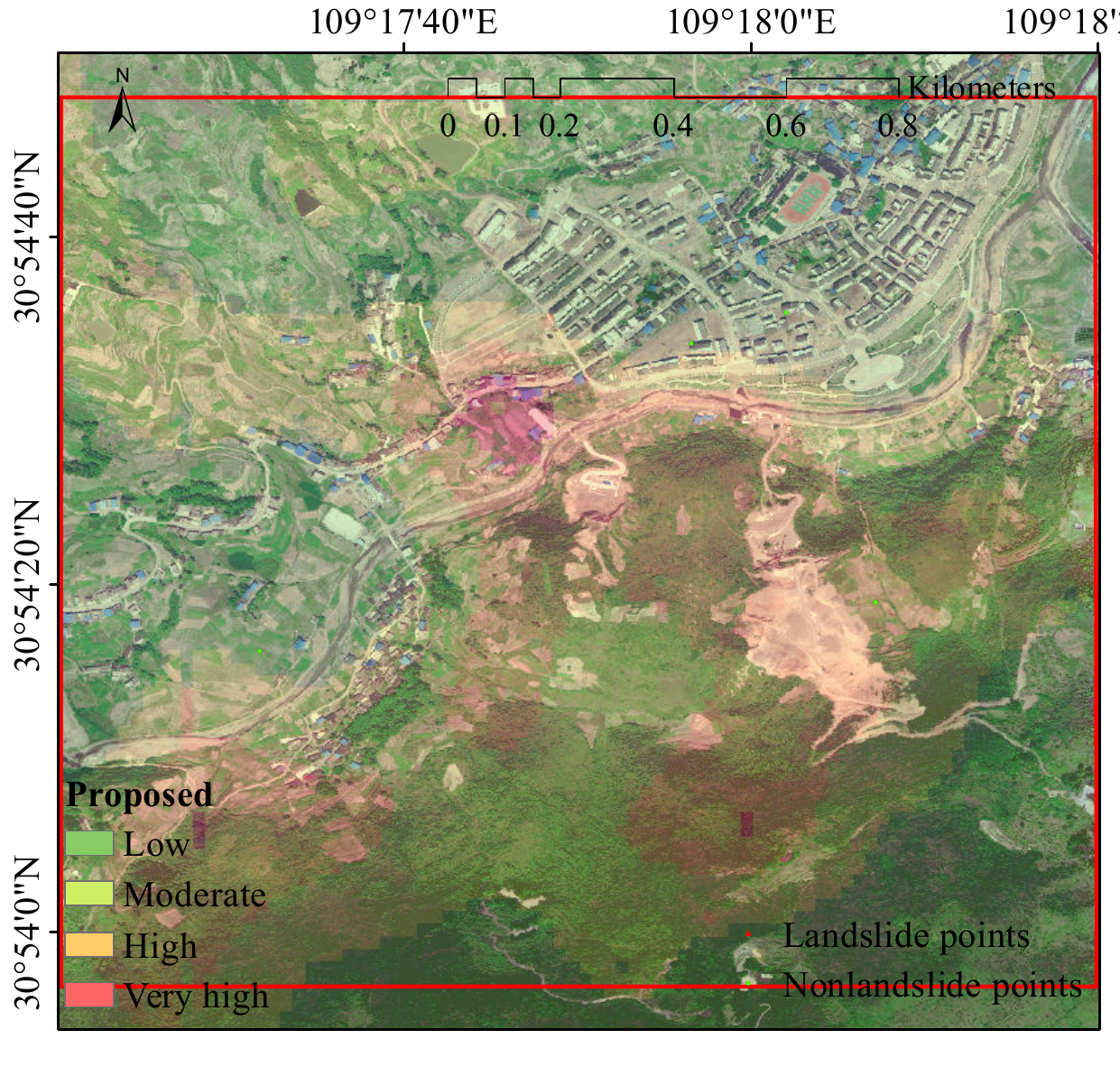}
        \caption{LSM of Part C}
    \end{subfigure}
    \caption{Display of LSMs in various parts of FJ. Specifically, we chose three typical landforms: water (part A), mountainous (part B), and residential areas (part C).}
    \label{fig:partLSM_FJ}
\end{figure}

\begin{figure}[H]
    \centering
    \begin{subfigure}{0.48\textwidth}
        \includegraphics[width=\linewidth]{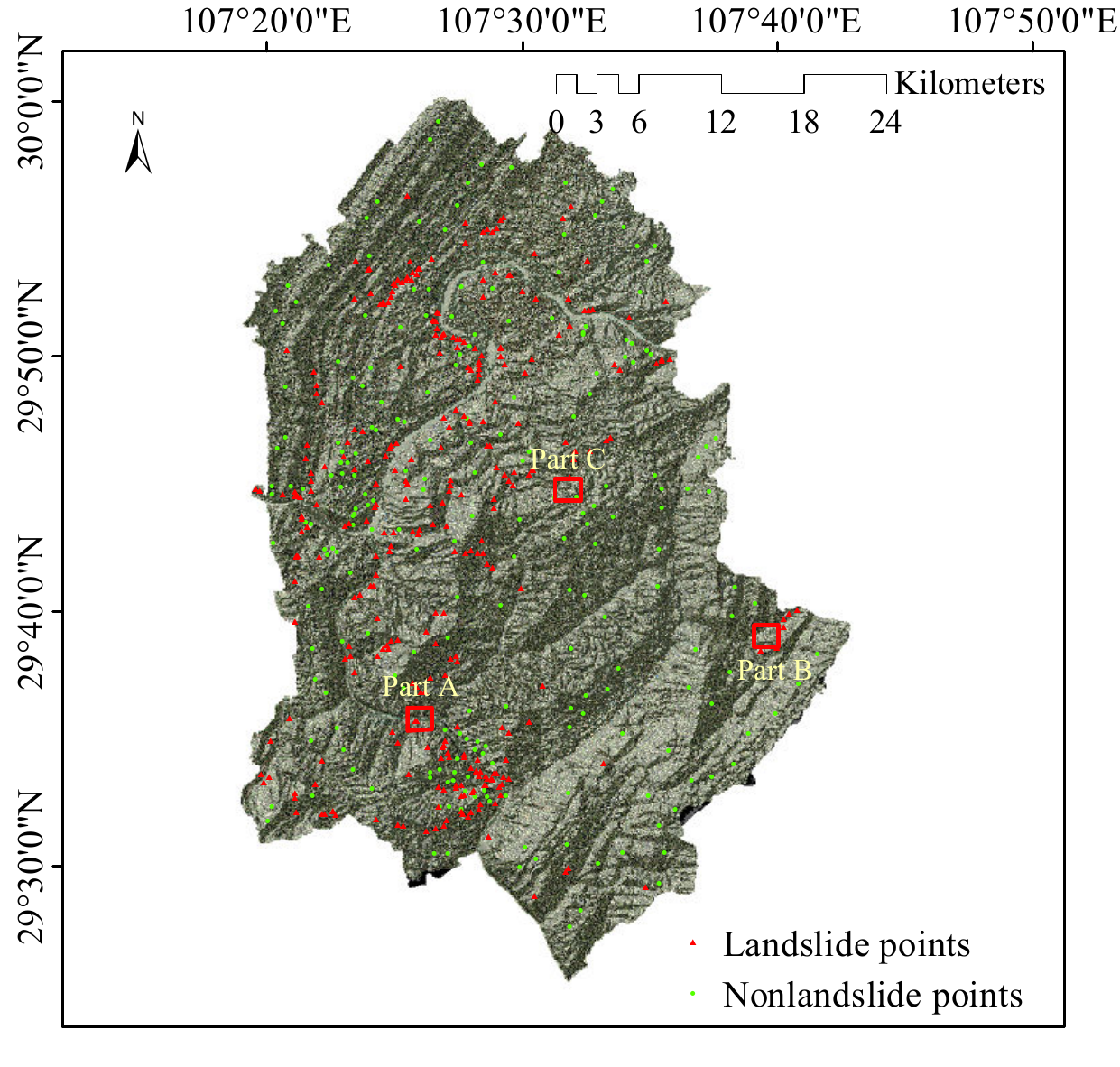}
        \caption{FJ area}
    \end{subfigure}
    \begin{subfigure}{0.48\textwidth}
        \includegraphics[width=\linewidth]{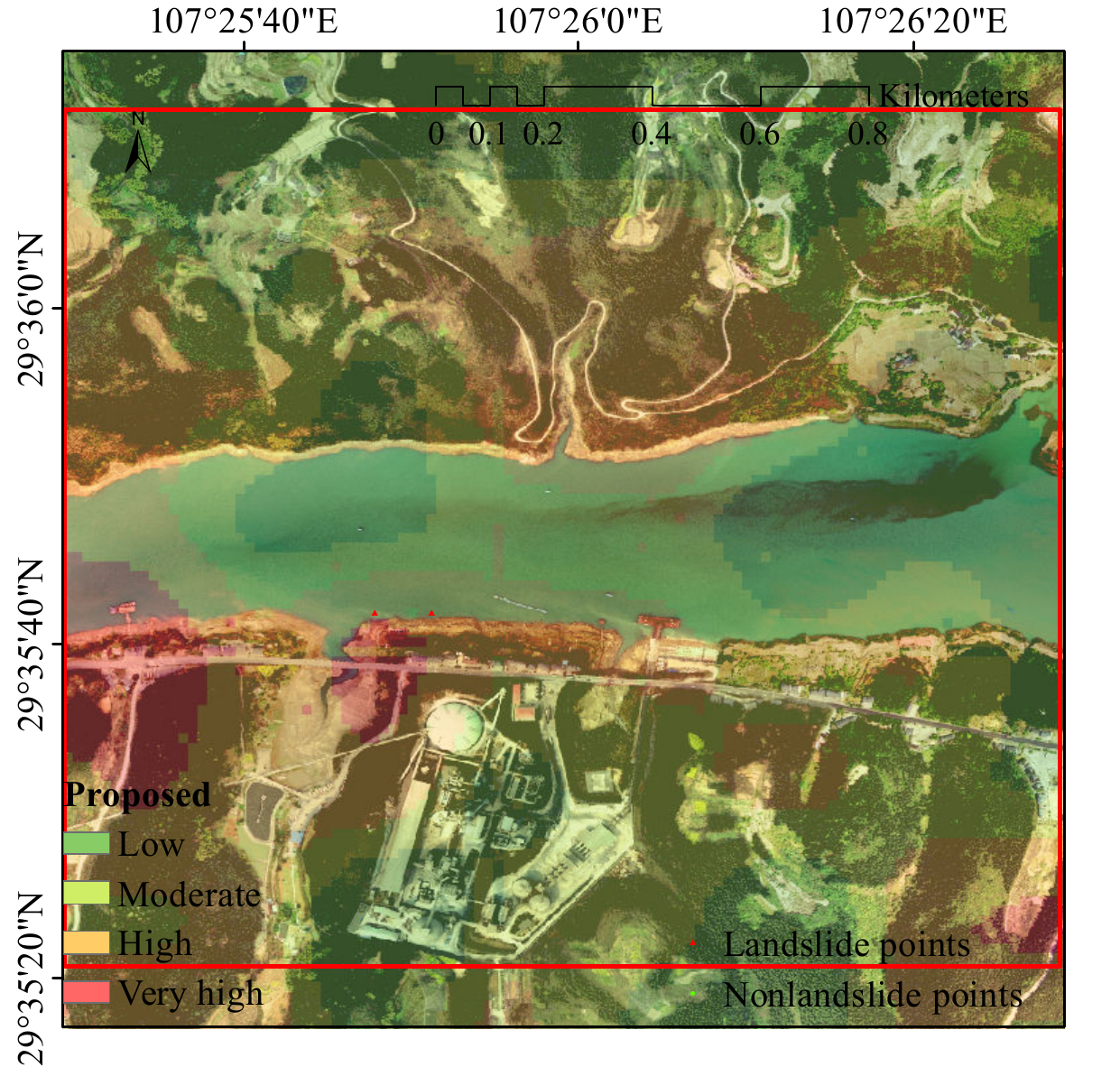}
        \caption{LSM of Part A}
    \end{subfigure}
	\begin{subfigure}{0.48\textwidth}
        \includegraphics[width=\linewidth]{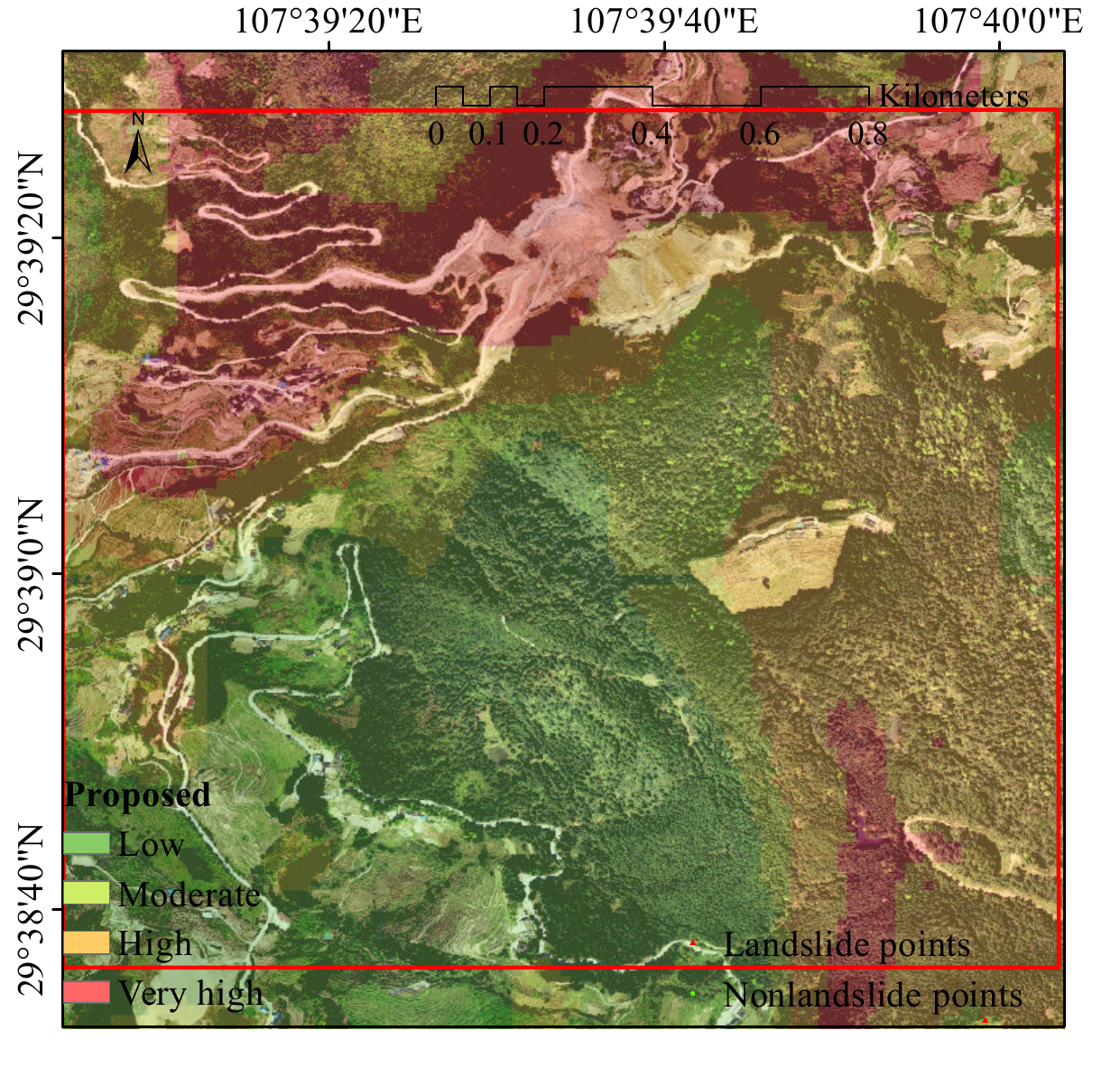}
        \caption{LSM of Part B}
    \end{subfigure}
    \begin{subfigure}{0.48\textwidth}
        \includegraphics[width=\linewidth]{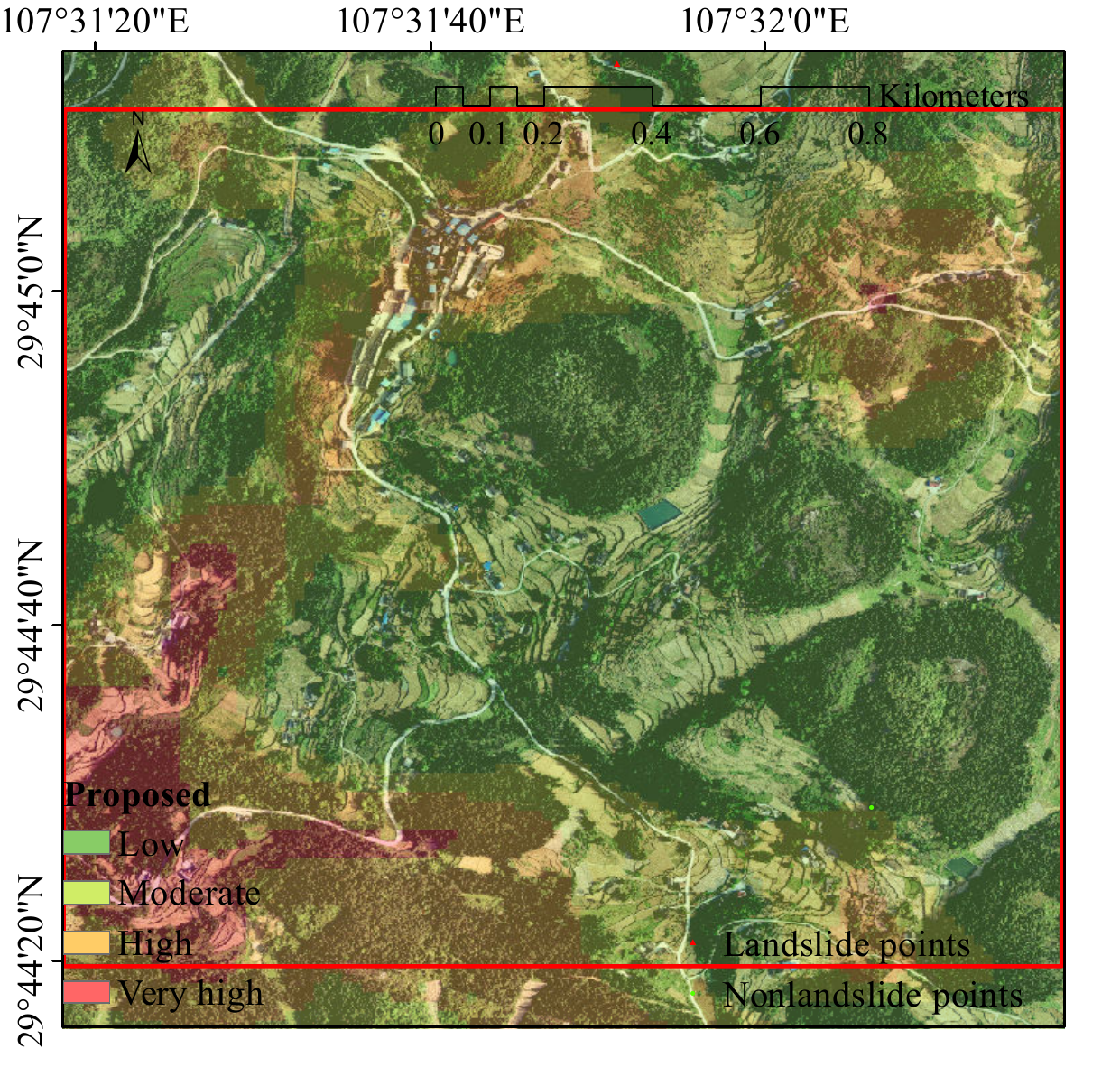}
        \caption{LSM of Part C}
    \end{subfigure}
    \caption{Display of LSMs in various parts of FL. Specifically, we chose three typical landforms: water (part A), mountainous (part B), and residential areas (part C).}
    \label{fig:partLSM_FL}
\end{figure}

For the intuitionistic perception of the landslide susceptibility prediction results, the LSMs of some parts of the FJ and FL regions are displayed in Fig. \ref{fig:partLSM_FJ} and Fig. \ref{fig:partLSM_FL}.
The distance to the river and road has played an important role in landslides, whereas the residential areas are normally assigned with low landslide susceptibility. 
The proposed model has likely captured certain vital aspects in the few-shot adaption process.
In addition, the figures also highlight indeed occurred landslides with high susceptibility, which proves the validity and reasonability of our method.

\subsubsection{Performance evaluation}
\label{subsubs:performance evaluation}
In this section, we bring in \citep{merghadi2018landslide} some statistical measures to prove the validity of our method, which includes accuracy, precision, recall, and F1-score. They are given in Eq. \ref{eq:measurement},

\begin{equation}
    \begin{split}
    \begin{aligned}
        Accuracy&=\frac{TP+TN}{TP+FP+TN+FN},\\
        Precision&=\frac{TP}{TP+FP},\\
        Recall&=\frac{TP}{TP+FN},\\
        F1-score&=\frac{2TP}{2TP+FP+FN},
    \end{aligned}
  \label{eq:measurement}
\end{split}
\end{equation}
where $TN$, $TP$, $FN$, and $FP$ denote the Ture negative, true positive, false negative, and false positive, respectively.

\begin{table*}[htbp]
    \caption{Performance comparison of various LSM models in FJ and FL. Despite the difficulties to few-shot learn the predictive models, our method still reaches the state-of-the-art.}
    \centering
    \begin{tabular}{clcccc}
      \hline
      Study areas                    & Models           & Accuracy(\%)     &Precision(\%)     & Recall(\%)      & F1-score(\%) \\
      \hline
      \multirow{5}{*}{FJ}            & MLP              & 73.5             & 72.3             & 69.9            & 70.9   \\
                                     & RF               & 74.8             & \textcolor{blue}{\textbf{74.7}}    & 65.3            & 69.4   \\
                                     & RL               & 74.9             & 70.2             & \textcolor{blue}{\textbf{77.5}}   & \textcolor{blue}{\textbf{77.3}}   \\
                                     & MAML             & 72.6             & 69.3             & 72.7            & 68.3   \\ 
                                     & Proposed         & \textcolor{blue}{\textbf{75.1}}    & 72.0             & 76.6            & 74.0   \\
      \hline
      \multirow{5}{*}{FL}            & MLP              & 76.7             & 77.1             & 76.6            & 76.6   \\
                                     & RF               & 77.8             & 80.1             & 73.3            & 76.3   \\
                                     & RL               & \textcolor{blue}{\textbf{81.6}}    & 78.7             & \textcolor{blue}{\textbf{85.6}}   & 81.2   \\
                                     & MAML             & 76.4             & 83.7             & 83.4            & 82.0   \\
                                     & Proposed         & 78.4             & \textcolor{blue}{\textbf{86.9}}    & 83.5            & \textcolor{blue}{\textbf{85.0}}   \\ \hline
    \end{tabular}
    \label{tab:statistical measures}
  \end{table*}

Table \ref{tab:statistical measures} shows the performance of various methods in FJ and FL. 
The proposed method provides the highest accuracy of 75.1\% in FJ. Also, it performs pretty well on other statistical measures, even with a few samples and gradient updates for adaption. 
It is different from MLP-based \citep{zare2013landslide}, RF-based \citep{catani2013landslide}, and RL-based \citep{zhu2020unsupervised} methods trained by whole samples within a ministrative region with hundreds or thousands of update iterations. 
It verifies the feasibility of assuming that various parts of a ministrative region could possess different landslide causes, and the LSM predictive model of each block could be few-shot learned. 
In addition, compared to the typical few-shot learning approach MAML \citep{finn2017model}, the adopted tricks realize distinct improvements in performance.


\subsection{Effect of unsupervised pretraining}
\label{subs:effect of unsupervised pretraining}
To explore the properties of unsupervised learned representation, we utilized some visualization approaches, including principal component analysis (PCA) \citep{wold1987principal}, Isomap \citep{balasubramanian2002isomap}, t-distributed stochastic neighbor embedding (t-SNE) \citep{van2008visualizing}, and uniform manifold approximation and projection (UMAP) \citep{mcinnes2018umap}, to visualize the embedding spaces of the original data, the MAML-based method without an unsupervised training process, and the proposed method, as shown in Table \ref{tab:embedding space}.  
The left column shows the embedding results inferred from the original input space. 
The middle column shows the embedding results inferred from representations learned by the MAML-based method. 
The right column shows the embedding results inferred from the proposed learned representation. 
The red and blue points respectively denote landslide and non-landslide samples. 

\begin{table*}[h]
    \caption{Three-dimensional visualization of embedding space when given only 5 samples ($K=5$) for adaption in FJ.}
    \centering
    \begin{tabular}{p{0.1\textwidth}cccc}
        \hline
        Embedding                   &Original data                             &Without pretraining                                  &Proposed \\ 
        PCA                         &\begin{minipage}[b]{0.25\textwidth}\centering\raisebox{-.5\height}{\includegraphics[width=\linewidth]{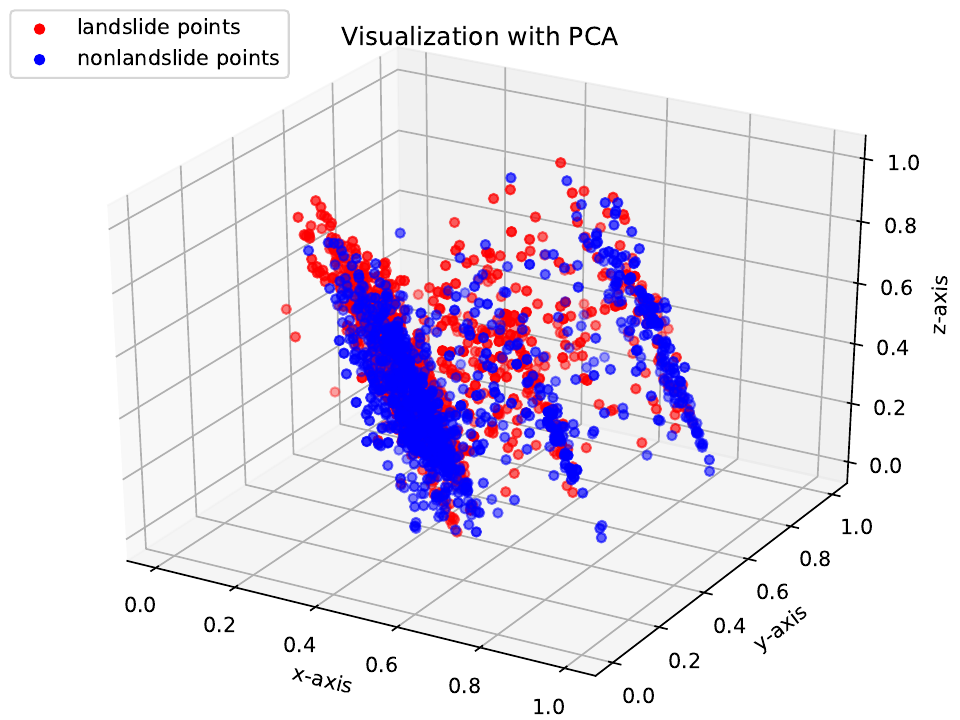}}\end{minipage}     &\begin{minipage}[b]{0.25\textwidth}\centering\raisebox{-.5\height}{\includegraphics[width=\linewidth]{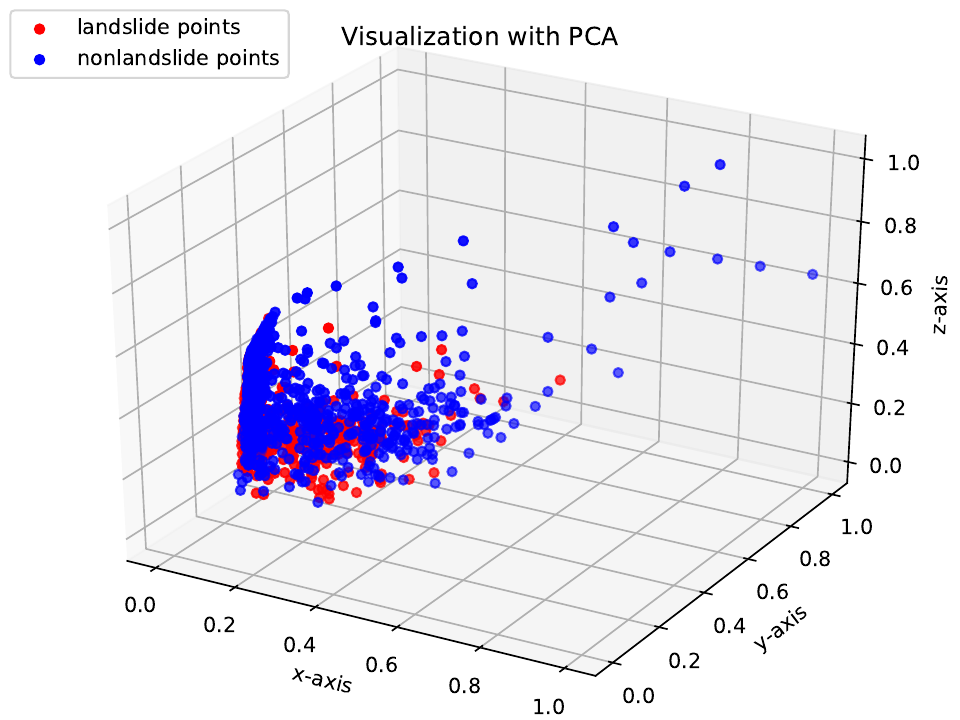}}\end{minipage}     &\begin{minipage}[b]{0.25\columnwidth}\centering\raisebox{-.5\height}{\includegraphics[width=\linewidth]{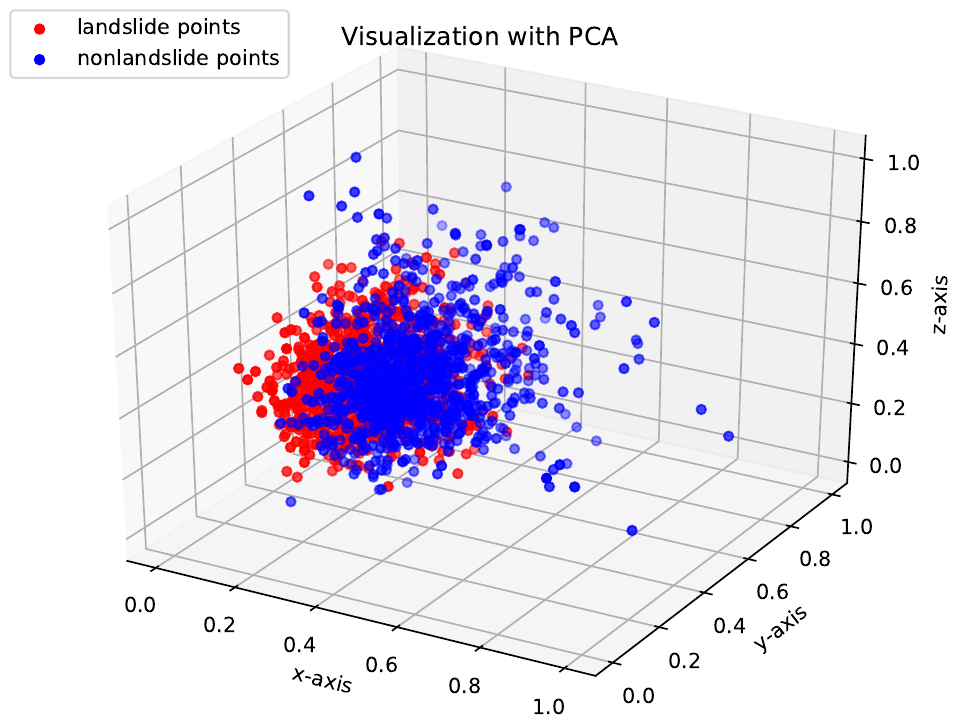}}\end{minipage}    \\                         
        Isomap                      &\begin{minipage}[b]{0.25\textwidth}\centering\raisebox{-.5\height}{\includegraphics[width=\linewidth]{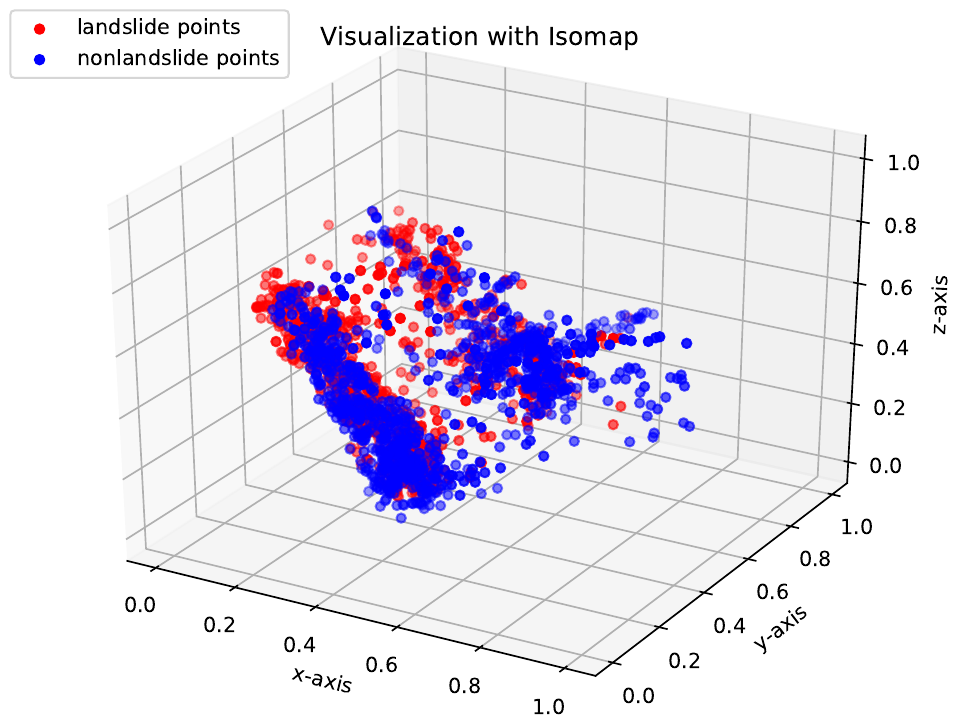}}\end{minipage}  &\begin{minipage}[b]{0.25\textwidth}\centering\raisebox{-.5\height}{\includegraphics[width=\linewidth]{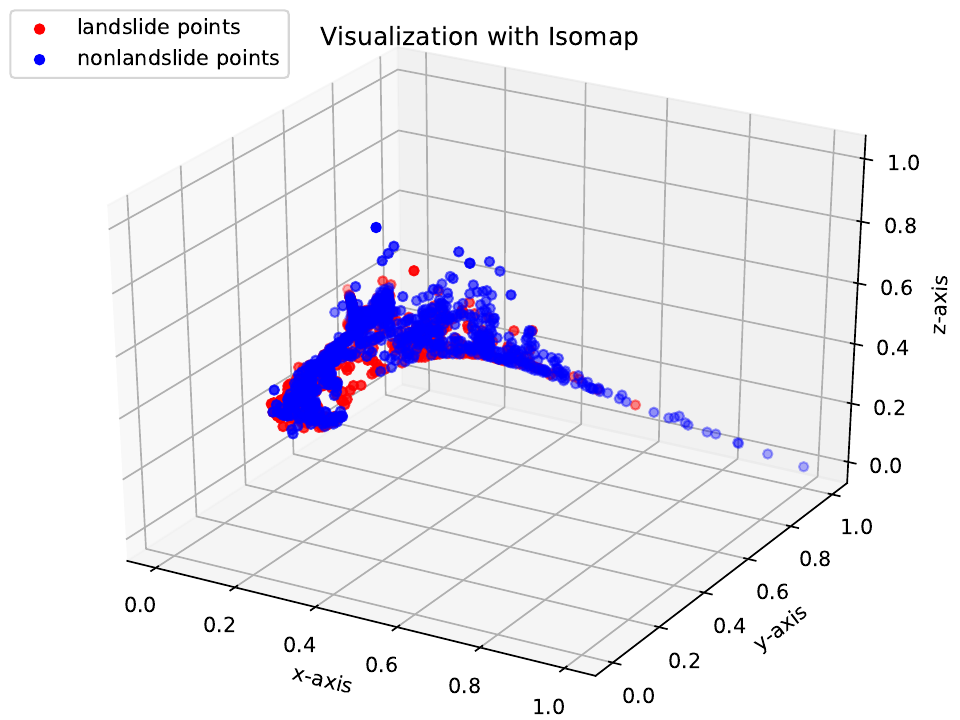}}\end{minipage}  &\begin{minipage}[b]{0.25\columnwidth}\centering\raisebox{-.5\height}{\includegraphics[width=\linewidth]{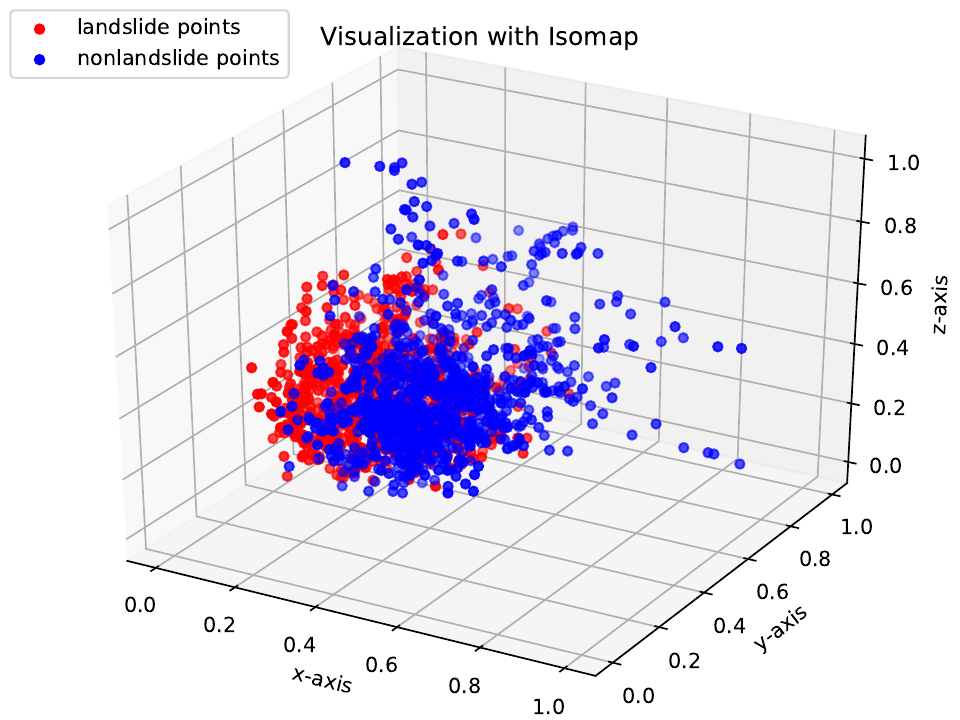}}\end{minipage}  \\
        t-SNE                       &\begin{minipage}[b]{0.25\textwidth}\centering\raisebox{-.5\height}{\includegraphics[width=\linewidth]{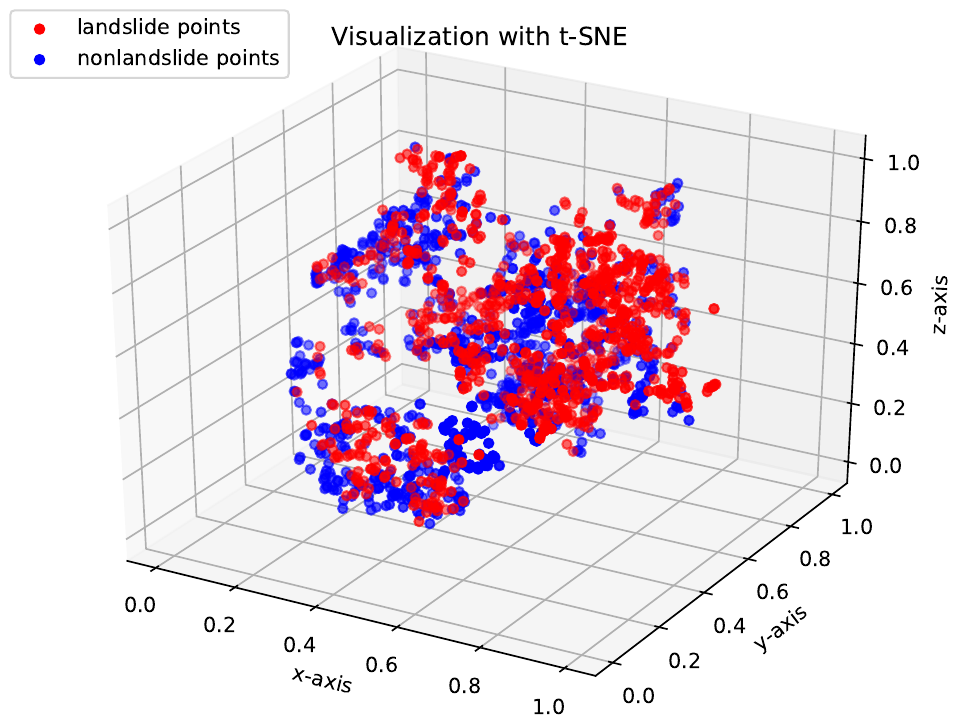}}\end{minipage}   &\begin{minipage}[b]{0.25\textwidth}\centering\raisebox{-.5\height}{\includegraphics[width=\linewidth]{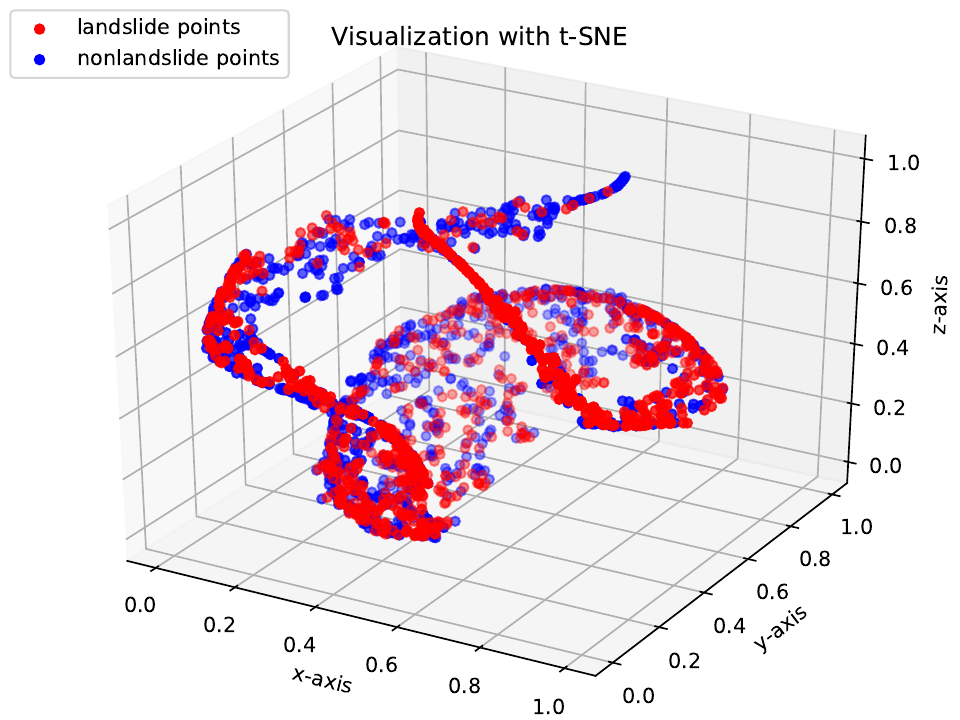}}\end{minipage}   &\begin{minipage}[b]{0.25\columnwidth}\centering\raisebox{-.5\height}{\includegraphics[width=\linewidth]{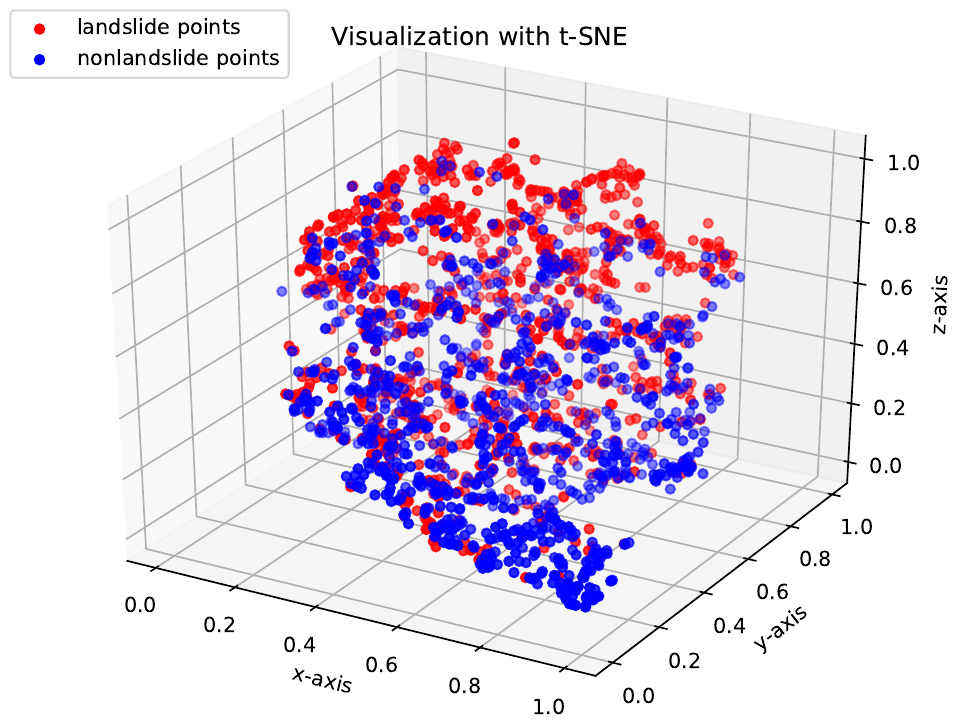}}\end{minipage}  \\
        UMAP                        &\begin{minipage}[b]{0.25\textwidth}\centering\raisebox{-.5\height}{\includegraphics[width=\linewidth]{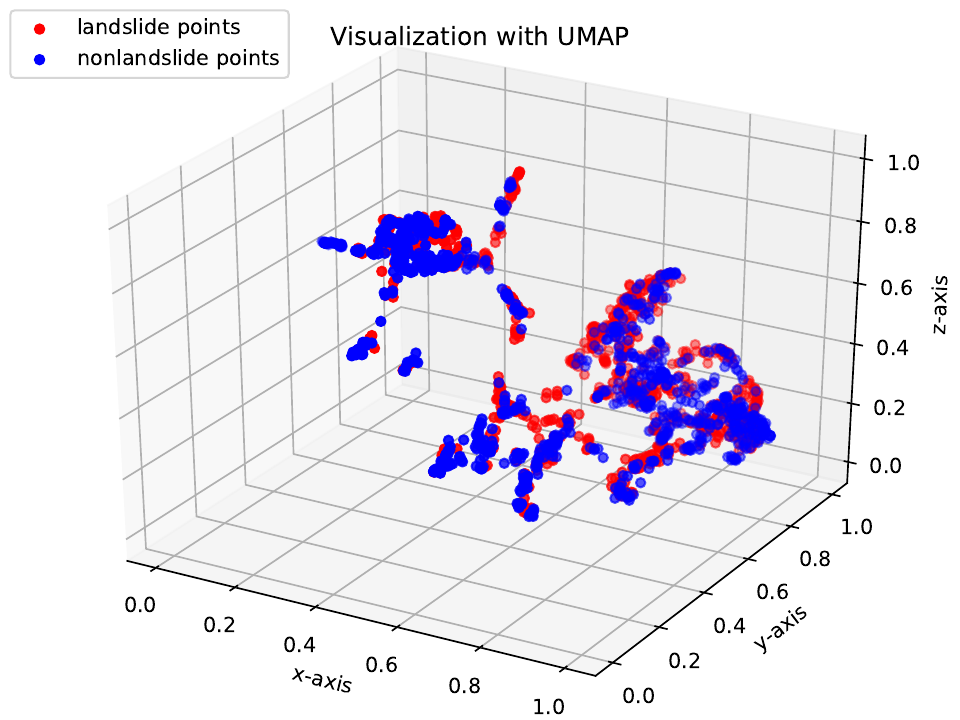}}\end{minipage}    &\begin{minipage}[b]{0.25\textwidth}\centering\raisebox{-.5\height}{\includegraphics[width=\linewidth]{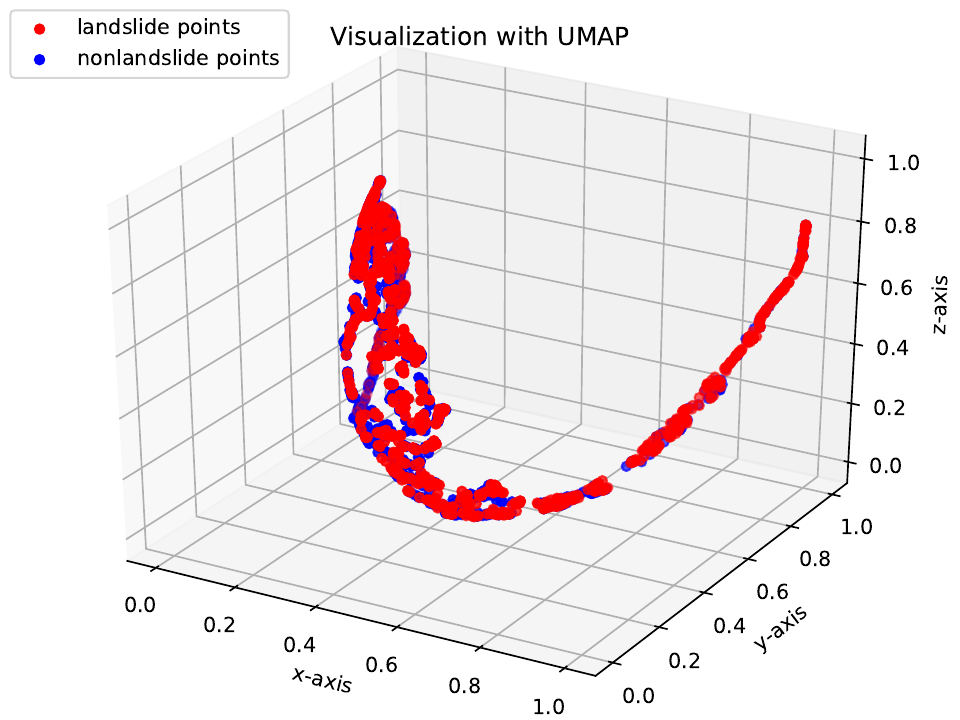}}\end{minipage}    &\begin{minipage}[b]{0.25\columnwidth}\centering\raisebox{-.5\height}{\includegraphics[width=\linewidth]{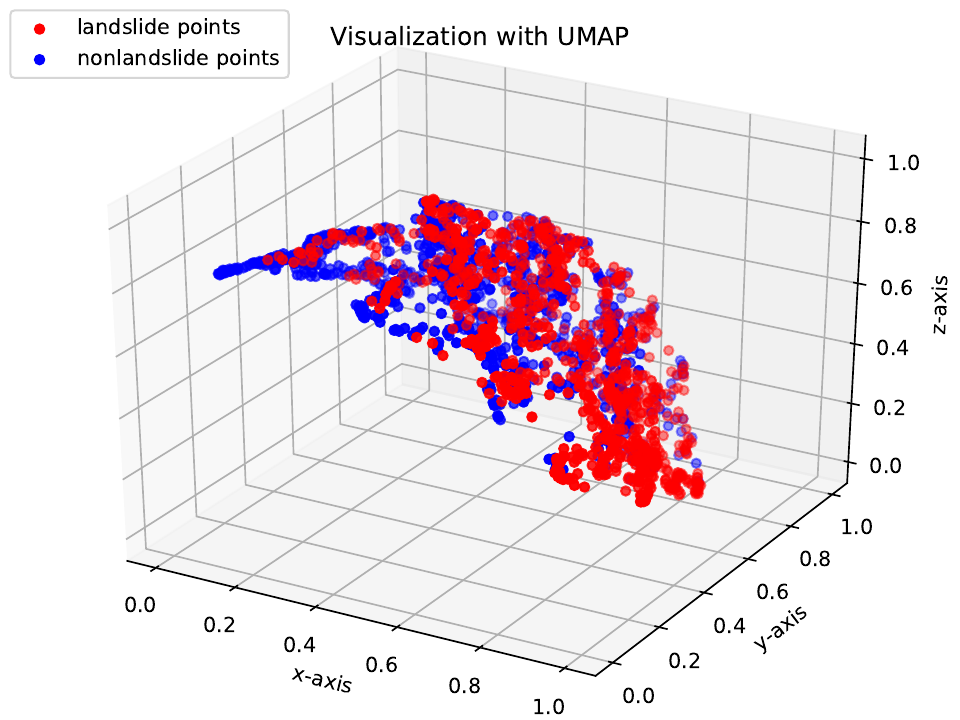}}\end{minipage}  \\ \hline
    \end{tabular}
    \label{tab:embedding space}
\end{table*}

\begin{table*}[h]
    \caption{The ablation experiment where we meta-train part of FJ and FL, and test the other part of FJ and FL. ($M$ is the number of samples within a task)}
    \centering
    \begin{tabular}{ccccc}
        \hline
        \multirow{2}{*}{Ablation experiment}                         &\multirow{2}{*}{Backbone}               & \multicolumn{3}{c}{Number of labelled samples}  \\ \cline{3-5}
                                                                     &                                        & $K=1$          & $K=5$          & $K=M/2$         \\ \hline 
        \multirow{2}{*}{Without unsupervised pretraining}            & MAML-based                             & 67.1$\pm$4.5   & 68.8$\pm$4.2   & 72.3$\pm$3.4  \\
                                                                     & proposed                               & 67.4$\pm$3.6   & 72.1$\pm$\textcolor{blue}{3.1}   & \textcolor{blue}{75.5}$\pm$3.3  \\ \hline
        \multirow{2}{*}{With unsupervised pretraining}               & MAML-based                             & 69.0$\pm$3.7   & 70.9$\pm$3.1   & 73.7$\pm$2.9  \\
                                                                     & proposed                               & 69.5$\pm$2.8   & 73.4$\pm$2.4   & \textcolor{blue}{76.9}$\pm$\textcolor{blue}{2.1}  \\ \hline
    \end{tabular}
    \label{tab:ablation experiment}
\end{table*}

The essence of these visualization approaches is to explore a discriminative embedding space, wherein the consecutiveness or aggregation of points of the same kind can reflect the performance of the visualization approach \citep{becht2019dimensionality}. 
Certainly, the unsupervised training process finds a representation that is easier to be embedded in a discriminative space that is suited for supervision tasks. 
In addition, we present the measurement performance of the ablation experiment, as shown in Table \ref{tab:ablation experiment}. 
The overall performance improvements proved the validity of the unsupervised pretraining process.

\subsection{Analysis of the meta-learner}
\label{subs:analysis of the meta-learner}
\subsubsection{Weakening the adverse effect of inaccurate supervision}
\begin{figure}[H]
    \centering
    \begin{subfigure}{0.45\textwidth}
        \includegraphics[width=\linewidth]{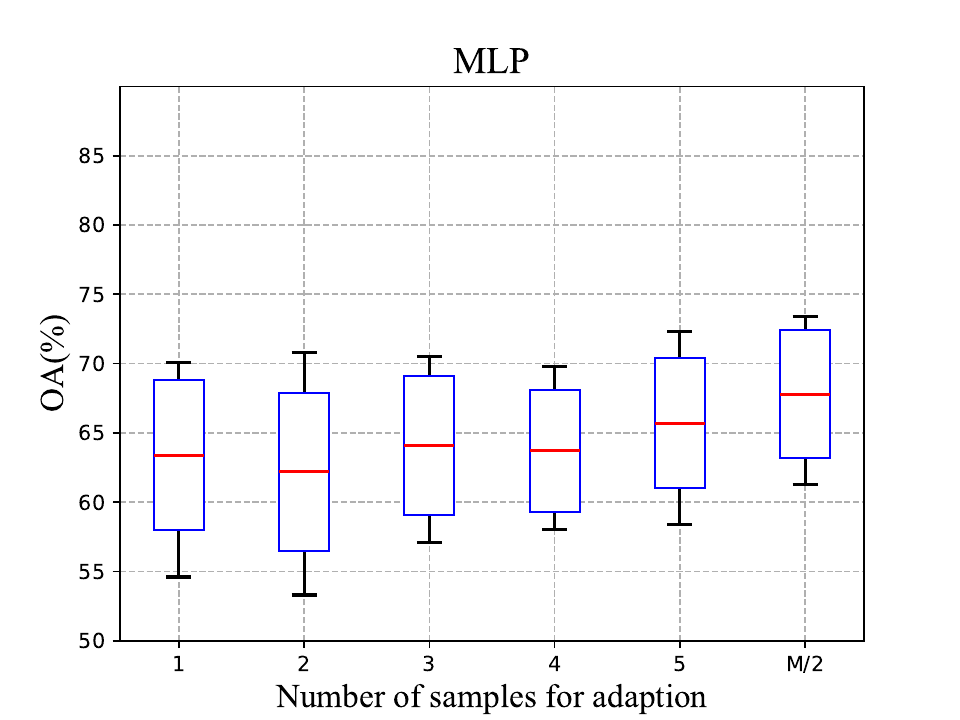}
    \end{subfigure}
    \begin{subfigure}{0.45\textwidth}
        \includegraphics[width=\linewidth]{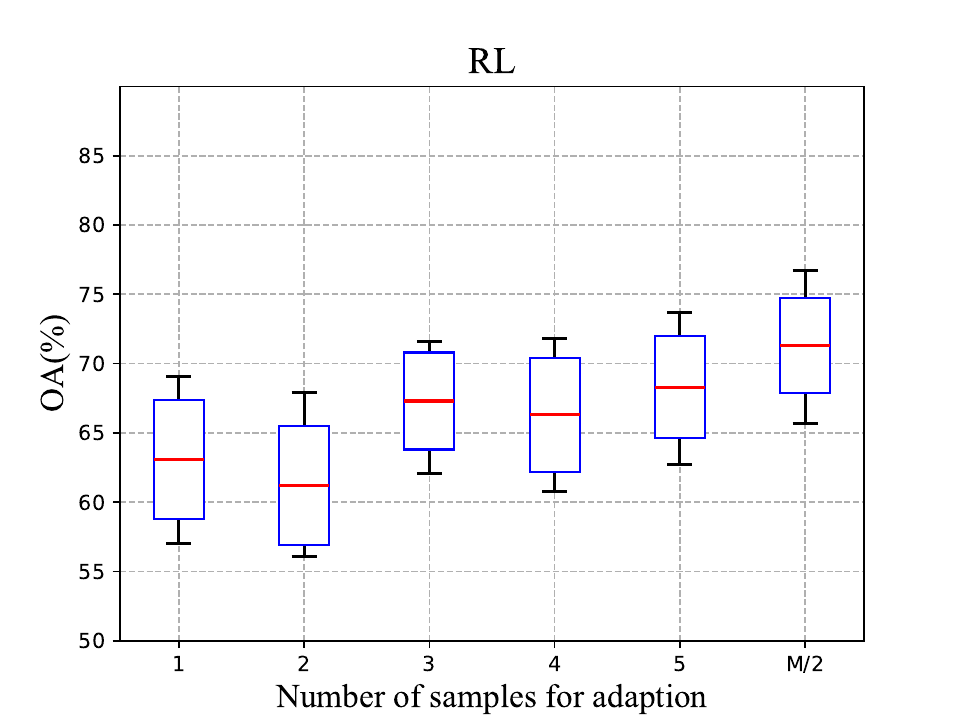}
    \end{subfigure}
	\begin{subfigure}{0.45\textwidth}
        \includegraphics[width=\linewidth]{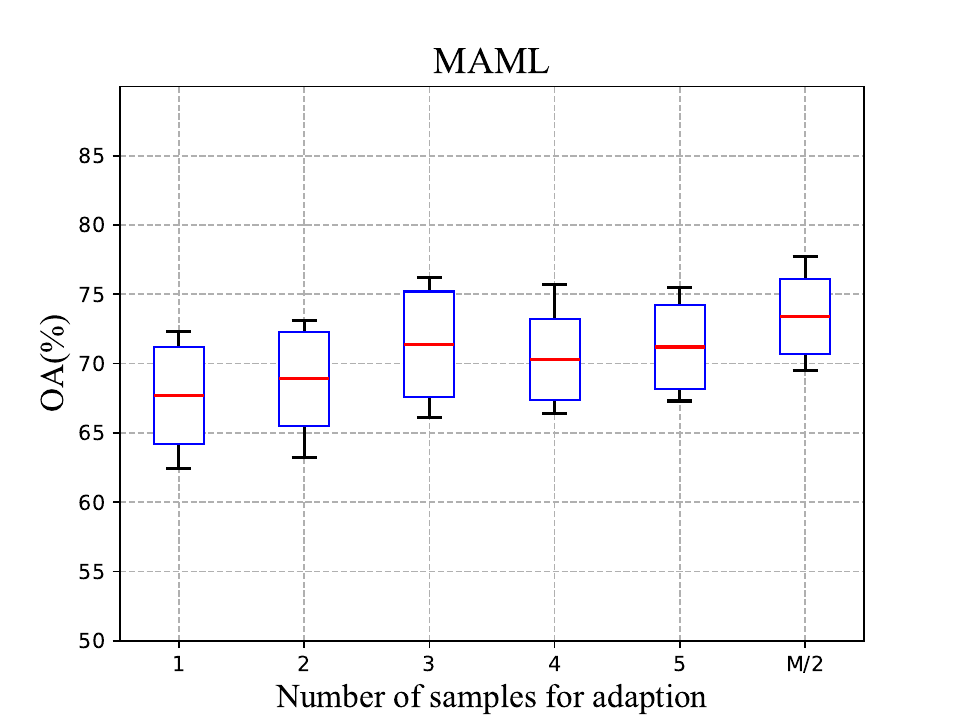}
    \end{subfigure}
    \begin{subfigure}{0.45\textwidth}
        \includegraphics[width=\linewidth]{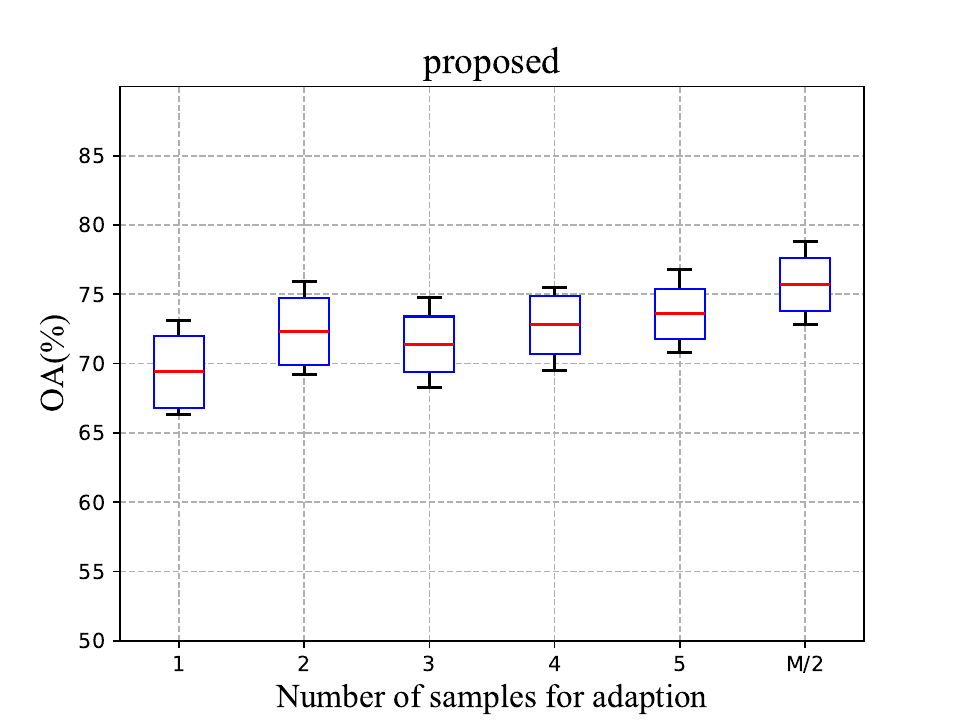}
    \end{subfigure}
    \caption{The candle charts for evaluating the overall performance of different methods. 
    The red line represents the mean OA, which lies in the middle of the bar. The height of the bar indicates two times the standard deviation. The horizontal black line indicates the maximum OA and minimum OA.}
    \label{fig:candle charts}
\end{figure}

One crucial problem this study aims to deal with is inaccurate supervision, in which labeled samples do not always reflect the ground truth \citep{zhang2021learning}. 
It occurs because landslide samples are collected from historical records and potential inferred landslides. 
The occurrence of historical landslides is a certain contingency, and experts infer potential landslides according to prior knowledge, both of which would lead to certain inaccuracies in the marking of landslide samples. 
That is different from a picture sample from which precise contents and objects can be confirmed, or samples can be identified with definite labeling meanings. 
Given this, the proposed meta-learner designed a trick to weigh the task losses to alleviate the adverse effects of inaccurate supervision, as discussed in Section \ref{subs:meta-learner}. 
To compare the performance in terms of stability and overall accuracy (OA), we (meta) train part of the samples of FJ and FL and then test the other part of the samples of FJ and FL. 
The performance can be seen in Fig. \ref{fig:candle charts}. 

The OA of the MLP-based method did not improve significantly as the number of samples increased, illustrating the limitation of learning with only a few samples. 
The RL-based method slightly improves but is unstable with a high standard deviation. 
The MAML-based method increases the OA to a new height but also suffers from instability owing to the adverse effect of inaccurate supervision. 
Instead, the proposed method realizes the best OA and stability. 
Because of the correlation of input features, direct feeding of the original data to the supervision task would mislead the learner to weigh the contribution of the features wrongly. 
It can reflect the embedding space of the MAML-based method that not a few positive and negative samples are still confusedly mixed in some dimensions. 

\subsubsection{Few-shot learning performance and generalization ability comparison}
\label{subsubs:few-shot learning performance and generalization ability comparison}
\begin{figure}[H]
    \centering
    \begin{subfigure}{0.45\textwidth}
        \includegraphics[width=\linewidth]{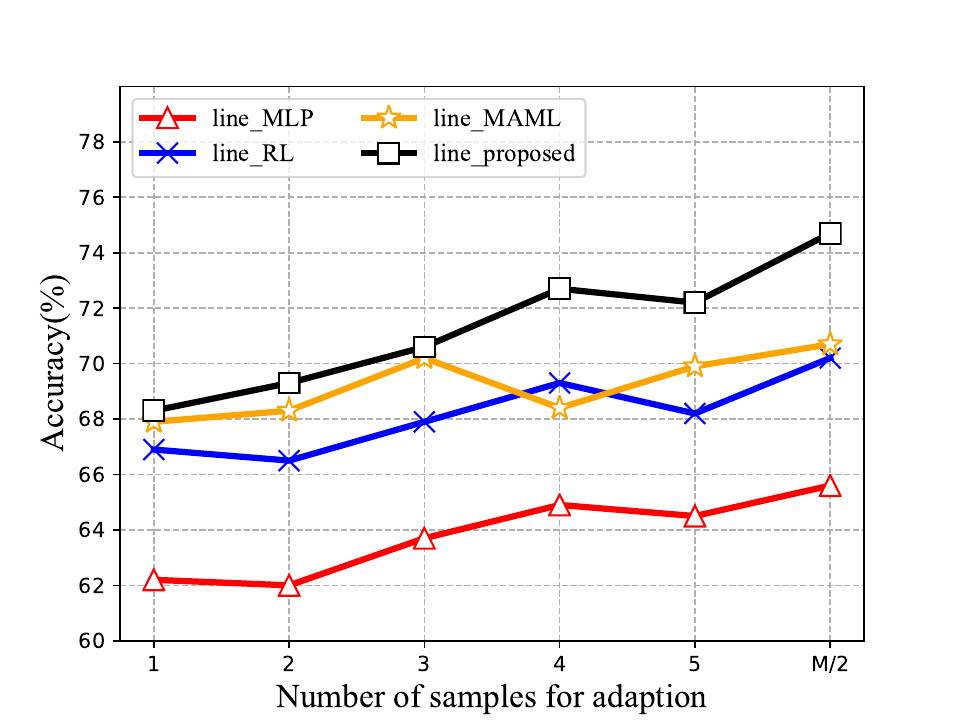}
        \caption{Experiment \textit{A}}
    \end{subfigure}
    \begin{subfigure}{0.45\textwidth}
        \includegraphics[width=\linewidth]{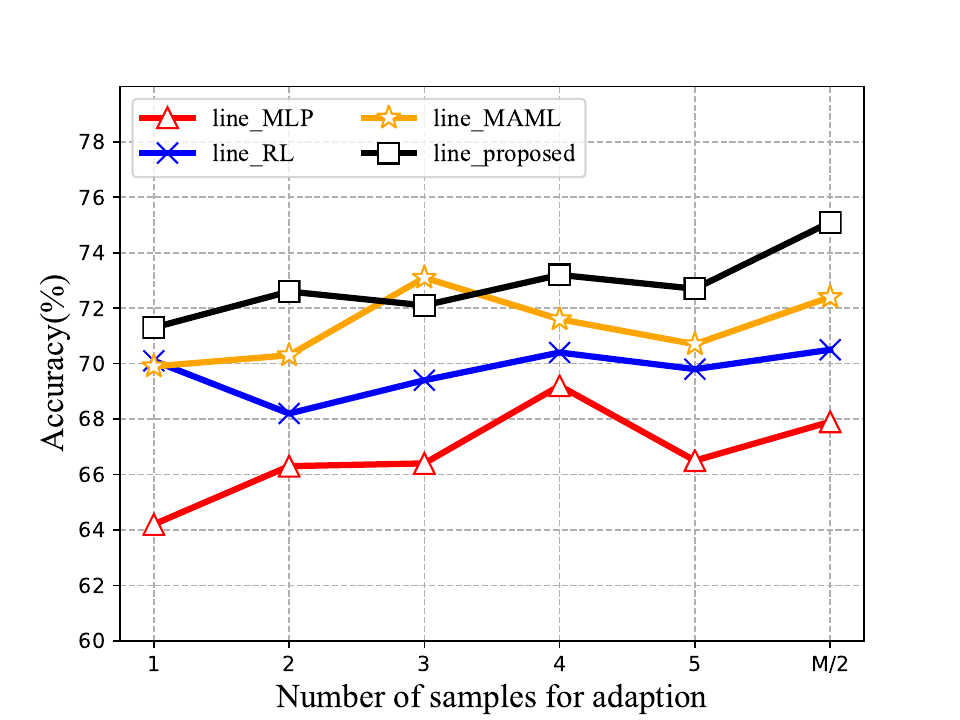}
        \caption{Experiment \textit{B}}
    \end{subfigure}
	\begin{subfigure}{0.45\textwidth}
        \includegraphics[width=\linewidth]{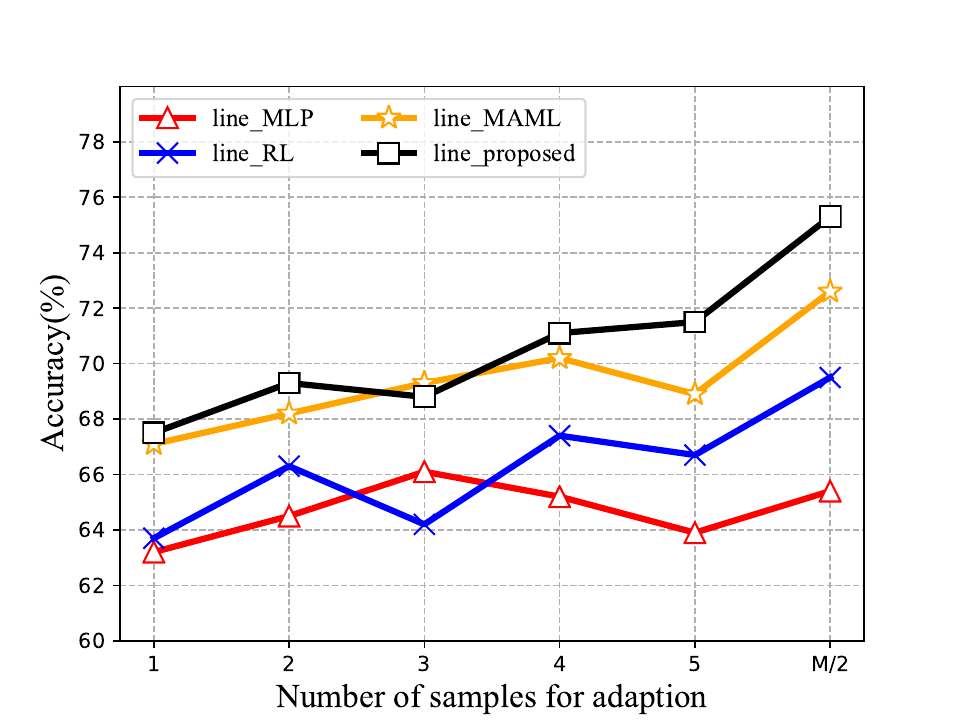}
        \caption{Experiment \textit{C}}
    \end{subfigure}
    \begin{subfigure}{0.45\textwidth}
        \includegraphics[width=\linewidth]{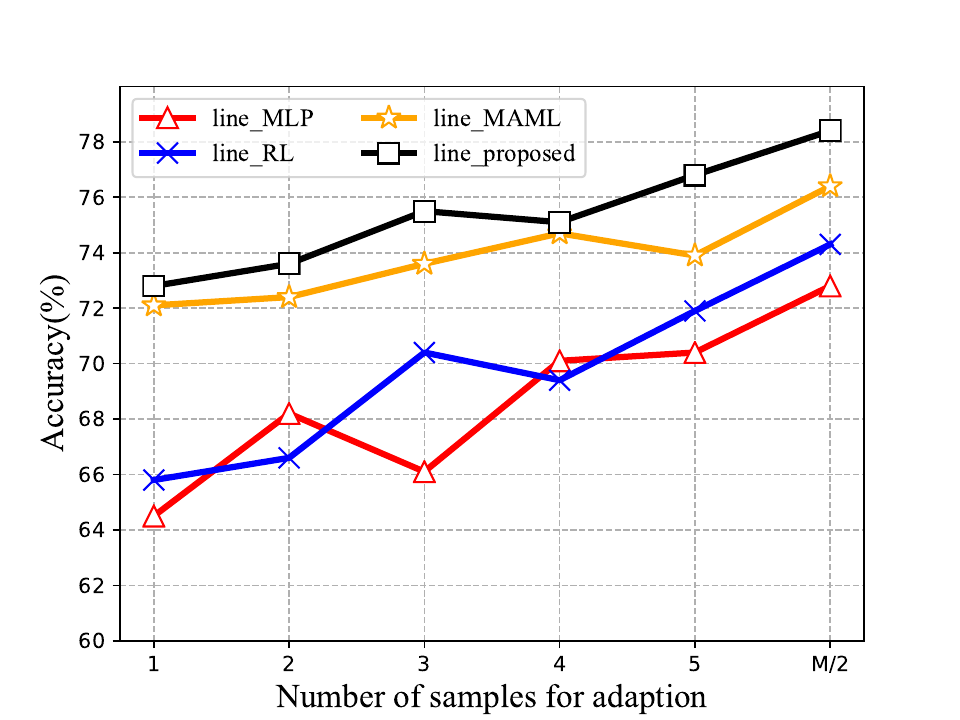}
        \caption{Experiment \textit{D}}
    \end{subfigure}
    \caption{OA performance with different numbers of labeled samples.}
    \label{fig:broken lines}
\end{figure}

To comprehensively explore the few-shot learning performance and generalization ability comparison of various methods, we designed four groups of comparative experiments, as shown in Tables \ref{tab:A experiment}, \ref{tab:B experiment}, \ref{tab:C experiment}, and \ref{tab:D experiment}, where $K$ denotes the number of samples used for adaptation, and $M$ ($M>5$) denotes the number of samples within a task. 
The methods that participated in the comparison were classified as transferring-based and meta-learning-based for their adaption. 
The four experiments were: \textit{A}. (meta) training part samples of FJ, and then testing the other part samples of FJ; \textit{B}. (meta) training all samples of FJ, and then testing all samples of FL; 
\textit{C}. (meta) training part samples of FJ and part samples of FL, then testing the other part samples of FJ; \textit{D}. (meta) training all samples of FJ and part samples of FL, then testing the other part samples of FL. 
The number of gradient descent updates for adaption in all experiments was set to 5.  
Fig. \ref{fig:AUROC} gives the receiver operating characteristic (ROC) curves of these methods in each experiment, where class 1 and class 2 represent the non-landslide and landslide classes, respectively.
Also, Fig. \ref{fig:broken lines} present the OA performance with a different number of labeled samples.

\begin{table*}[htb]
    \caption{Overall accuracy (OA) of various methods in experiment \textit{A}: (meta) training part samples of FJ, and then testing the other part of FJ.}
    \centering
    \begin{tabular}{cccccc}
        \hline
        \multirow{2}{*}{Experiment}                         &\multirow{2}{*}{Categories}               &\multirow{2}{*}{Backbone}               & \multicolumn{3}{c}{Number of labelled samples}  \\ \cline{4-6}
                                                            &                                          &                                        & $K=1$          & $K=5$          & $K=M/2$         \\ \hline 
        \multirow{4}{*}{\textit{A}}                         &\multirow{2}{*}{Transferring-based}       & MLP-based                              & 62.2$\pm$4.8   & 64.5$\pm$4.9   & 65.6$\pm$4.2  \\
                                                            &                                          & RL-based                               & 66.9$\pm$4.3   & 68.2$\pm$3.5   & 70.2$\pm$3.1  \\ \cline{2-6}
                                                            &\multirow{2}{*}{Meta-learning-based}      & MAML-based                             & 67.9$\pm$3.4   & 69.9$\pm$2.3   & 70.7$\pm$1.7  \\
                                                            &                                          & Proposed                               & 68.3$\pm$2.4   & 72.2$\pm$1.6   & \textcolor{blue}{74.7}$\pm$\textcolor{blue}{0.7}  \\ \hline
    \end{tabular}
    \label{tab:A experiment}
\end{table*}
\begin{table*}[htb]
    \caption{Overall accuracy (OA) of various methods in experiment \textit{B}: (meta) training all samples of FJ, then testing all samples of FL.}
    \centering
    \begin{tabular}{cccccc}
        \hline
        \multirow{2}{*}{Experiment}                         &\multirow{2}{*}{Categories}               &\multirow{2}{*}{Backbone}               & \multicolumn{3}{c}{Number of labelled samples}  \\ \cline{4-6}
                                                            &                                          &                                        & $K=1$          & $K=5$          & $K=M/2$         \\ \hline 
        \multirow{4}{*}{\textit{B}}                         &\multirow{2}{*}{Transferring-based}       & MLP-based                              & 64.2$\pm$6.2   & 66.5$\pm$5.4   & 67.9$\pm$4.1  \\
                                                            &                                          & RL-based                               & 70.1$\pm$4.6   & 69.8$\pm$4.7   & 70.5$\pm$3.2  \\ \cline{2-6}
                                                            &\multirow{2}{*}{Meta-learning-based}      & MAML-based                             & 69.9$\pm$3.2   & 70.7$\pm$2.9   & 72.4$\pm$2.6  \\
                                                            &                                          & Proposed                               & 71.3$\pm$2.5   & 72.7$\pm$2.3   & \textcolor{blue}{75.1}$\pm$\textcolor{blue}{1.9}  \\ \hline
    \end{tabular}
    \label{tab:B experiment}
\end{table*}
\begin{table*}[htb]
    \caption{Overall accuracy (OA) of various methods in experiment \textit{C}: (meta) training part samples of FJ and part samples of FL, then testing the other part of FJ.}
    \centering
    \begin{tabular}{cccccc}
        \hline
        \multirow{2}{*}{Experiment}                         &\multirow{2}{*}{Categories}               &\multirow{2}{*}{Backbone}               & \multicolumn{3}{c}{Number of labelled samples}  \\ \cline{4-6}
                                                            &                                          &                                        & $K=1$          & $K=5$          & $K=M/2$         \\ \hline 
        \multirow{4}{*}{\textit{C}}                         &\multirow{2}{*}{Transferring-based}       & MLP-based                              & 63.2$\pm$5.5   & 63.9$\pm$5.9   & 65.4$\pm$4.5  \\
                                                            &                                          & RL-based                               & 63.7$\pm$4.9   & 66.7$\pm$3.1   & 69.5$\pm$3.2  \\ \cline{2-6}
                                                            &\multirow{2}{*}{Meta-learning-based}      & MAML-based                             & 67.1$\pm$3.2   & 68.9$\pm$2.7   & 72.6$\pm$2.4  \\
                                                            &                                          & Proposed                               & 67.5$\pm$2.0   & 71.5$\pm$\textcolor{blue}{1.4}   & \textcolor{blue}{75.3}$\pm$1.7  \\ \hline
    \end{tabular}
    \label{tab:C experiment}
\end{table*}
\begin{table*}[htb]
    \caption{Overall accuracy (OA) of various methods in experiment \textit{D}: (meta) training all samples of FJ and part samples of FL, then testing the other part FL.}
    \centering
    \begin{tabular}{cccccc}
        \hline
        \multirow{2}{*}{Experiment}                         &\multirow{2}{*}{Categories}               &\multirow{2}{*}{Backbone}               & \multicolumn{3}{c}{Number of labelled samples}  \\ \cline{4-6}
                                                            &                                          &                                        & $K=1$          & $K=5$          & $K=M/2$         \\ \hline 
        \multirow{4}{*}{\textit{D}}                         &\multirow{2}{*}{Transferring-based}       & MLP-based                              & 64.5$\pm$6.2   & 70.4$\pm$6.1   & 72.8$\pm$5.4  \\
                                                            &                                          & RL-based                               & 65.8$\pm$5.6   & 71.9$\pm$4.1   & 74.3$\pm$4.6  \\ \cline{2-6}
                                                            &\multirow{2}{*}{Meta-learning-based}      & MAML-based                             & 72.1$\pm$4.2   & 73.2$\pm$3.8   & 76.4$\pm$3.5  \\
                                                            &                                          & Proposed                               & 72.8$\pm$3.0   & 75.3$\pm$2.6   & \textcolor{blue}{78.4}$\pm$\textcolor{blue}{2.5}  \\ \hline
    \end{tabular}
    \label{tab:D experiment}
\end{table*}

\begin{figure}[H]
    \centering
    \begin{subfigure}{0.4\textwidth}
        \includegraphics[width=\linewidth]{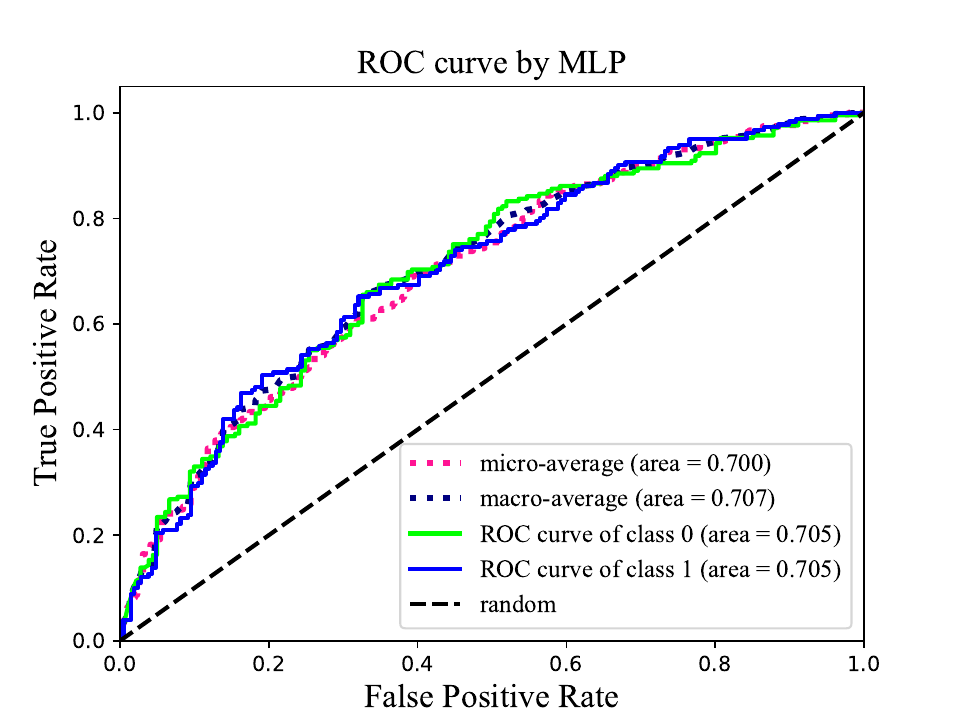}
    \end{subfigure}
    \begin{subfigure}{0.4\textwidth}
        \includegraphics[width=\linewidth]{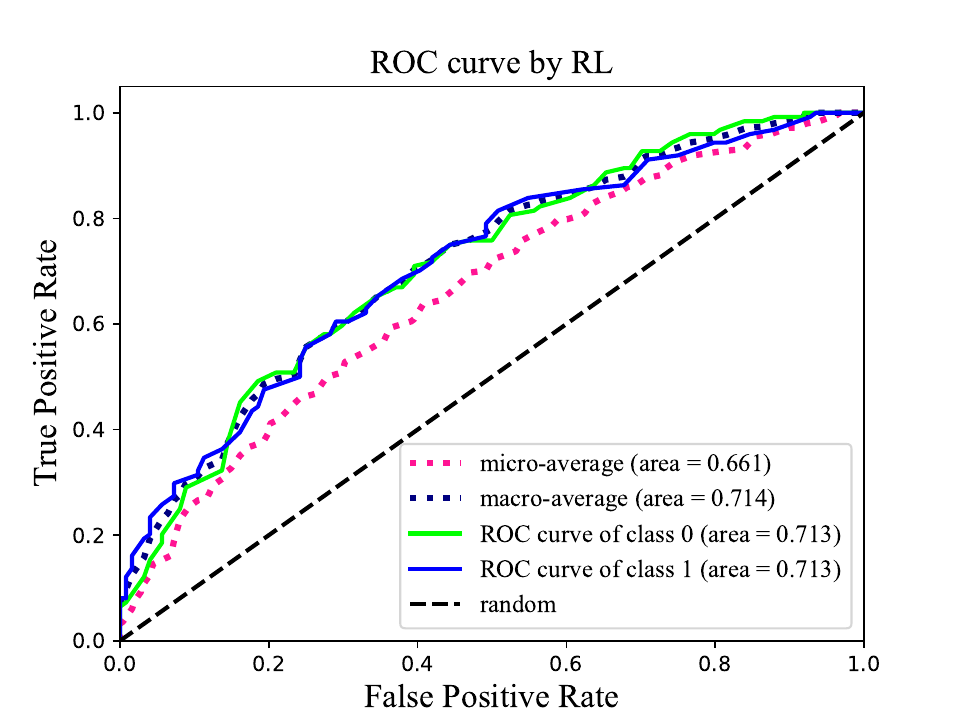}
    \end{subfigure}
    \begin{subfigure}{0.4\textwidth}
        \includegraphics[width=\linewidth]{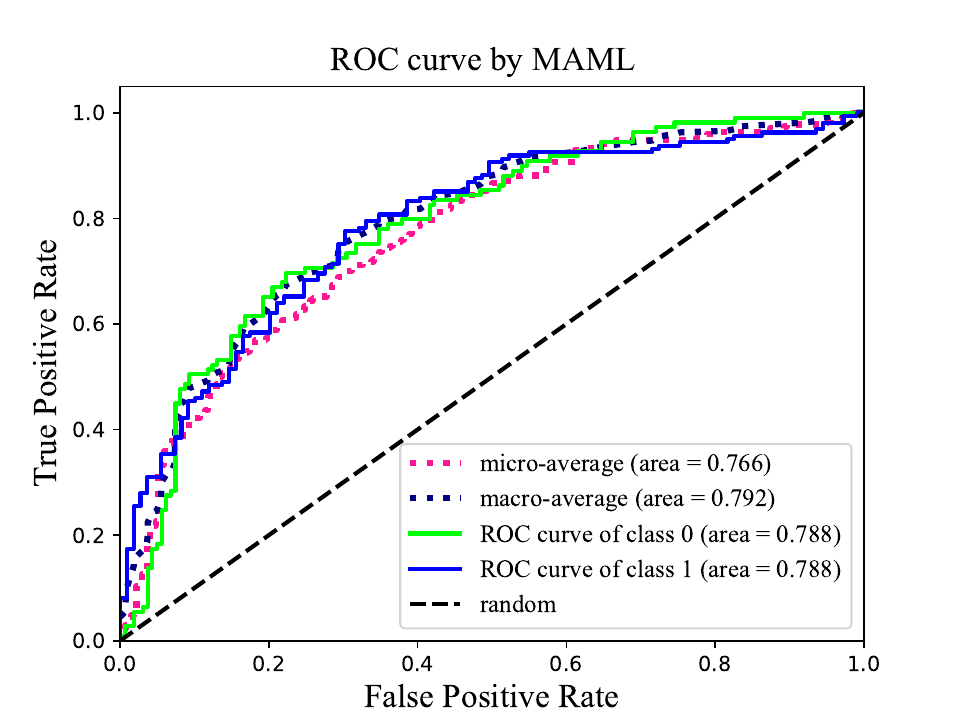}
    \end{subfigure}
	\begin{subfigure}{0.4\textwidth}
        \includegraphics[width=\linewidth]{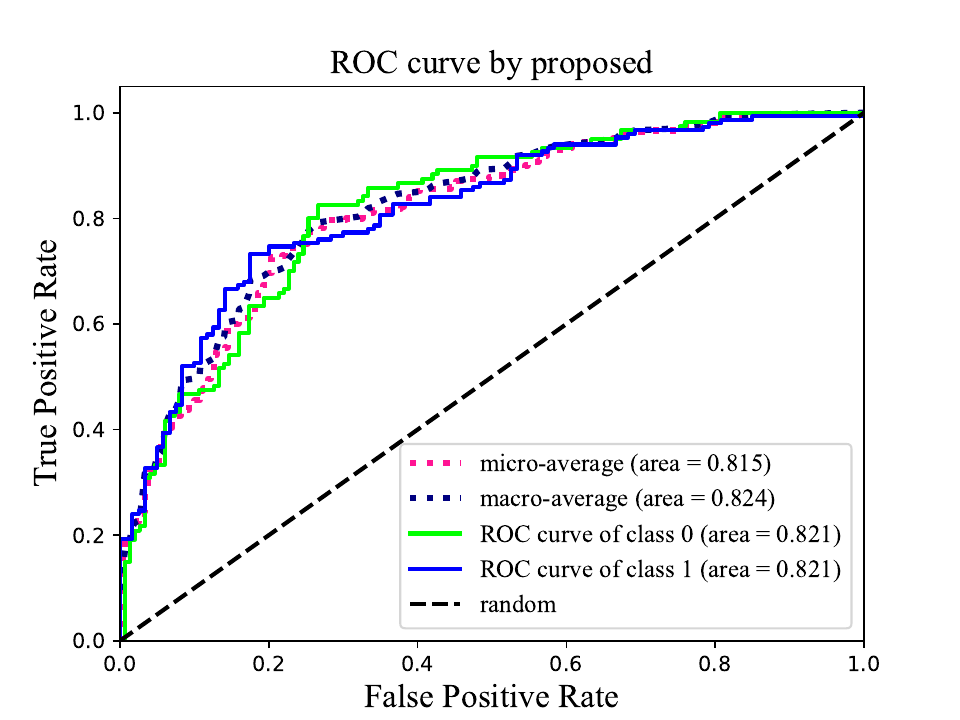}
    \end{subfigure}
    \caption{ROC curves of Experiment \textit{A}.}
    \label{fig:AUROC_A}
\end{figure}

\begin{figure}[H]
    \centering
    \begin{subfigure}{0.4\textwidth}
        \includegraphics[width=\linewidth]{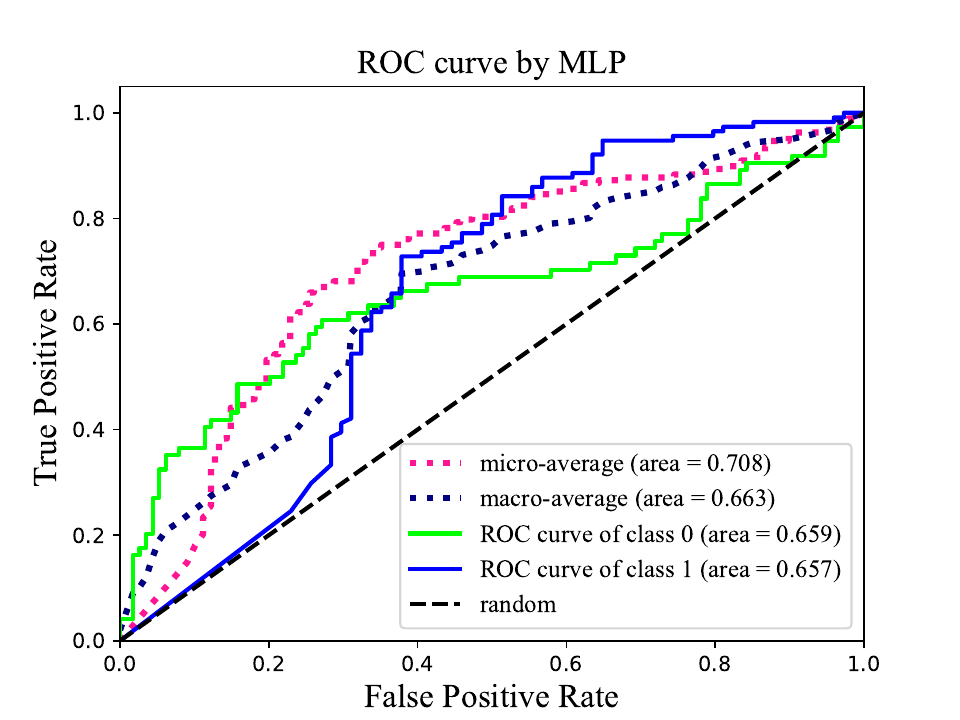}
    \end{subfigure}
    \begin{subfigure}{0.4\textwidth}
        \includegraphics[width=\linewidth]{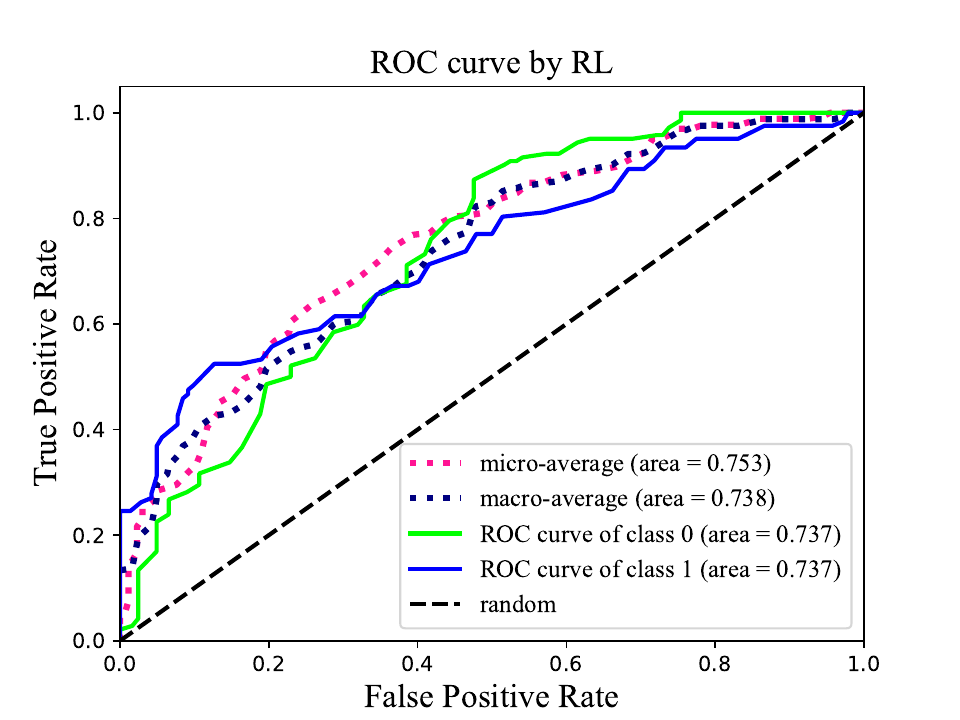}
    \end{subfigure}
    \begin{subfigure}{0.4\textwidth}
        \includegraphics[width=\linewidth]{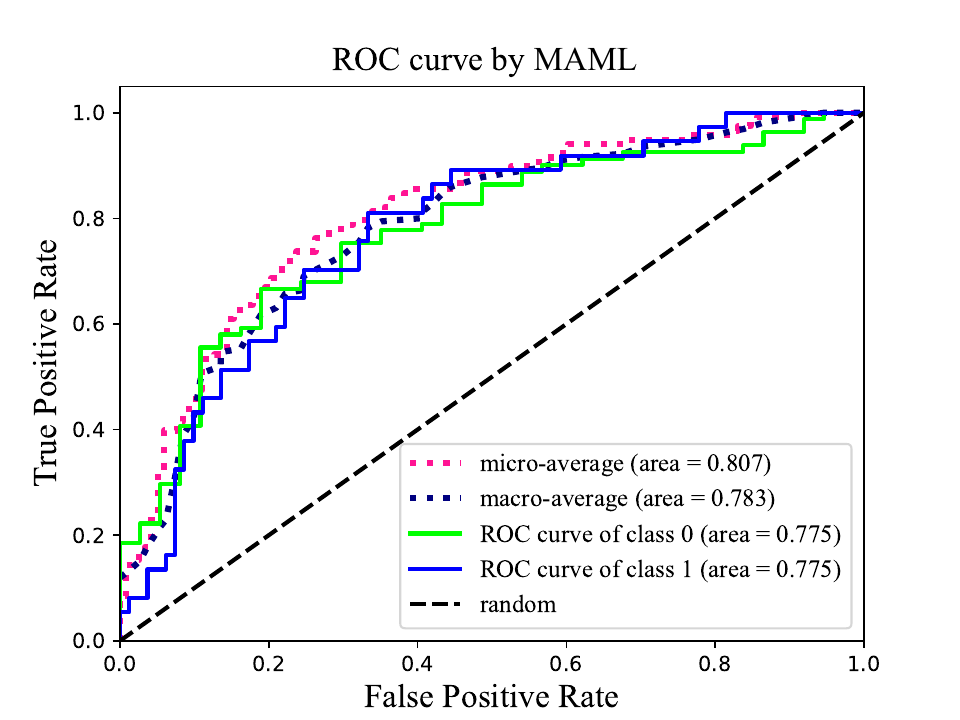}
    \end{subfigure}
	\begin{subfigure}{0.4\textwidth}
        \includegraphics[width=\linewidth]{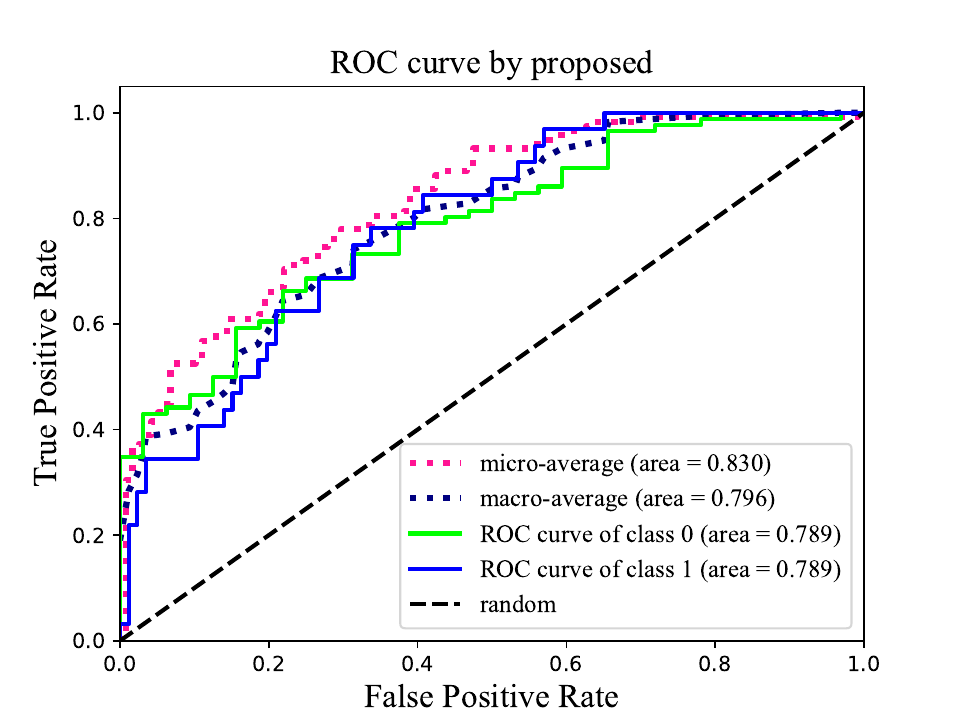}
    \end{subfigure}
    \caption{ROC curves of Experiment \textit{B}.}
    \label{fig:AUROC_B}
\end{figure}

\begin{figure}[H]
    \centering
    \begin{subfigure}{0.4\textwidth}
        \includegraphics[width=\linewidth]{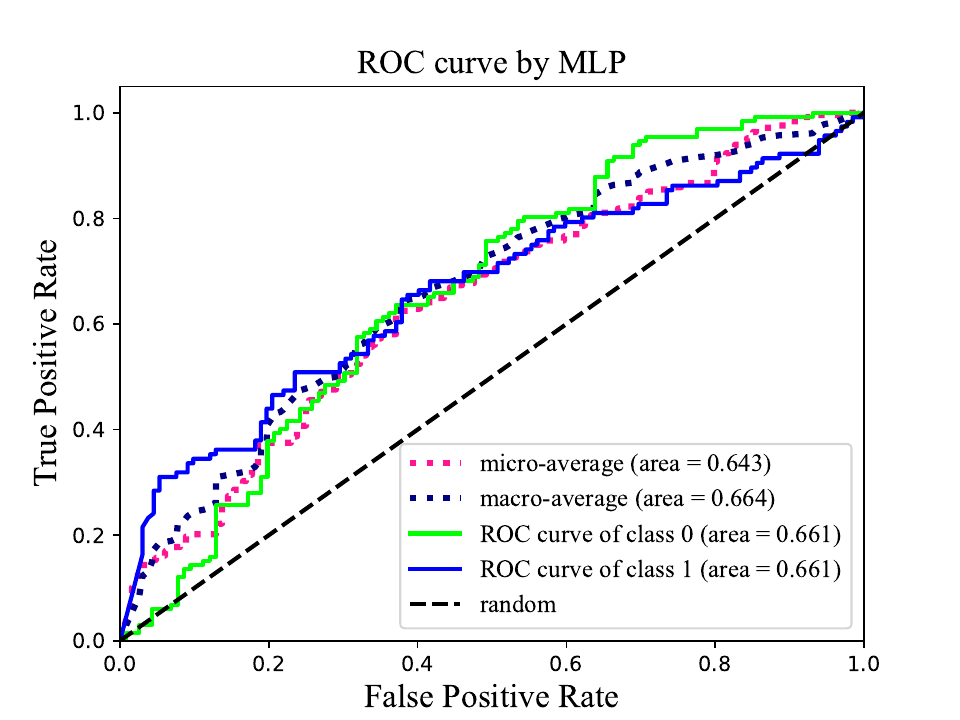}
    \end{subfigure}
    \begin{subfigure}{0.4\textwidth}
        \includegraphics[width=\linewidth]{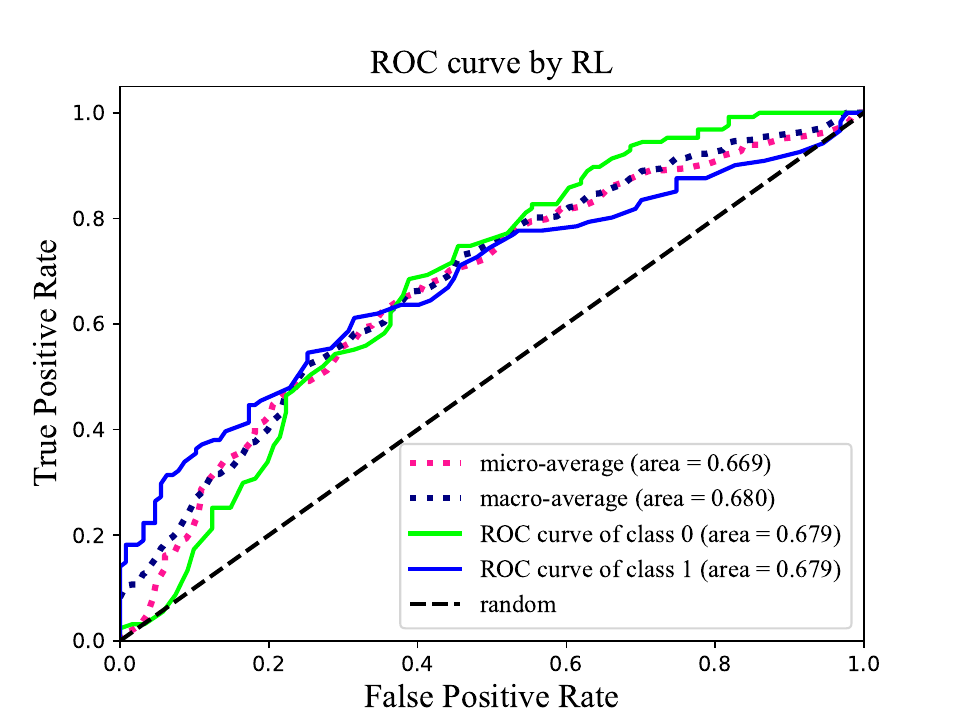}
    \end{subfigure}
    \begin{subfigure}{0.4\textwidth}
        \includegraphics[width=\linewidth]{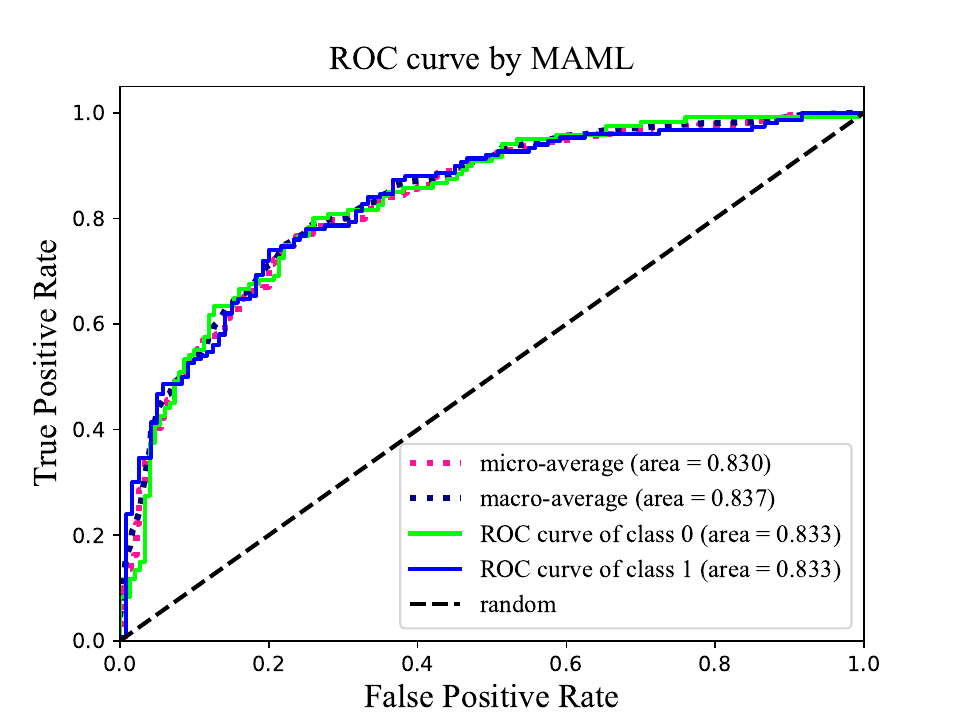}
    \end{subfigure}
	\begin{subfigure}{0.4\textwidth}
        \includegraphics[width=\linewidth]{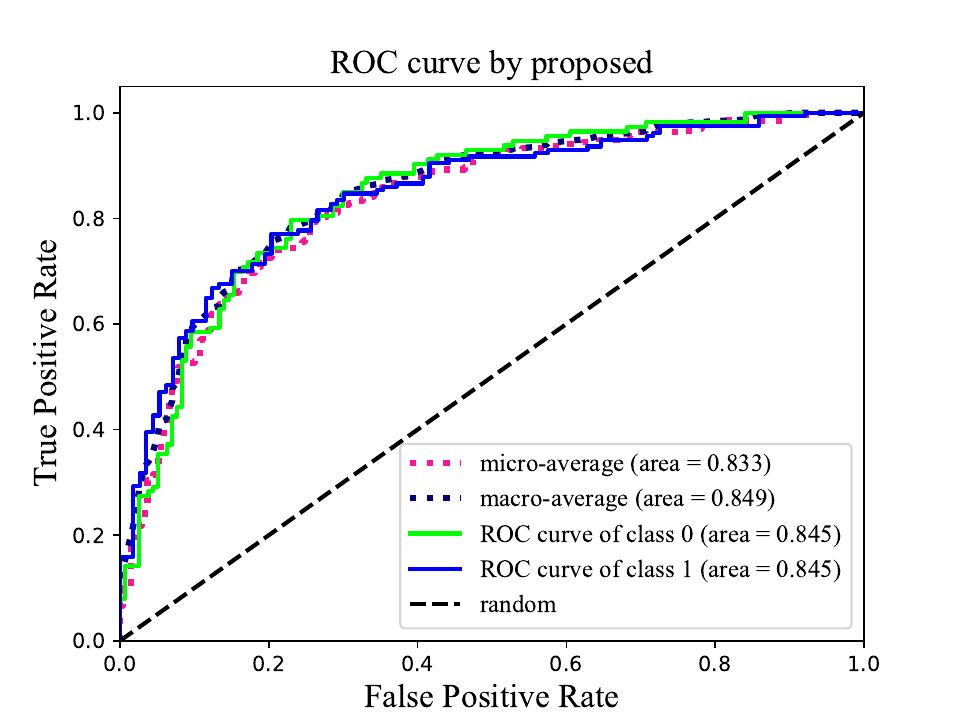}
    \end{subfigure}
    \caption{ROC curves of Experiment \textit{C}.}
    \label{fig:AUROC}
\end{figure}

\begin{figure}[H]
    \centering
    \begin{subfigure}{0.4\textwidth}
        \includegraphics[width=\linewidth]{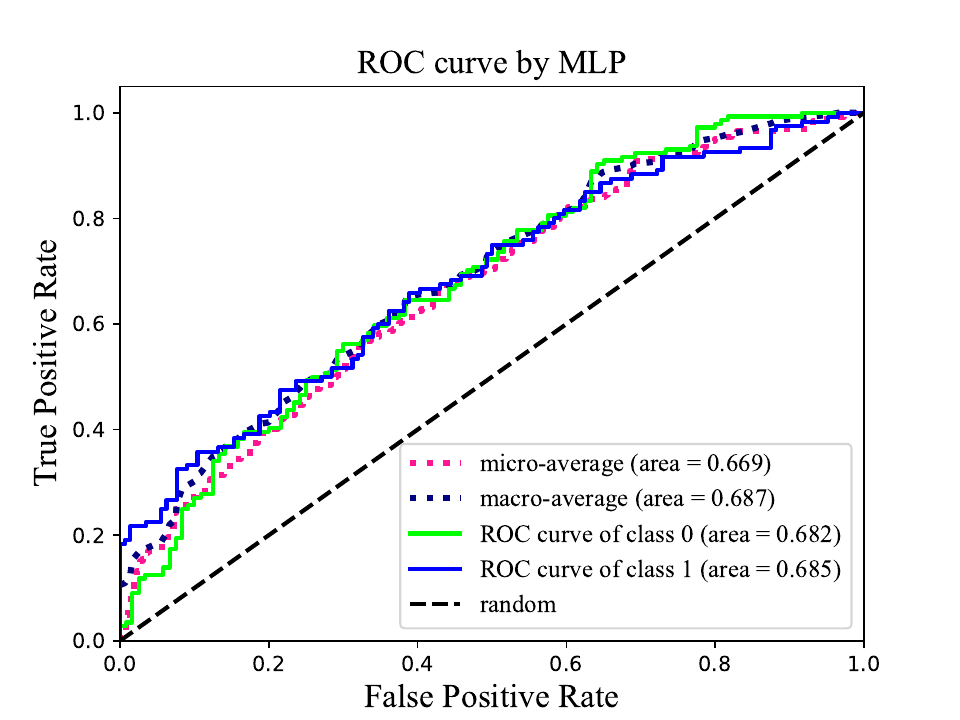}
    \end{subfigure}
    \begin{subfigure}{0.4\textwidth}
        \includegraphics[width=\linewidth]{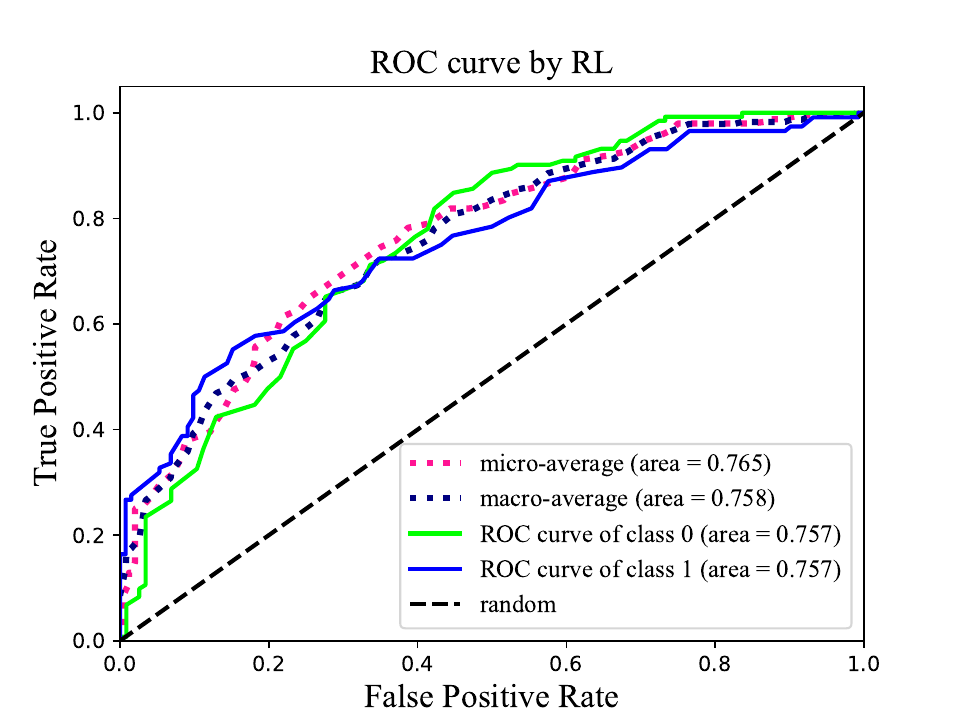}
    \end{subfigure}
    \begin{subfigure}{0.4\textwidth}
        \includegraphics[width=\linewidth]{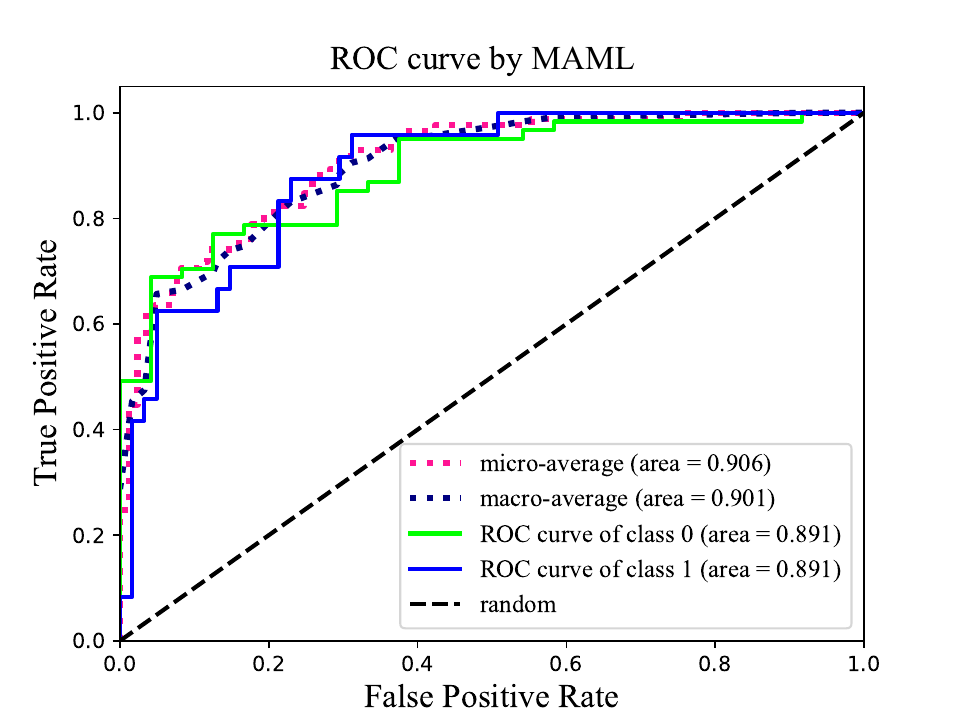}
    \end{subfigure}
	\begin{subfigure}{0.4\textwidth}
        \includegraphics[width=\linewidth]{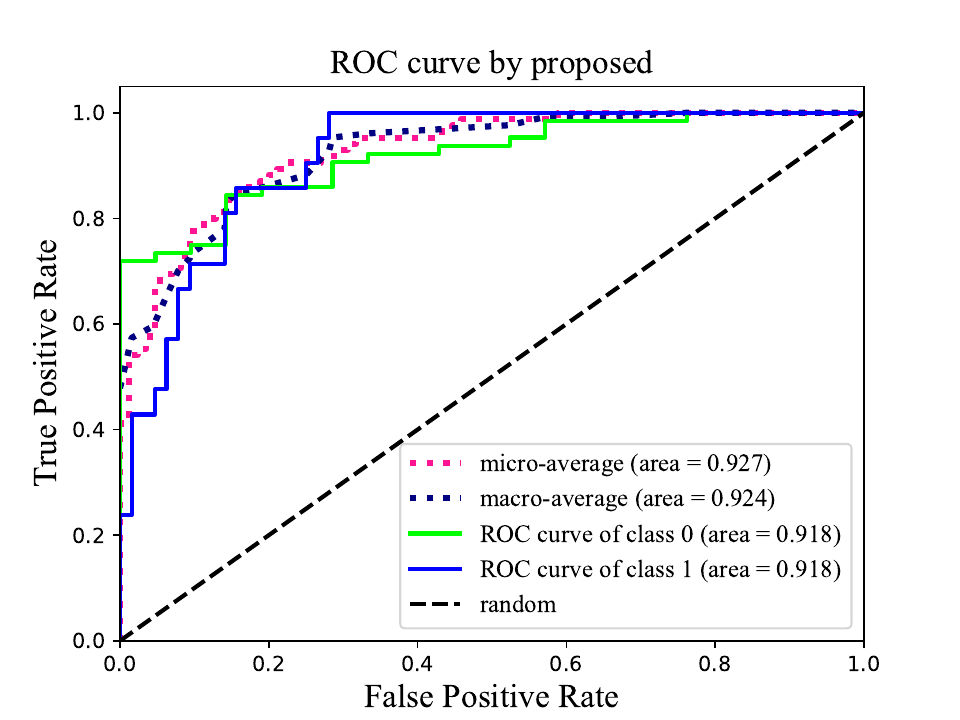}
    \end{subfigure}
    \caption{ROC curves of Experiment \textit{D}.}
    \label{fig:AUROC}
\end{figure}

From all the experiments, we find that the meta-learning-based methods evidently improve their performance with an increase in $K$, whereas the transferring-based methods improve relatively little. 
That demonstrates the validity of applying the meta-learning ideology to realize fast adaption with very few samples and iterations in the LSM territory. 
In addition, with our proposed tricks, we further improve the performance to the top level.

In experiments \textit{A} and \textit{B}, where there are plenty of samples in FJ, the few-shot learning performance of the proposed method is better than the others. 
From experiment \textit{B}, we find that the generalization ability of the proposed methods from FJ to FL also outperforms the others. 
It is worth noting that the accuracy in \cite{zhu2020unsupervised} is close to meta-learning-based methods when $K=1$ because the applied adversarial training has learned shared factors that can be directly transferred between FJ and FL scenarios \citep{goodfellow2014generative}.

From the comparison with experiment \textit{A} and \textit{C}, we can see that meta-training together with FJ and FL tasks improves the overall performance of meta-learning-based methods. 
In this study, we attribute this to the fact that productive tasks containing various landslide-inducing environments contribute to a more general intermediate model, that is, improving the generalization ability of the intermediate model. 
This claim can also be confirmed from a comparison with experiments \textit{B} and \textit{D}. 
Thus, experiment \textit{C} was used to predict the LSM of FJ, while experiment \textit{D} was used to predict the LSM of FL.

\section{Conclusion}
\label{s:conclusion}
In response to the two problems mentioned in Section \ref{subs:limited generalization ability of a single model} and \ref{subs:few samples for block-wise analysis}, our proposed method meta-learns an intermediate representation for the few-shot prediction of landslide susceptibility. 
The experiment and results demonstrate the validity of (1) applying unsupervised pretraining to discover discriminative embedding space; 
(2) segmenting large scenarios into pieces for individual analysis; and (3) meta-learning an intermediate representation for fast adaptation to new local tasks.

This study presents the first example of a block-wise prediction of landslide susceptibility in a meta-learning manner to the best of our knowledge. 
Considering the tiny amount of samples and gradient descent updates used when adapting to a new task, the proposed architecture outperforms in terms of stability to inaccurate supervision, generalization ability, and fast adaption performance. 
It proves the validity and superiority of applying our method to large scenarios where the landslide-inducing environment is various and complex, and landslide samples are very few. We also summarize the limitations of our proposed method as follows.

(1) Samples in meta-training were limited in FJ and FL, thus if possible, samples from other areas, accounting for different landslide causes, could be added into the meta-learner to improve the generalization ability and fast adaption ability of the intermediate model.

(2) Because many of the parameter settings, for example, the update learning rate, number of meta tasks, and number of adapting updates, are not determined automatically, but experimentally, it becomes necessary to validate the appropriate value by repeating the running of the code.

(3) The unbalanced distribution of positive and negative samples would cause an adverse tendency to predict LSM. For example, if one block is only full of positive samples, the adapted model will tend to give a positive prediction of that block. 
It is demonstrated in Fig. \ref{fig:LSM of FL}. Nevertheless, the proposed method still performs better than the other methods.

\section*{CRediT author statement}
\textbf{Li Chen}: Conceptualization, Methodology, Software, Investigation, Writing - Original Draft, Data Curation.
\textbf{Yulin Ding}: Supervision, Project administration, Funding acquisition.
\textbf{Saeid Pirasteh}: Formal analysis, Writing - Review \& Editing.
\textbf{Han Hu}: Methodology, Writing - Review \& Editing, Funding acquisition.
\textbf{Qing Zhu}: Project administration, Funding acquisition.
\textbf{Xuming Ge}: Resources.
\textbf{Haowei Zeng}: Validation.
\textbf{Haojia Yu}: Formal analysis.
\textbf{Qisen Shang}: Formal analysis.
\textbf{Yongfei Song}: Funding acquisition.

\section*{Declaration of Competing Interest}
The authors declare that they have no known competing financial interests or personal relationships that could have appeared to influence the work reported in this paper.

\section*{Acknowledgments}
This work was financially supported by the National Natural Science Foundation of China (grant No. 41941019, 41871291, and 42071355) and the Science and Technology Department of Ningxia (grant No. 2021BEG03001). 
The authors gratefully acknowledge the provision of landslide inventories from Chongqing Geomatics and Remote Sensing Center. 
Also, the authors are thankful to the science teams who produce and manage online datasets in CNIC and USGS.

\bibliographystyle{model2-names}
\bibliography{LSM_ML}

\end{document}